\documentclass[11pt]{article}

\usepackage[margin=1in]{geometry}
\usepackage[T1]{fontenc}
\usepackage[utf8]{inputenc}
\usepackage{lmodern}
\usepackage{microtype}
\usepackage{amsmath}
\usepackage{amssymb}
\usepackage{amsthm}
\usepackage{booktabs}
\usepackage{array}
\usepackage{longtable}
\usepackage{enumitem}
\usepackage{graphicx}
\usepackage{float}
\usepackage{flafter}
\usepackage{placeins}
\usepackage{caption}
\usepackage{fvextra}
\usepackage[authoryear,round]{natbib}
\usepackage{hyperref}

\hypersetup{
  colorlinks=true,
  linkcolor=blue,
  urlcolor=blue
}

\newcommand{\dep}{\mathrm{dep}}

\newcommand{\TrainExact}{\operatorname{TrainExact}}
\newcommand{\HeldoutExact}{\operatorname{HeldoutExact}}

\title{ReplaySCM: A Benchmark for Executable Causal Mechanism Induction from Interventions}
\author{Serafim Batzoglou\\\texttt{serafim.batzoglou@gmail.com}}
\date{}

\begin{document}
\makeatletter
\begin{center}
\vspace*{0.12in}
{\LARGE\bfseries \@title\par}
\vskip 0.8em
{\large \@author\par}
\end{center}
\makeatother

\begin{abstract}
Most causal benchmarks for language models score local answers or graph structure. We introduce ReplaySCM, a 1,300-item benchmark for executable causal mechanism induction from finite interventional evidence. Each item contains binary worlds generated by a latent fully observed acyclic Boolean structural causal model (SCM). A system must output a mechanism map in a restricted Boolean DSL; the submission is parsed, checked for legality and acyclicity, and replayed on training and held-out intervention worlds. Scoring uses replay behavior rather than formula strings, so syntactically different mechanisms receive credit when they behave correctly. ReplaySCM varies the structural information disclosed to the model through Ordered, Block-order, Hidden-order, and Hidden-roots settings, and includes Alternative-SCM tasks that supply a valid reference SCM and ask for a semantically distinct alternative that fits the training worlds, together with a separating intervention and witness. Frontier LLMs infer parts of the functional-parent structure, but held-out replay drops sharply when order or root structure is hidden. We also evaluate a matched support-audit ladder—Original, Extra Worlds, and Counterexample Audit (CEx)—that raises mean local predecessor-pattern coverage from 0.8949 to 0.9815 to 1.0; under the audited searches, no discovered semantic alternative remains consistent with the training worlds. The Ordered/Hidden-order gap persists under this stronger evidence. ReplaySCM complements answer-level causal reasoning and graph-discovery benchmarks by evaluating executable replay generalization from finite interventional evidence, without claiming unique identification of the latent SCM.
\end{abstract}

\section{Introduction}

Recent causal benchmarks for language models usually score local outputs: an answer to a causal question, a predicted counterfactual value, a
graph edge, or a written explanation. These tasks are useful, but they leave open whether a system can output a causal model
that can be reused under new interventions. We introduce ReplaySCM, a 1,300-item benchmark for executable causal mechanism induction from 
finite interventional evidence in a controlled Boolean setting.

Structural causal models (SCMs) are a natural target for this evaluation because they represent mechanisms and interventions in one
formal object \citep{pearl2009causality}. ReplaySCM uses small, finite, binary, fully observed, acyclic SCMs whose
endogenous mechanisms are Boolean expressions. ReplaySCM asks for a mechanism map in a restricted Boolean DSL.
The evaluator parses the map, checks legality and acyclicity, and replays it on both training and held-out intervention 
worlds. Scoring is semantic: formulas receive credit for their replay behavior, independent of their textual form.

{\looseness=-1
The benchmark is organized around matched versions of the same latent SCMs. A revealed-structure ladder varies what the model is 
told: Ordered gives the full topological order, Block-order gives only coarse precedence blocks, Hidden-order hides the endogenous 
order, and Hidden-roots also hides the root set. Two Alternative-SCM tasks, Ordered and Hidden-order, 
supply a valid reference SCM and ask for a semantically distinct 
alternative with a separating intervention and witness. Extra Worlds and Counterexample Audit (CEx) tasks (Ordered and Hidden-order) progressively add
training worlds to reduce finite-evidence ambiguity. In CEx, no discovered semantic alternative from LLM outputs, symbolic exact-search, or 50-seed bnlearn+DSL searches fits the training worlds. The matched revealed-structure settings isolate disclosure on the same latent SCMs; CEx preserves latent SCMs and held-out worlds while auditing ambiguity.

\par}

{\looseness=-1
We evaluate frontier LLMs and include two non-LLM calibration rows.
The empirical picture is consistent across these comparisons. Frontier LLMs often infer some functional-parent relationships, 
but exact executable replay remains hard, especially when order or roots are hidden. Held-out replay is much higher among responses 
that exactly fit all training worlds, while responses that fail on the training worlds rarely replay all held-out worlds. 
Supplying a valid SCM in Alternative-SCM substantially raises performance, showing that local editing with a supplied causal 
object is easier than inferring that object from intervention worlds. The Ordered/Hidden-order gap persists in the
Extra Worlds and CEx settings.
\par}

{\looseness=-1
\textbf{Contributions.} We make four contributions. First, we introduce executable-SCM induction from interventions as a publicly released
benchmark evaluated by exact replay on training and held-out worlds. Second, we organize the benchmark as a matched
revealed-structure ladder—Ordered, Block-order, Hidden-order, and Hidden-roots—together with Alternative-SCM, which separates
reasoning with a supplied SCM from hidden-structure inference. Third, we build a generator that filters out trivial
shortcuts and tracks residual ambiguity through support filters, shortcut checks, bounded audits, and a three-level matched support-audit ladder: Original, Extra Worlds, and Counterexample Audit (CEx).
Fourth, we benchmark frontier LLMs under a shared evaluator and use two non-LLM
baselines to calibrate the difficulty of executable mechanism induction.

We release a public artifact of benchmark instances, prompt-export and scoring code, replay and validation scripts, and documentation for reproducing the reported evaluations.

\par}

\section{Related Work}

Causal reasoning benchmarks for language models mostly evaluate local outputs: commonsense causal judgments, intervention questions, 
counterfactual answers, or graph-level predictions. Representative examples include COPA, WIQA, Corr2Cause, CLadder, CounterBench, 
ExpliCa, CausalFlip, CausalGraphBench, and CausalGraph2LLM \citep{roemmele2011copa,tandon2019wiqa,jin2024corr2cause,jin2023cladder,chen2025counterbench,miliani2025explica,wang2026causalflip,babakov2025causalgraphbench,sheth2024causalgraph2llm}. ReplaySCM differs in the target object: it scores a full executable SCM by replay under interventions.

{\looseness=-1
ReplaySCM is directly connected to classical causal discovery, interventional structure learning, Boolean network inference, 
inductive logic programming, and program synthesis. Constraint-based, score-based, interventional, functional-causal-model, 
and continuous-optimization methods estimate graphs or equivalence classes from observational and interventional 
data \citep{spirtes2000causation,chickering2002optimal,hauser2012gies,shimizu2006lingam,zheng2018notears,glymour2019review}. 
Solver-checkable symbolic induction has long been studied in Boolean network inference, logic-model inference, ILP, 
and synthesis \citep{liang1998reveal,quinlan1990learning,muggleton1991ilp,muggleton1994ilp,solarlezama2008program,alur2013syntaxguided,torlak2014lightweight}. 
For ReplaySCM, the closest prior work is exact symbolic induction from finite interventional evidence, because the benchmark rewards acyclic structure 
search together with exact Boolean mechanism synthesis under a shared semantic evaluator. 
ReplaySCM does not introduce a new causal-discovery algorithm; it formulates this discrete mechanism-induction problem as an LLM 
benchmark with a fixed final-object contract: output one executable causal mechanism and evaluate it by intervention replay.
\par}

\section{Benchmark Definition}

Each benchmark instance consists of multiple interventional worlds generated by a small binary acyclic SCM. The required output is one executable Boolean mechanism map. Credit is assigned by semantic replay on training and held-out worlds; the evaluator does not compare formula strings. Any executable SCM that induces the correct replay behavior on the scored worlds is counted as correct, even if its Boolean formulas differ syntactically from the latent gold mechanisms.

\textbf{Latent SCM and intervention worlds.} Each problem is generated from a small binary acyclic SCM with observed roots and endogenous variables. 
The benchmark provides multiple intervention worlds, each with hard intervention targets, row-level assignments, and observed rows produced 
by executing the latent SCM under that intervention. In Ordered, Block-order, and Hidden-order, a submission is an executable mechanism map 
in the benchmark Boolean DSL. In Hidden-roots, the submission must also predict the root set. In Alternative-SCM, 
the model is given a valid reference SCM and must return a semantically distinct alternative that fits the training worlds, together
with a separating intervention and witness.

\textbf{Replay and metrics.} Replay is semantic: intervened variables are clamped to their assigned values, non-intervened roots are copied from the observed row, and non-intervened endogenous variables are computed in a valid topological order of the submitted SCM. Only non-intervened endogenous cells are scored. TrainExact requires exact replay of all scored training cells. TrainWorldExact and HeldoutWorldExact average exact replay over training and held-out worlds. HeldoutExact is stricter: it requires exact training replay and exact replay of every held-out world. Appendix~\ref{app:formal_replay_metrics} gives the formal replay definitions.

\textbf{Replay example.} With one root variable $R$ and one endogenous variable $Y$ with gold mechanism $Y = \mathrm{not}\ R$, the candidate $\hat{f}_Y = \mathrm{not}\ R$ exactly replays both an observational row $(R, Y) = (0, 1)$ and an intervention row with $R := 1$ and $(R, Y) = (1, 0)$, whereas $\hat{f}_Y = R$ fails.

\textbf{Semantic structure metrics.} We use the following semantic structure metrics. A variable $U$ is a functional parent of a local
mechanism $\hat f_V$ if flipping $U$ can change the truth table of $\hat f_V$; self-loops are excluded. This yields a directed 
functional-parent graph $G(\widehat M)$ with edges $U \to V$. Parent F1 is the edge-level F1 of $G(\widehat M)$ against the gold 
functional-parent graph. Exact parent map requires every endogenous variable's functional-parent set to match the gold set. Parent SHD 
is the structural Hamming distance between the directed functional-parent graphs, with additions/deletions costing 1 and reversals also costing 1. 

\textbf{Benchmark settings and pools.} ReplaySCM uses four revealed-structure settings---Ordered, Block-order, Hidden-order, 
and Hidden-roots---and the supplementary Alternative-SCM family. Ordered (Ord-Full) and Hidden-order (Hid-Full) are the two
250-problem pools. Within these pools, 100 problems are matched with the same latent SCM in each pair. These 100 problems form the pool for matched Ordered (Ord-Match), Block-order (Block), matched Hidden-order (Hid-Match), Hidden-roots (Hid-Roots), and Alternative-SCM (Alt-Ord, Alt-Hid). The same 100 problems also form a three-level support-audit ladder: Original (Ord-Match/Hid-Match), Extra Worlds (Ord-Ext/Hid-Ext, with additional training worlds and unchanged held-out worlds), and Counterexample Audit (Ord-CEx/Hid-CEx, with further worlds that complete local predecessor-pattern coverage and add counterexamples against discovered train-consistent alternatives). The benchmark pools, sizes, and relations are listed in Appendix Table~\ref{tab:benchmark_inventory_naming}.

\vspace{-0.55\baselineskip}
\section{Benchmark Construction}

Naively sampling latent SCMs and intervention worlds yields many under-constrained problems: simple shortcut formulas can fit the observed training worlds, and some local mechanisms may never be queried on the predecessor assignments needed to determine them. ReplaySCM therefore generates latent SCMs and world sets jointly. Gold mechanisms must depend semantically on every declared parent, attain both Boolean outputs, and appear only in instances that satisfy local support, intervention coverage, distribution shift, and shortcut-resistance checks.

The generator then applies two ambiguity-reduction stages. First, a bounded survivor-reduction loop keeps a pool of shortcut candidates that fit the training worlds and adds new worlds that rule out as many as possible. Second, a targeted disambiguation stage searches for local semantic alternatives to each endogenous mechanism and adds compact worlds that rule out many alternatives at once. After generation, bounded ambiguity audits enumerate local alternatives and coordinated upstream/downstream alternative pairs under fixed search budgets. These audits quantify residual ambiguity under bounded searches; they do not prove uniqueness.

The benchmark balances finite-evidence support with structural diversity. Instances are first generated in Matched Hidden-order form and then converted into matched Ordered, Block-order, and Hidden-roots variants of the same latent SCM. Alternative-SCM is constructed from valid alternatives that fit the training worlds in paired result pools, deduplicated by semantic signature, and retained only when a single-variable separating intervention and witness exist.

{\looseness=-1
\textbf{Support-audit ladder.} The three-level support-audit ladder uses the same 100 Matched Ordered/Hidden-order SCMs to ask whether the Ordered/Hidden-order gap remains after adding more evidence. The Original level is Ord-Match/Hid-Match. The Extra Worlds level adds the same 3--4 gold-simulated training worlds per problem to both disclosure settings, excludes held-out intervention signatures, and raises mean local predecessor-pattern coverage from 0.8949 to 0.9815 while preserving the held-out worlds. The Counterexample Audit (CEx) level starts from Extra Worlds, completes local predecessor-pattern coverage with gold-simulated worlds, and then adds separating worlds until no discovered semantic alternative from LLM outputs, symbolic exact-search, or 50-seed bnlearn+DSL searches still fits the training worlds. These variants add evidence, but they also make the prompts longer. The complete generator specification is in Appendix B.2.
\par}

\section{Experimental Setup}

{\looseness=-1
All experiments use the same fixed benchmark snapshot. The two full pools are Ordered (full) and Hidden-order (full), each with 250 problems. All matched, Alternative-SCM, Extra Worlds, and CEx settings are derived from the same 100-problem same-latent subpool. Support-audit variants share latent SCMs and held-out worlds. Original and Extra Worlds also share training worlds, while CEx may add setting-specific counterexample training worlds. Every system output is parsed, checked, and replayed by the same evaluator. Appendix Table~\ref{tab:benchmark_inventory_naming} summarizes the inventory and naming convention.
\par}

\subsection{Systems and shared evaluator}

Figure~\ref{fig:evaluation_pipeline} shows the shared evaluation contract. For each instance, every system receives the same benchmark record: structured training worlds, intervention annotations, observed variables, task disclosure, allowed operators, and the required output schema. Systems differ only in candidate generation; all scored objects are executable SCMs in the benchmark DSL, with any extra fields required by supplementary settings.

We include two fixed-protocol non-LLM baselines. bnlearn+DSL uses the bnlearn structure-learning toolkit \citep{scutari2010bnlearn}
to propose candidate parent structure and then fits executable Boolean mechanisms in the benchmark DSL. The symbolic exact-search baseline searches directly for Boolean mechanisms that replay all training worlds exactly under fixed staged budgets.

{\looseness=-1
\textbf{LLM prompting.} Each LLM receives the same benchmark record rendered as a structured prompt: task metadata, variable roles, revealed structural information, DSL grammar, intervention modes, and tabular training worlds. The model is asked for one schema-compatible answer object. The evaluation uses a direct-generation protocol: no tool use, self-consistency voting, evaluator-guided revision, or semantic repair is allowed. Appendix~\ref{app:llm_evaluation_protocol} lists the model identifiers, run dates, decoding settings, response-extraction rule, stored-answer selection policy, and provider/model references, and notes that the final snapshot does not store a uniform per-item retry log. Appendix~\ref{app:prompt_excerpt} shows a prompt excerpt.
\par}

\textbf{Non-LLM baselines.} The two non-LLM calibration rows receive the same training worlds and revealed-structure fields as the LLM prompt. The symbolic exact-search baseline searches directly for train-exact Boolean DSL mechanisms under a fixed staged budget. The bnlearn+DSL baseline uses bnlearn only for structure proposal; a shared exact Boolean fitter then synthesizes executable DSL mechanisms over the proposed parents. Both baselines are scored by the same parser, legality checks, acyclicity check, and replay evaluator as the LLM submissions. These rows are benchmark-specific induction procedures, distinct from off-the-shelf end-to-end causal discovery systems. They show how much of the task remains difficult even for fixed symbolic or hybrid search pipelines under the same evaluator. Appendix~\ref{app:baseline_details} gives the fixed procedures and budgets.

We report TrainExact, TrainWorldExact, HeldoutWorldExact, and HeldoutExact to distinguish exact replay on the exposed worlds from replay under new interventions.
Coverage is the fraction of benchmark problems with a scored result. Conditional held-out metrics, reported mainly in the appendix, compute held-out replay only for responses that are train-exact. Alternative-SCM and Hidden-roots use task-specific summaries: joint success for Alternative-SCM,
and root-set exactness together with downstream mechanism induction for Hidden-roots. 
Any reported conditional rate with denominator 1–5 is suppressed and shown as *, while – indicates that the quantity is undefined because the conditioning event is empty.
Figure~\ref{fig:paired_information_tax} bootstrap intervals resample the 100 matched latent problem IDs, preserve same-latent pairing, and treat stored model outputs as fixed.

\section{Results}

We evaluate training replay, held-out replay, semantic parent structure, and held-out replay among train-exact responses (Tables 1 and 2). TrainExact and HeldoutWorldExact generally decline as structural information is withheld along the same-latent disclosure ladder (Figure 1).

\subsection{Exact executable induction remains unsolved}

On Ordered, Block-order, and Hidden-order, frontier LLMs remain well below perfect exact replay. TrainExact, HeldoutWorldExact, and HeldoutExact are all far from 1.0, with the largest difficulty on Hidden-order. Among the LLM rows, GPT-5.4 is strongest on Hidden-order, followed by Claude Opus 4.6 and Gemini 3.1 Pro.

Held-out replay is much higher among responses that replay all training worlds exactly. Responses that fail on the training worlds also tend to fail on held-out worlds. The non-LLM calibration rows in Table 1 show that many items admit high-scoring executable solutions under the same parser and replay evaluator. This gap therefore reflects the difficulty of direct executable mechanism induction, rather than a problem with the DSL or scorer.

\vspace{-0.60\baselineskip}
\subsection{Analysis of semantic structure and local mechanisms}

For frontier LLMs, schema, parsing, and executability failures are uncommon. The main errors come after parsing: models often infer some parent relationships but miss the exact parent map (Table 2).

We measure semantic parent structure with parent recall, parent F1, structural Hamming distance (SHD), per-variable parent exactness, and exact parent-map matching. A variable counts as a parent only if flipping it can change the parsed local Boolean function. On Hidden-order, GPT-5.4 achieves Parent F1 0.891 but Exact parent map only 0.280, showing that frontier models often infer informative dependency structure without matching the exact structure and local mechanisms needed for full replay. This pattern holds across the reported models: partial parent-structure inference is common, while exact parent-map and mechanism matching are much harder.

Once an exact parent map is matched, models usually produce mechanisms that are exact on the exposed worlds and on the held-out worlds, with most conditional accuracies ranging from 0.8 to 1.0 (Table 2, TrainExact and HeldoutExact | exact parent map).

% ---- inlined from generated/tables/topline_results_main.tex ----
\begin{table}[!htbp]
\centering
\scriptsize
\setlength{\tabcolsep}{2.4pt}
\resizebox{\linewidth}{!}{%
\begin{tabular}{p{0.19\linewidth}p{0.23\linewidth}rrrrrr}
\toprule
Setting & System & Valid & TrainExact & HeldoutWorld & HeldoutExact & \shortstack{HeldoutWorld\\$|$ TrainExact} & \shortstack{HeldoutExact\\$|$ TrainExact} \\
\midrule
\textbf{Ord-Full} & GPT-5.4 & 1.000 & 0.612 & \textbf{0.731} & \textbf{0.344} & 0.849 & 0.562 \\
 & Opus~4.6 & 0.980 & \textbf{0.640} & 0.725 & 0.292 & 0.816 & 0.456 \\
 & DeepSeek4Pro & 1.000 & 0.360 & 0.518 & 0.152 & 0.783 & 0.422 \\
 & Gemini3.1 & 1.000 & 0.392 & 0.640 & 0.260 & 0.885 & 0.663 \\
 & Grok~4.20 & 1.000 & 0.332 & 0.532 & 0.164 & 0.825 & 0.494 \\
 & Grok~4 & 0.980 & 0.296 & 0.519 & 0.148 & 0.818 & 0.500 \\
 & Grok~4.3 & 0.988 & 0.120 & 0.243 & 0.068 & 0.892 & 0.567 \\
 & KimiK2t & 0.972 & 0.084 & 0.164 & 0.024 & 0.786 & 0.286 \\
 & DSReasoner & 0.988 & 0.048 & 0.116 & 0.036 & \textbf{0.917} & \textbf{0.750} \\
\addlinespace[2pt]
\cmidrule(lr){2-8}
 & bnlearn+DSL & 0.996 & 0.996 & 0.880 & 0.596 & 0.880 & 0.598 \\
 & symbolic exact-search & 0.980 & 0.980 & 0.884 & 0.596 & 0.884 & 0.608 \\
\addlinespace[4pt]
\textbf{Block} & GPT-5.4 & 1.000 & \textbf{0.600} & \textbf{0.774} & \textbf{0.410} & 0.912 & 0.683 \\
 & Opus~4.6 & 0.950 & 0.420 & 0.671 & 0.240 & 0.884 & 0.571 \\
 & DeepSeek4Pro & 1.000 & 0.350 & 0.410 & 0.170 & 0.861 & 0.486 \\
 & Gemini3.1 & 1.000 & 0.280 & 0.559 & 0.220 & 0.942 & 0.786 \\
 & Grok~4.20 & 1.000 & 0.150 & 0.455 & 0.130 & 0.967 & 0.867 \\
 & Grok~4 & 0.990 & 0.240 & 0.437 & 0.070 & 0.745 & 0.292 \\
 & Grok~4.3 & 0.970 & 0.110 & 0.206 & 0.060 & 0.875 & 0.545 \\
 & KimiK2t & 0.980 & 0.010 & 0.079 & 0.010 & * & * \\
 & DSReasoner & 0.990 & 0.010 & 0.090 & 0.010 & * & * \\
\addlinespace[2pt]
\cmidrule(lr){2-8}
 & bnlearn+DSL & 1.000 & 1.000 & 0.899 & 0.620 & 0.899 & 0.620 \\
 & symbolic exact-search & 0.960 & 0.960 & 0.927 & 0.650 & 0.927 & 0.677 \\
\addlinespace[4pt]
\textbf{Hid-Full} & GPT-5.4 & 1.000 & \textbf{0.628} & \textbf{0.697} & \textbf{0.292} & 0.843 & 0.465 \\
 & Opus~4.6 & 0.972 & 0.416 & 0.595 & 0.164 & 0.827 & 0.394 \\
 & DeepSeek4Pro & 0.980 & 0.296 & 0.375 & 0.096 & 0.772 & 0.324 \\
 & Gemini3.1 & 1.000 & 0.184 & 0.431 & 0.120 & 0.891 & \textbf{0.652} \\
 & Grok~4.20 & 0.984 & 0.228 & 0.422 & 0.104 & 0.838 & 0.456 \\
 & Grok~4 & 0.976 & 0.208 & 0.370 & 0.084 & 0.815 & 0.404 \\
 & Grok~4.3 & 0.972 & 0.084 & 0.191 & 0.032 & 0.833 & 0.381 \\
 & KimiK2t & 0.984 & 0.008 & 0.049 & 0.000 & * & * \\
 & DSReasoner & 0.992 & 0.000 & 0.027 & 0.000 & – & – \\
\addlinespace[2pt]
\cmidrule(lr){2-8}
 & bnlearn+DSL & 0.996 & 0.996 & 0.896 & 0.620 & 0.896 & 0.623 \\
 & symbolic exact-search & 0.900 & 0.900 & 0.892 & 0.536 & 0.892 & 0.596 \\
\bottomrule
\end{tabular}
}
\caption{\textbf{ReplaySCM core benchmark performance.} Ord-Full and Hid-Full contain 250 problems; Block contains the matched 100-problem block-order pool. The central comparison is among LLMs; non-LLM rows calibrate task difficulty. Valid denotes executable SCM submissions. TrainExact is exact replay on all scored training cells. HeldoutWorld denotes HeldoutWorldExact, the unconditioned mean exact held-out world replay rate. HeldoutExact is strict: it requires exact training replay and exact replay of every held-out world. Here and below, * denotes a defined conditional rate suppressed because the denominator is between 1 and 5, and – denotes an undefined conditional rate with zero denominator. Boldface, where present, marks the best LLM value within the setting and column; non-LLM calibration rows are excluded from this highlighting.}
\label{tab:topline_results}
\end{table}
% ---- end inlined generated/tables/topline_results_main.tex ----

% ---- inlined from generated/tables/mechanism_failure_decomposition_main.tex ----
\begin{table}[!t]
\centering
\scriptsize
\setlength{\tabcolsep}{2.0pt}
\resizebox{\linewidth}{!}{%
\begin{tabular}{p{0.12\linewidth}p{0.19\linewidth}rrrrrrrr}
\toprule
Setting & Model & \shortstack{Parent\\Recall} & \shortstack{Parent\\F1} & \shortstack{Parent\\SHD $\downarrow$} & \shortstack{Per-var.\\parent exact} & \shortstack{Exact\\parent map} & \shortstack{Mean\\local match} & \shortstack{TrainExact\\$|$ exact\\parent map} & \shortstack{HeldoutExact\\$|$ exact\\parent map} \\
\midrule
\textbf{Ord-Full} & GPT-5.4 & 0.956 & \textbf{0.945} & \textbf{2.02} & \textbf{0.798} & \textbf{0.388} & 0.715 & 0.835 & 0.753 \\
 & Opus~4.6 & \textbf{0.958} & 0.944 & 2.12 & 0.793 & 0.335 & \textbf{0.717} & 0.854 & 0.683 \\
 & DeepSeek4Pro & 0.906 & 0.880 & 4.20 & 0.645 & 0.168 & 0.578 & 0.976 & 0.833 \\
 & Gemini3.1 & 0.914 & 0.919 & 2.87 & 0.764 & 0.332 & 0.686 & 0.807 & 0.747 \\
 & Grok~4.20 & 0.904 & 0.894 & 3.81 & 0.672 & 0.180 & 0.606 & 0.889 & 0.844 \\
 & Grok~4 & 0.917 & 0.881 & 4.58 & 0.638 & 0.147 & 0.583 & 0.917 & 0.806 \\
 & Grok~4.3 & 0.750 & 0.764 & 6.76 & 0.473 & 0.073 & 0.419 & 0.944 & 0.833 \\
 & KimiK2t & 0.746 & 0.722 & 8.46 & 0.393 & 0.029 & 0.321 & 0.857 & 0.714 \\
 & DSReasoner & 0.638 & 0.683 & 8.43 & 0.354 & 0.036 & 0.310 & \textbf{1.000} & \textbf{1.000} \\
\addlinespace[4pt]
\textbf{Block} & GPT-5.4 & \textbf{0.958} & \textbf{0.952} & \textbf{1.63} & \textbf{0.838} & \textbf{0.460} & \textbf{0.778} & 0.891 & 0.804 \\
 & Opus~4.6 & 0.927 & 0.917 & 2.81 & 0.773 & 0.316 & 0.704 & 0.800 & 0.733 \\
 & DeepSeek4Pro & 0.822 & 0.820 & 5.09 & 0.625 & 0.170 & 0.587 & 0.941 & 0.882 \\
 & Gemini3.1 & 0.878 & 0.889 & 3.46 & 0.717 & 0.260 & 0.669 & 0.846 & 0.846 \\
 & Grok~4.20 & 0.849 & 0.860 & 4.38 & 0.677 & 0.150 & 0.631 & 0.800 & 0.800 \\
 & Grok~4 & 0.837 & 0.830 & 5.25 & 0.615 & 0.121 & 0.567 & 0.667 & 0.583 \\
 & Grok~4.3 & 0.654 & 0.680 & 8.14 & 0.425 & 0.051 & 0.398 & * & * \\
 & KimiK2t & 0.637 & 0.633 & 10.16 & 0.334 & 0.020 & 0.272 & * & * \\
 & DSReasoner & 0.605 & 0.649 & 9.12 & 0.355 & 0.010 & 0.341 & * & * \\
\addlinespace[4pt]
\textbf{Hid-Full} & GPT-5.4 & \textbf{0.903} & \textbf{0.891} & \textbf{3.35} & \textbf{0.742} & \textbf{0.280} & \textbf{0.689} & 0.957 & 0.771 \\
 & Opus~4.6 & 0.852 & 0.844 & 4.75 & 0.658 & 0.189 & 0.611 & 0.913 & 0.804 \\
 & DeepSeek4Pro & 0.745 & 0.740 & 7.29 & 0.495 & 0.086 & 0.463 & 1.000 & 0.857 \\
 & Gemini3.1 & 0.786 & 0.805 & 5.39 & 0.619 & 0.156 & 0.571 & 0.795 & 0.692 \\
 & Grok~4.20 & 0.768 & 0.774 & 6.45 & 0.560 & 0.122 & 0.524 & 0.733 & 0.633 \\
 & Grok~4 & 0.770 & 0.752 & 7.25 & 0.514 & 0.066 & 0.476 & 1.000 & \textbf{0.938} \\
 & Grok~4.3 & 0.520 & 0.543 & 11.22 & 0.285 & 0.029 & 0.258 & 0.857 & 0.714 \\
 & KimiK2t & 0.457 & 0.456 & 13.53 & 0.153 & 0.000 & 0.119 & – & – \\
 & DSReasoner & 0.351 & 0.391 & 13.53 & 0.123 & 0.000 & 0.107 & – & – \\
\bottomrule
\end{tabular}
}
\caption{\textbf{Structural and local-semantic diagnostics.} Parent Recall, Parent F1, Parent SHD, Per-var. parent exact, and Exact parent map compare semantic dependency structure against the gold SCM among executable responses. Mean local match is the fraction of endogenous variables whose Boolean mechanisms match semantically. The two conditional columns are conditioned on exact parent-map matching among executable responses. The * and – notation follows Table~\ref{tab:topline_results}. Boldface, where present, marks the best LLM value within the setting and column.}
\label{tab:mechanism_failure_decomposition}
\end{table}
% ---- end inlined generated/tables/mechanism_failure_decomposition_main.tex ----

\subsection{Hiding structure drives the largest losses}

Across the structural-information ladder from Matched Ordered to Hidden-roots, TrainExact and HeldoutWorldExact generally fall as less structure is revealed (Figure 1).

The comparison uses the same latent SCMs across disclosure conditions, so score changes reflect how much structural information is revealed. The largest and most consistent drops occur on Matched Ordered→Matched Hidden-order and Matched Hidden-order→Hidden-roots; paired deltas are reported in Appendix D (Table D.1). Some individual model/metric curves are not monotone. Block-order falls between the two extremes: coarse precedence information helps, but hidden structure remains difficult.

\par\medskip\noindent\begin{minipage}{\linewidth}
\centering
\includegraphics[width=0.768\linewidth]{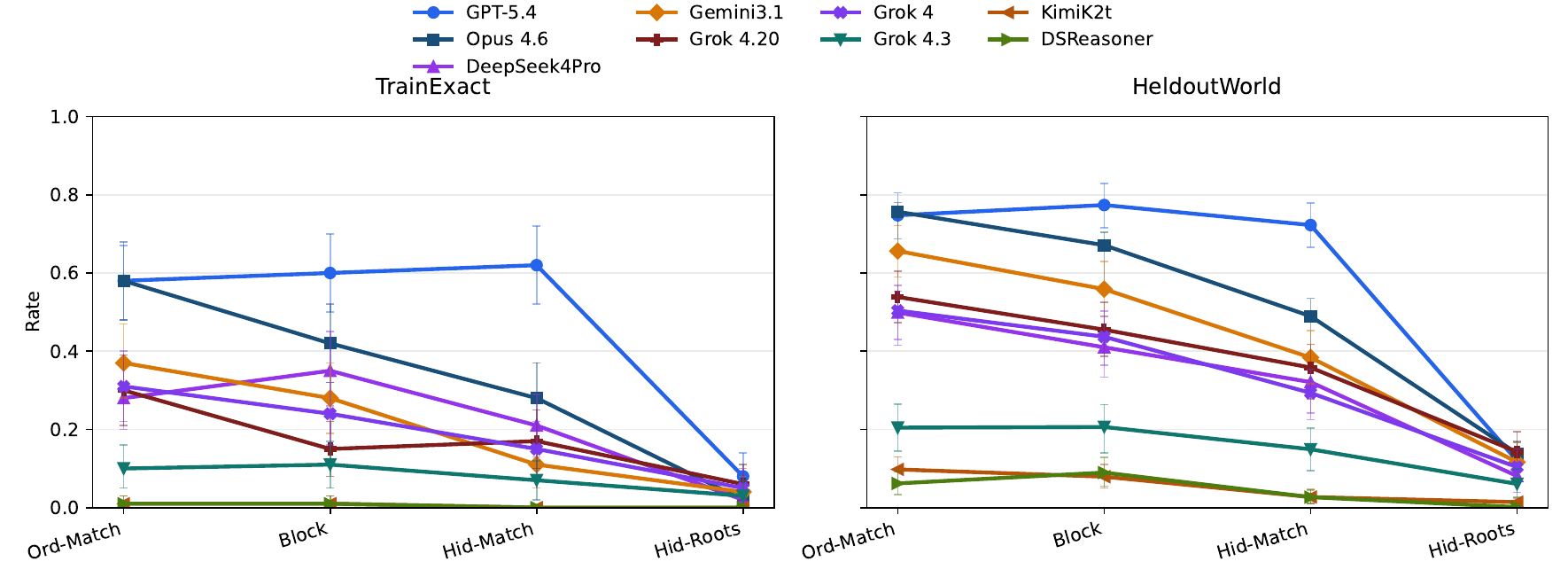}
\captionof{figure}{\textbf{Same-latent disclosure ladder for LLMs.} Left: TrainExact; right: HeldoutWorldExact, labeled HeldoutWorld in the panel title. Markers are means over the 100 matched latent SCMs; vertical bars are 95\% bootstrap intervals over matched problem IDs.}
\label{fig:paired_information_tax}
\end{minipage}\par\medskip

Alternative-SCM supplies a valid reference SCM, so the model does not have to infer the structure from the intervention worlds. The task is to construct a semantically distinct alternative that fits the training worlds, along with a separating intervention and witness. Joint success remains high under the leave-model-out restriction (Table 3). Successful alternatives typically preserve most of the supplied structure and make local edits. Alternative-SCM therefore tests local editing with a supplied SCM; broad alternative discovery is outside its scope (Appendix G).

Hidden-roots separates root-set prediction from mechanism induction. Once the root set is hidden, frontier LLMs struggle with both parts of the task. Table 3 reports root-set exactness separately from mechanism replay. Exact mechanism replay remains weak even when the root set is correct (Appendix Tables G.6--G.8).

Stronger disambiguation during benchmark generation improves held-out replay on Ordered and reduces train-to-held-out gaps on Hidden-order. The Ordered/Hidden-order gap remains in the support-audit ladder (Appendix E.2; Appendix F).

% ---- inlined from generated/tables/diagnostic_decomposition_main.tex ----
\par\noindent\begin{minipage}{\linewidth}
\centering
\scriptsize
\renewcommand{\arraystretch}{0.84}
\setlength{\tabcolsep}{2.5pt}
\resizebox{\linewidth}{!}{%
\begin{tabular}{p{0.12\linewidth}p{0.14\linewidth}rrrrr}
\toprule
\multicolumn{7}{l}{\textbf{Panel A: Alternative-SCM with supplied reference SCM}} \\
Setting & System & \shortstack{Paired\\TrainCorrect} & \shortstack{Alt-SCM\\joint} & \shortstack{Joint,\\leave-model-out} & \shortstack{Alt\\TrainExact} & \shortstack{Experiment +\\witness} \\
\midrule
\textbf{Alt-Hid} & GPT-5.4 & \textbf{0.620} & 0.840 & 0.833 & 0.840 & 0.840 \\
 & Opus~4.6 & 0.280 & 0.700 & 0.745 & 0.700 & 0.700 \\
 & DeepSeek4Pro & 0.210 & 0.670 & 0.704 & 0.680 & 0.670 \\
 & Gemini3.1 & 0.110 & \textbf{0.950} & \textbf{0.926} & \textbf{0.970} & \textbf{0.950} \\
 & Grok~4.20 & 0.170 & 0.730 & 0.757 & 0.730 & 0.730 \\
 & Grok~4 & 0.150 & 0.870 & 0.922 & 0.870 & 0.870 \\
 & Grok~4.3 & 0.070 & 0.360 & 0.408 & 0.390 & 0.360 \\
 & KimiK2t & 0.000 & 0.180 & 0.211 & 0.200 & 0.180 \\
 & DSReasoner & 0.000 & 0.030 & 0.042 & 0.030 & 0.030 \\
\addlinespace[2pt]
\textbf{Alt-Ord} & GPT-5.4 & \textbf{0.580} & \textbf{0.960} & \textbf{0.963} & \textbf{0.960} & \textbf{0.960} \\
 & Opus~4.6 & \textbf{0.580} & 0.730 & 0.765 & 0.730 & 0.730 \\
 & DeepSeek4Pro & 0.280 & 0.800 & 0.789 & 0.800 & 0.800 \\
 & Gemini3.1 & 0.370 & 0.920 & 0.941 & 0.920 & 0.920 \\
 & Grok~4.20 & 0.300 & 0.800 & 0.800 & 0.800 & 0.800 \\
 & Grok~4 & 0.310 & 0.890 & 0.922 & 0.900 & 0.890 \\
 & Grok~4.3 & 0.100 & 0.420 & 0.479 & 0.430 & 0.420 \\
 & KimiK2t & 0.010 & 0.240 & 0.296 & 0.310 & 0.240 \\
 & DSReasoner & 0.010 & 0.030 & 0.042 & 0.050 & 0.030 \\
\bottomrule
\end{tabular}
}
\resizebox{\linewidth}{!}{%
\begin{tabular}{p{0.23\linewidth}rrrrrrr}
\toprule
\multicolumn{8}{l}{\textbf{Panel B: Hid-Roots induction}} \\
System & RootExact & TrainExact & HeldoutWorld & HeldoutExact & \shortstack{TrainExact\\$|$ RootExact} & \shortstack{HeldoutWorld\\$|$ RootExact} & \shortstack{HeldoutExact\\$|$ RootExact} \\
\midrule
GPT-5.4 & 0.240 & \textbf{0.080} & 0.120 & 0.020 & \textbf{0.333} & \textbf{0.307} & 0.083 \\
Opus~4.6 & 0.240 & 0.020 & 0.138 & 0.000 & 0.083 & 0.260 & 0.000 \\
DeepSeek4Pro & 0.360 & 0.020 & 0.081 & 0.010 & 0.056 & 0.195 & 0.028 \\
Gemini3.1 & 0.340 & 0.040 & 0.116 & 0.030 & 0.118 & 0.290 & 0.088 \\
Grok~4.20 & \textbf{0.390} & 0.060 & \textbf{0.143} & \textbf{0.050} & 0.128 & 0.215 & 0.128 \\
Grok~4 & 0.300 & 0.050 & 0.104 & 0.030 & 0.133 & 0.192 & 0.100 \\
Grok~4.3 & 0.170 & 0.030 & 0.061 & 0.030 & 0.176 & 0.243 & \textbf{0.176} \\
KimiK2t & 0.060 & 0.000 & 0.014 & 0.000 & 0.000 & 0.062 & 0.000 \\
DSReasoner & 0.040 & 0.000 & 0.001 & 0.000 & * & * & * \\
bnlearn+DSL & 0.790 & 0.900 & 0.867 & 0.530 & 1.000 & 0.907 & 0.658 \\
symbolic exact-search & 0.820 & 0.950 & 0.880 & 0.560 & 1.000 & 0.931 & 0.683 \\
\bottomrule
\end{tabular}
}
\captionof{table}{\textbf{Alternative-SCM and Hidden-roots results.} Panel A compares paired-source induction with Alternative-SCM on the same matched source problems. Paired TrainCorrect is correctness on the paired source task. Alt-SCM joint requires a train-exact semantically distinct alternative together with a successful separating experiment and witness. Joint, leave-model-out excludes alternatives sourced from the evaluated model; because its eligible-case denominator can differ, this column need not be numerically below the all-source joint column. Panel B reports Hidden-roots RootExact and replay. RootExact is exact root-set prediction, and the RootExact-conditioned replay columns condition on exact root-set prediction. HeldoutWorld denotes HeldoutWorldExact. The * and – notation follows Table~\ref{tab:topline_results}. Boldface, where present, marks the best LLM value within the setting and column; non-LLM calibration rows are excluded from this highlighting.}
\label{tab:diagnostic_decomposition_main}
\end{minipage}\par
% ---- end inlined generated/tables/diagnostic_decomposition_main.tex ----

\subsection{Additional Analyses}

Appendix C breaks errors down into validity, parent structure, mechanism matching, and held-out replay. Frontier models often infer informative dependencies before they match the exact parent map or the exact local mechanisms needed for full replay. Held-out replay is much higher among train-exact responses, so many failed submissions already fail on the training worlds (Tables~\ref{tab:mechanism_validity_funnel_appendix}--\ref{tab:conditional_retention_appendix}).

{\looseness=-2
The matched support-audit ladder—Original, Extra Worlds, and CEx—asks whether ambiguity in the finite worlds explains the Ordered/Hidden-order gap. Extra Worlds raises mean local predecessor-pattern coverage from 0.8949 to 0.9815 and sharply reduces discovered alternatives that fit the training worlds. CEx completes bounded local predecessor-pattern coverage (mean 1.0, with all 100 problems fully covered), preserves the same held-out worlds, and adds counterexamples until no discovered semantic alternative from LLM outputs, symbolic exact-search, or 50-seed bnlearn+DSL searches still fits the training worlds. In the CEx rows, HeldoutExact equals TrainExact for every LLM row with TrainExact > 0, so train-exact CEx outputs also replay all held-out worlds exactly. Thus, in the most heavily audited matched setting, the main remaining difference is whether the model can find an executable SCM that fits the exposed worlds in the first place. Hidden-order nevertheless remains below Ordered on the same latent SCMs and held-out worlds despite the added evidence (Appendix E.2, Table E.1; Appendix F, Tables F.2 and F.7--F.14).
\par}

The supplementary settings show where models fail. Alternative-SCM performance remains strong, including under leave-model-out restrictions, showing that local edits are easier when a valid SCM is supplied. Hidden-roots adds a separate root-set prediction problem before mechanism induction. Together, these analyses show that inferring hidden structure is harder than producing a valid answer object or fitting isolated local mechanisms on the training worlds (Appendix G, Tables G.1--G.8).

\section{Discussion and Limitations}

ReplaySCM evaluates causal reasoning through an executable object: a reusable mechanism map that must continue to work under new interventions. This exposes failures that can remain hidden behind local answers and graph predictions. A model can produce a plausible explanation or a plausible edge set, yet still fail when its proposed mechanisms are replayed. The benchmark therefore separates two capabilities that are often conflated: describing causal structure and constructing a causal object that can be executed.

The main empirical pattern is that hidden structure is hard. Frontier LLMs often infer many functional-parent relationships, but they rarely assemble the exact parent map and Boolean mechanisms needed to replay all worlds. Alternative-SCM sharpens this point: when a valid SCM is supplied, models are much better at making local semantic edits and proposing separating interventions. Hidden-roots shows the complementary failure mode: hiding the root set adds a hard root-set prediction problem before mechanism induction even begins.

The support-audit results strengthen this interpretation. Extra Worlds and CEx add evidence, improve local predecessor-pattern coverage, and reduce discovered train-consistent alternatives. CEx reaches full bounded local predecessor-pattern coverage, and no discovered alternative from LLM outputs, symbolic exact-search, or 50-seed bnlearn+DSL searches remains train-consistent. Even under this stronger finite evidence, Hidden-order remains harder than Ordered on the same latent SCMs and held-out worlds.

The benchmark is intentionally narrow: small, finite, binary, fully observed Boolean SCMs. That restriction is what makes exact replay, solver-checkable scoring, same-latent comparisons, and bounded ambiguity audits possible. The audits rule out many easy finite-sample shortcuts within their search budgets, but finite worlds still do not establish global uniqueness.

ReplaySCM therefore provides a controlled testbed for executable causal mechanism induction. Natural next steps include noisy or partially observed SCMs, larger hidden test sets, natural-language problem statements, and interactive protocols in which models propose interventions, observe outcomes, and revise candidate mechanisms.

% ---- bibliography inlined from causal_reasoning_paper_preview.bbl ----

% ---- end bibliography ----

\appendix
\numberwithin{table}{section}
\numberwithin{figure}{section}

\section{Evaluator and protocol details}
\label{app:evaluator_details}

The appendices follow the structure of the paper. Appendix A specifies the evaluator and model-call protocol. Appendix B gives the formal replay definitions, benchmark-generation specification, and prompt excerpt. Appendix C breaks failures down into validity, structure, and replay errors. Appendix D supports the same-latent disclosure ladder. Appendix E summarizes robustness and support-audit results, with detailed construction and support-audit analyses in Appendix F. Appendix G analyzes Alternative-SCM and Hidden-roots settings, Appendix H specifies the non-LLM calibration rows, and Appendix I gives illustrative case studies.

\subsection{Benchmark inventory and evaluator contract}
\label{app:evaluator_contract}

\begin{center}
\fbox{\begin{minipage}{0.94\linewidth}
\small
\setlength{\tabcolsep}{4pt}
\renewcommand{\arraystretch}{1.08}
\begin{tabular}{@{}p{0.30\linewidth}p{0.62\linewidth}@{}}
\multicolumn{2}{@{}l@{}}{\textbf{Benchmark card.}} \\
\textbf{Primary and supplementary benchmark inventory} & See Table~\ref{tab:benchmark_inventory_naming} for benchmark names, sizes, and relations. \\
\textbf{Three-level support-audit matched subsets} & Original matched source problems, Ordered/Hidden-order + Extra Worlds, and Ordered/Hidden-order + Counterexample Audit (CEx). \\
\textbf{Study size} & 1300 scored prompt records: the two full pools plus Block, Hid-Roots, Alt-Ord/Alt-Hid, Ord-Ext/Hid-Ext, and Ord-CEx/Hid-CEx. Ord-Match and Hid-Match are subsets of the full pools, not additional records. \\
\textbf{SCM class} & Fully observed, finite, binary, acyclic SCMs with observed roots and endogenous variables; revealed structure varies by setting. \\
\textbf{Mechanism language} & Boolean DSL over not, and, or, xor, and iff; submissions are executable mechanism maps with full replay semantics. \\
\textbf{Evidence} & Multiple training and held-out intervention worlds per instance; world-level replay averages exact replay over worlds. \\
\textbf{Interventions and scoring} & Hard interventions clamp targeted variables; non-intervened endogenous cells are scored by replay. Roots provide observed per-row context unless intervened. \\
\textbf{Design emphasis} & High-precision solver-checkable evaluation, same-latent revealed-structure comparisons, and bounded ambiguity audits; large corpus scale is outside the design goal.
\end{tabular}
\end{minipage}}
\end{center}

% ---- inlined from generated/tables/benchmark_inventory.tex ----
\begin{table}[t]
\centering
\footnotesize
\setlength{\tabcolsep}{2.6pt}
\begin{tabular}{p{0.14\linewidth}p{0.20\linewidth}p{0.15\linewidth}p{0.055\linewidth}p{0.39\linewidth}}
\toprule
Pool & Full name in prose & Short label & Size & Relation to other benchmarks \\
\midrule
Full & Ordered (full) & Ord-Full & 250 & Full Ordered pool; 100 of these problems also appear in Matched Ordered \\
Full & Hidden-order (full) & Hid-Full & 250 & Full Hidden-order pool; 100 of these problems also appear in Matched Hidden-order \\
Matched source & Matched Ordered & Ord-Match & 100 & Same latent SCMs as Block, Hid-Match, and Hid-Roots; source of Alt-Ord, Ord-Ext, and Ord-CEx \\
Matched source & Matched Hidden-order & Hid-Match & 100 & Same latent SCMs as Ord-Match, Block, and Hid-Roots; source of Alt-Hid, Hid-Ext, and Hid-CEx \\
Matched derived & Block-order & Block & 100 & Same latent SCMs as Ord-Match and Hid-Match \\
Matched derived & Hidden-roots & Hid-Roots & 100 & Same latent SCMs as Ord-Match and Hid-Match \\
Supplementary & Alternative-SCM (Ordered) & Alt-Ord & 100 & Built from Matched Ordered source problems \\
Supplementary & Alternative-SCM (Hidden-order) & Alt-Hid & 100 & Built from Matched Hidden-order source problems \\
Extra Worlds & Ordered + Extra Worlds & Ord-Ext & 100 & Matched Ordered with additional training worlds; same held-out worlds \\
Extra Worlds & Hidden-order + Extra Worlds & Hid-Ext & 100 & Matched Hidden-order with additional training worlds; same held-out worlds \\
Counterexample Audit & Ordered + Counterexample Audit & Ord-CEx & 100 & Ord-Ext with further gold-simulated training worlds; same held-out worlds \\
Counterexample Audit & Hidden-order + Counterexample Audit & Hid-CEx & 100 & Hid-Ext with further gold-simulated training worlds; same held-out worlds \\
\bottomrule
\end{tabular}
\caption{\textbf{Benchmark inventory and naming convention.} Full names are used in explanatory prose; short labels are used in tables, figure axes, and compact references to matched variants. The benchmark is organized around two 250-problem full pools (\texttt{Ord-Full} and \texttt{Hid-Full}), a matched 100-problem same-latent pool (\texttt{Ord-Match}, \texttt{Block}, \texttt{Hid-Match}, and \texttt{Hid-Roots}), and derived settings built from that matched pool (\texttt{Alt-Ord} / \texttt{Alt-Hid}, \texttt{Ord-Ext} / \texttt{Hid-Ext}, and \texttt{Ord-CEx} / \texttt{Hid-CEx}). Extra Worlds uses the same added training worlds across disclosure settings; CEx preserves the same latent SCMs and held-out worlds, but counterexample training additions may be setting-specific. Here \texttt{Hid} abbreviates Hidden-order; Hidden-roots is always written \texttt{Hid-Roots}.}
\label{tab:benchmark_inventory_naming}
\end{table}
% ---- end inlined generated/tables/benchmark_inventory.tex ----

The evaluator first checks that a submission has the required schema, mentions only observed variables, assigns mechanisms to exactly the required endogenous variables, and induces an acyclic dependency graph. Each mechanism must parse in the benchmark Boolean DSL and may use only the variables allowed by the task disclosure. Valid SCM submissions are then replayed by the formal definitions in Appendix~\ref{app:formal_replay_metrics} on every training and held-out world. For the Alternative-SCM setting, the scorer additionally checks that the proposed alternative fits the training worlds, is semantically distinct from the supplied reference SCM on the bounded signature support, and is separated by the submitted single-variable intervention and witness.

\paragraph{LLM response handling.}
The prompt asks for exactly one schema-compatible answer object and nothing else. Strict one-line JSON compliance is reported separately. Replay metrics, however, are computed from the selected candidate object under the fixed extraction and selection policy in Appendix~\ref{app:llm_evaluation_protocol}; extracted mechanism strings are never rewritten, simplified, or semantically repaired.

\begin{figure}[!htbp]
\centering
\includegraphics[width=0.98\linewidth]{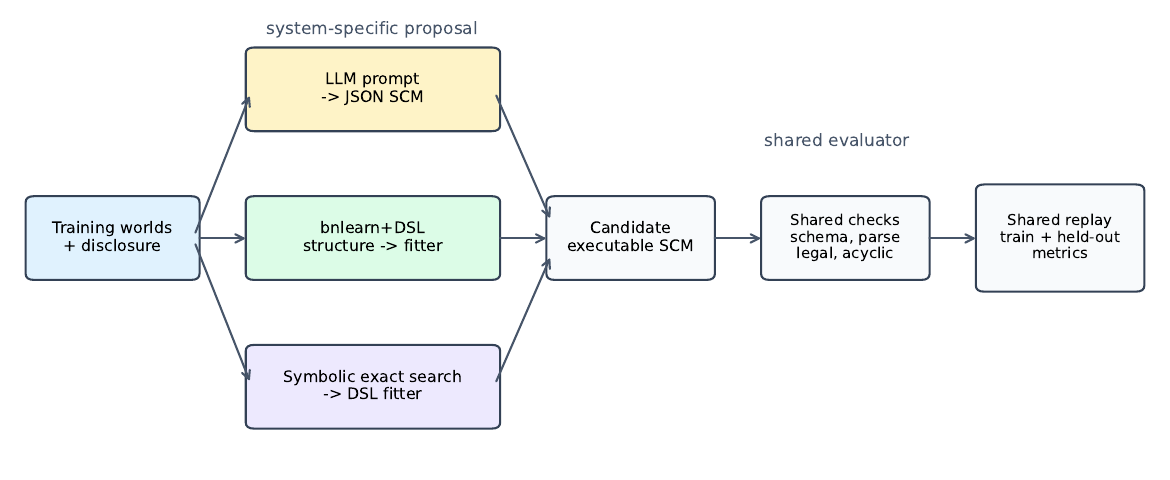}
\caption{\textbf{Shared evaluator and proposal pipeline.} Systems differ only in how they propose a candidate SCM; every candidate then passes through the same parsing, legality, acyclicity, and replay evaluator.}
\label{fig:evaluation_pipeline}
\end{figure}

\FloatBarrier
\subsection{LLM evaluation protocol}
\label{app:llm_evaluation_protocol}

All LLM rows use a direct-generation protocol. For each benchmark item, the rendered prompt is sent to the model without tool access, external computation, evaluator feedback, or iterative repair. Calls use the decoding settings in Table~\ref{tab:llm_model_protocol}. For each model/problem pair, the evaluation snapshot contains one selected response. When a pair was rerun, later calls used the same prompt and decoding settings and stopped once the stored response reached the validity criterion; however, the final snapshot does not store a uniform per-item retry count. Validation-stage rates are computed on the selected stored responses under the response-handling policy in Table~\ref{tab:llm_response_policy}. They are not one-call compliance rates. The evaluator parses the selected object and replays it exactly. It never edits mechanism strings or selects candidates using training or held-out replay scores.

\begin{table}[!htbp]
\centering
\scriptsize
\setlength{\tabcolsep}{2.4pt}
\resizebox{\linewidth}{!}{%
\begin{tabular}{@{}l p{0.25\linewidth}p{0.16\linewidth}p{0.10\linewidth}p{0.08\linewidth}p{0.31\linewidth}@{}}
\toprule
Paper label & Provider/API model identifier & Run date or date range & Temperature & Max toks & Other decoding settings \\
\midrule
GPT-5.4 & \texttt{gpt-5.4} & 2026-03-21--2026-04-26 & not set in request & 64000 & OpenAI Responses API; JSON-object response format; reasoning effort medium. \\
Opus 4.6 & \texttt{claude-opus-4-6} & 2026-04-13 & not set in request & 64000 & Thinking enabled with 32000-token budget; effort medium. \\
DeepSeek4Pro & \texttt{deepseek-v4-pro} & 2026-04-24 & 0.1 & 131072 & JSON-object response format; thinking enabled; reasoning effort medium. \\
Gemini3.1 & \texttt{gemini-3.1-pro-preview} & 2026-02-23--2026-04-26 & 0.1 & not logged & JSON output requested; thinking effort medium where exposed. \\
Grok 4.20 & \texttt{grok-4.20-0309-reasoning} & 2026-04-07--2026-04-26 & 0.1 & 64000 & JSON-object response format. \\
Grok 4 & \texttt{grok-4-0709} & 2026-02-12--2026-04-26 & 0.1 & 64000 & JSON-object response format. \\
KimiK2t & \texttt{moonshotai/kimi-k2-thinking} & 2026-02-12--2026-04-26 & 0.1 & not logged & OpenRouter call with JSON response format and reasoning enabled at high effort. \\
DSReasoner & \texttt{deepseek-reasoner} & 2026-02-08--2026-04-26 & 0.1 & not logged & JSON-object response format; provider thinking mode enabled. \\
\bottomrule
\end{tabular}
}
\caption{\textbf{LLM model identifiers and decoding settings.} Dates are taken from logged rows in the result snapshot. ``Not logged'' indicates absence from the final result records, and ``not set in request'' indicates that the provider request omitted the parameter. Provider/model references are listed immediately below the table.}
\label{tab:llm_model_protocol}
\end{table}

Provider/model references. Table~\ref{tab:llm_model_protocol} is the source of truth for the exact queried model identifiers and run windows. We cite official provider model cards or API documentation for the corresponding model families and identifiers where available: GPT-5.4 \citep{openai2026gpt54api,openai2026gpt54}, Claude Opus 4.6 \citep{anthropic2026opus46system,anthropic2026opus46}, DeepSeek V4 Pro \citep{deepseek2026v4modelcard,deepseek2026v4prohf}, Gemini 3.1 Pro \citep{googledeepmind2026gemini31card,google2026gemini31api}, Grok 4.20 and Grok 4 \citep{xai2026grok420api,xai2025grok4card,xai2026grok40709api}, Kimi K2 Thinking \citep{moonshot2026kimik2thinking,openrouter2025kimik2thinking}, and DeepSeek Reasoner \citep{deepseek2026reasonerapi,deepseek2026pricing}.

\begin{table}[!htbp]
\centering
\small
\begin{tabular}{@{}p{0.28\linewidth}p{0.65\linewidth}@{}}
\toprule
Component & Protocol \\
\midrule
Stored response & One selected response is stored for each model/problem pair. When a pair was rerun, the same prompt and decoding settings were used; the final snapshot does not store a uniform per-item retry count. \\
Sampling independence & Repeated calls, when present, are independent API calls under the same prompt and decoding settings. Outputs are not pooled, voted, or aggregated. \\
Extraction target & The extractor searches for one JSON object matching the required task schema: a mechanism map, a root set plus mechanisms, or an Alternative-SCM answer object. \\
Multiple JSON-like objects & If multiple JSON-like objects are present, the extractor chooses the task-shaped object by a deterministic schema-oriented score, then by object size, decoded span length, and earliest start position. \\
Wrapper text & Wrapper text violates the strict prompt instruction and lowers Strict JSON compliance; a deterministically extractable candidate object can still be replay-scored. \\
Candidate selection within a response & Candidate texts are examined in deterministic extractor order. The selected candidate is the first candidate that reaches executable-valid status; if none is executable-valid, the selected candidate is the best extracted answer by the extraction rule and is then counted as non-valid if it fails downstream checks. \\
No semantic repair & Mechanism strings are passed unchanged to DSL parsing and replay. \\
Missing or invalid final row & If the selected stored response is not valid and executable, the entry is counted as non-valid and receives zero on replay metrics. \\
Strict JSON diagnostic & Strict JSON measures literal compliance with the prompt's one-line JSON instruction on the selected stored response. It is reported separately from Extracted JSON and Valid. \\
Held-out isolation & Held-out worlds are excluded from prompts, extraction, stored-answer selection, and retry decisions. Selection never uses TrainExact, HeldoutWorldExact, HeldoutExact, or any held-out replay score. \\
\bottomrule
\end{tabular}
\caption{\textbf{Response extraction and final-answer policy.} The policy distinguishes literal prompt compliance from deterministic extraction of a candidate object for replay evaluation.}
\label{tab:llm_response_policy}
\end{table}

The protocol evaluates direct candidate generation from stored responses. Repeated invocations, when present in the stored snapshot, were used only to obtain one extractable or executable candidate under the response-handling rule; they were never used to search over training or held-out replay scores.

\FloatBarrier

\section{Formal replay and benchmark generation}
\label{app:formal_replay_generation}

\subsection{Formal replay definitions}
\label{app:formal_replay_metrics}

\paragraph{Latent SCM and intervention worlds.}
Let
\[
\mathcal{V}=\mathcal{R}\sqcup\mathcal{E}
\]
be the observed variables, partitioned into roots $\mathcal R$ and endogenous variables $\mathcal E$. The latent SCM is
\[
M^\star=(\mathcal{R},\mathcal{E},F^\star),
\qquad
F^\star=\{f_V^\star:V\in\mathcal{E}\},
\]
where each $f_V^\star$ is a Boolean expression in the benchmark DSL and the endogenous dependency graph is acyclic. A world $w$ specifies a hard intervention target set $I_w\subseteq\mathcal V$ and row-level assigned values $a_w^{(i)}(V)$ for $V\in I_w$. It contains rows $i\in[n_w]$, each with an observed binary assignment $x^{(w,i)}\in\{0,1\}^{\mathcal V}$ generated by executing $M^\star$ under that intervention. The problem exposes training worlds $\mathcal W_{\mathrm{tr}}$ and withholds held-out worlds $\mathcal W_{\mathrm{ho}}$.

\paragraph{Submitted objects.}
For the primary induction settings---``Ordered,'' ``Block-order,'' and ``Hidden-order''---a submission is an executable mechanism map
\[
\widehat F=\{\hat f_V:V\in\widehat{\mathcal E}\}
\]
in the restricted Boolean DSL over \texttt{not}, \texttt{and}, \texttt{or}, \texttt{xor}, and \texttt{iff}. The task disclosure determines the candidate root/endogenous partition except in ``Hidden-roots,'' where the submission must also predict the root set and hence the endogenous set. The evaluator converts the submitted object into a candidate SCM
\[
\widehat M=(\widehat{\mathcal R},\widehat{\mathcal E},\widehat F),
\]
checks schema validity, variable use, DSL parseability, and acyclicity, and then executes $\widehat M$. In the ``Alternative-SCM'' setting, the answer object is different: the system receives a valid reference SCM and must return a semantically distinct alternative SCM that fits the training worlds, a separating hard intervention, and a witness assignment.

\paragraph{Semantic replay.}
Replay executes the submitted SCM row by row. For each world-row, intervened variables are clamped to their assigned values; non-intervened roots are copied from the observed row; and non-intervened endogenous variables are computed in a valid topological order of $\widehat M$. For a valid candidate $\widehat M$, replay produces
\[
\tilde x_{\widehat M}^{(w,i)}(V)=
\begin{cases}
a_w^{(i)}(V), & V\in I_w,\\
x^{(w,i)}(V), & V\in \widehat{\mathcal R}\setminus I_w,\\
\hat f_V\!\left(\tilde x_{\widehat M}^{(w,i)}\!\restriction_{\dep(\hat f_V)}\right), & V\in \widehat{\mathcal E}\setminus I_w.
\end{cases}
\]
Only non-intervened endogenous cells are scored. For a world $w$, define the scored cells
\[
S_w(\widehat M)=\{(i,V): i\in[n_w],\ V\in\widehat{\mathcal E}\setminus I_w\}
\]
and the world replay indicator
\[
E_w(\widehat M)=
\mathbf{1}\!\left[
\forall (i,V)\in S_w(\widehat M),\
\tilde x_{\widehat M}^{(w,i)}(V)=x^{(w,i)}(V)
\right].
\]
Invalid submissions receive zero on replay metrics.

\paragraph{Replay metrics.}
Strict training exactness requires exact replay of every scored cell in every training world:
\[
\TrainExact(\widehat M)=
\mathbf{1}\!\left[
\forall w\in\mathcal W_{\mathrm{tr}},\ E_w(\widehat M)=1
\right].
\]
World-level training and held-out replay rates average exact replay over worlds:
\[
\mathrm{TrainWorldExact}(\widehat M)=
\frac{1}{|\mathcal W_{\mathrm{tr}}|}
\sum_{w\in\mathcal W_{\mathrm{tr}}}E_w(\widehat M),
\qquad
\mathrm{HeldoutWorldExact}(\widehat M)=
\frac{1}{|\mathcal W_{\mathrm{ho}}|}
\sum_{w\in\mathcal W_{\mathrm{ho}}}E_w(\widehat M).
\]
Strict held-out replay requires both exact training fit and exact replay of all held-out worlds:
\[
\HeldoutExact(\widehat M)=
\TrainExact(\widehat M)\,
\mathbf{1}\!\left[
\forall w\in\mathcal W_{\mathrm{ho}},\ E_w(\widehat M)=1
\right].
\]

\paragraph{Sanity checks.}
Invalid submissions receive zero on replay metrics. The aggregate tables must therefore satisfy $\TrainExact \leq \mathrm{TrainWorldExact}$ and $\HeldoutExact \leq \TrainExact$. When $\HeldoutExact \mid \TrainExact$ is reported and not suppressed, $\HeldoutExact$ equals $\TrainExact$ times $\HeldoutExact \mid \TrainExact$ up to rounding. $\mathrm{HeldoutWorldExact}$ can exceed $\TrainExact$ because it is an unconditioned per-world held-out average. Conditional rates with denominators 1--5 are suppressed as ``*''; zero-denominator cases are shown as ``--''.

Retention is the ratio $\mathrm{HeldoutWorldExact}/\mathrm{TrainWorldExact}$ when the denominator is nonzero. Conditional held-out metrics compute held-out replay only when $\TrainExact=1$. For the primary induction settings, TrainExact asks whether one executable SCM explains the exposed worlds exactly. HeldoutWorldExact and HeldoutExact are the stricter mechanistic criteria, because they ask whether that same executable object survives new interventions.

\subsection{Benchmark generation specification}
\label{app:benchmark_generation_spec}

ReplaySCM instances are generated programmatically. The generator first samples a small acyclic Boolean SCM, then simulates observational and interventional worlds from that SCM. Candidate instances are rejected or strengthened until they satisfy support, intervention-coverage, shortcut-resistance, and bounded ambiguity checks. These procedures reduce obvious finite-sample shortcuts while keeping the task finite: the evidence still does not guarantee identification.

\begin{table}[!htbp]
\centering
\small
\begin{tabular}{@{}p{0.25\linewidth}p{0.68\linewidth}@{}}
\toprule
Component & Specification \\
\midrule
Variables & 6--10 observed binary variables; 3 roots; 3--7 endogenous variables. \\
Variable labels & Observed labels $X1,\ldots,Xn$ are randomly permuted over latent slots, so numeric suffixes contain no topological-order information. \\
Topological order & Roots precede endogenous variables in the latent order; endogenous variables are sampled in an acyclic order. \\
Parent sets & For each endogenous variable, candidate parents are earlier latent variables, bounded by MaxPredecessorsPerMechanism. Parent count is sampled uniformly from the admissible range, with a lower bound of two whenever at least two predecessors are available. \\
Predecessor bound & MaxPredecessorsPerMechanism is in $\{2,3,4,5\}$ in the released snapshot; most records use 4. \\
Operators & \texttt{not}, \texttt{and}, \texttt{or}, \texttt{xor}, \texttt{iff}. \\
Constants & Constants are not allowed in gold mechanisms or submissions. \\
Mechanism sampling & Gold mechanisms are sampled as Boolean DSL expressions over declared parents, canonicalized, and rejected unless every declared parent is semantically active and the local truth table attains both Boolean outputs. \\
Mechanism size & Gold mechanisms in the released snapshot have AST sizes 3--14 and depths 2--6. \\
Rows per ordinary world & 10--12 unit rows per ordinary generated world. \\
Root contexts & Each unit/root pair has a latent threshold sampled uniformly from $[0,1]$. Each world/root has an environment level in $\{0.2,0.35,0.5,0.65,0.8\}$; the non-intervened root value is $1[\mathrm{threshold}<\mathrm{level}]$. \\
Training worlds & Initial target of 8 training worlds; final core records contain 8--11 after disambiguation additions. \\
Held-out worlds & 8 held-out worlds per core record. \\
Intervention modes & \texttt{none}, \texttt{hard\_constant}, \texttt{hard\_assigned}. \\
Intervention targets & Roots and endogenous variables may both be intervention targets; multiple targets are allowed. Target sizes are usually 1--3, with size 0 for observational worlds and occasional size 4 in forced-focus worlds. \\
\texttt{hard\_constant} & Each target is clamped to one uniformly sampled Boolean value for the whole world. \\
\texttt{hard\_assigned} & Each target receives row-level Boolean assignments sampled with bias in $\{0.3,0.5,0.7\}$; all-equal multi-row assignments are repaired to include both values. \\
Held-out separation & Held-out intervention signatures are excluded from training-world and added-world selection. \\
\bottomrule
\end{tabular}
\caption{\textbf{Latent SCM and intervention-world sampling.}}
\label{tab:scm_world_sampling_spec}
\end{table}

\begin{table}[!htbp]
\centering
\small
\begin{tabular}{@{}p{0.25\linewidth}p{0.68\linewidth}@{}}
\toprule
Stage & Criterion or budget \\
\midrule
Local support & Counts assignments to bounded predecessor subsets on rows where the target variable is not intervened. The local-support probes use subset size 3 for nearly all records and size 4 for a small number of records. \\
Scored exposure & Where stricter thresholds apply, each endogenous variable must have 3--4 scored training worlds and 33/40/44 scored cells, depending on the generation stratum. \\
Intervention coverage & Where stricter thresholds apply, records require 3--4 \texttt{hard\_assigned} worlds, 1--2 \texttt{hard\_constant} worlds, and at most 4--5 intervened worlds per endogenous variable. \\
Held-out balance & Held-out target novelty is constrained to lie between lower bounds 0.20--0.25 and upper bounds 0.65--0.72, depending on the stratum. \\
Shortcut class & Bounded Boolean DSL formulas over admissible predecessor sets; AST cap 5, with floor 2--3. \\
Survivor reduction & Maintains shortcut candidates that fit the training worlds; proposes 170--340 candidate worlds; uses 8--17 iterations/restarts; requires survivor reduction fraction 0.75--0.95. \\
Targeted local disambiguation & Searches local alternatives with AST slack 2, max cap 8, 50,000 states per size, and 2.5 seconds per variable; accepted records contain 0--3 compact added training worlds. \\
Stronger post-generation local audit & AST slack 4, cap 10, 80,000 states per size, 4 seconds per variable. \\
Coordinated pair audit & AST slack 3, cap 9, up to 5 upstream alternatives per variable, and 120 seconds per problem. \\
Added-worlds selector & For the 100 matched source problems, adds 3--4 additional gold-simulated training worlds per problem, mean 3.09, selected to maximize new local predecessor-pattern coverage while excluding held-out signatures. \\
Counterexample Audit selector & Starts from the Extra Worlds records, completes local predecessor-pattern coverage to 1.0, and adds separating worlds until no discovered alternative from LLM outputs, symbolic exact-search, or 50-seed bnlearn+DSL audit pools still fits the training worlds. \\
Alternative-SCM retention & Uses discovered valid train-exact alternatives; deduplicates by semantic signature; retains only cases with a single-variable separating intervention and witness. \\
\bottomrule
\end{tabular}
\caption{\textbf{Construction filters and bounded audit budgets.}}
\label{tab:construction_filter_budget_spec}
\end{table}

\subsubsection{Latent SCM sampling}

Each latent SCM has binary observed variables $X1,\ldots,Xn$ with $n\in\{6,\ldots,10\}$. Exactly three latent slots are roots, and all remaining variables are endogenous. Visible labels are randomly permuted over the latent slots, so label order contains no causal-order information.

Parent sets are sampled only from earlier latent variables, which enforces acyclicity by construction. The predecessor window is bounded by MaxPredecessorsPerMechanism, and the parent count is sampled uniformly from the admissible range, using a lower bound of two whenever at least two predecessors are available. Local mechanisms are sampled as Boolean DSL expressions over the declared parents using \texttt{not}, \texttt{and}, \texttt{or}, \texttt{xor}, and \texttt{iff}; constants are not allowed. A mechanism is rejected if any declared parent is semantically inactive or if its truth table is constant.

\par\smallskip\noindent\textbf{Algorithm 1: Sample a latent SCM}
\begin{Verbatim}[fontsize=\footnotesize,breaklines=true]
Input: size stratum, predecessor bound, operator set
1. Sample n in {6,...,10}; create n latent slots.
2. Mark the first three latent slots as roots and the remaining slots as endogenous.
3. Randomly permute visible labels X1,...,Xn over the latent slots.
4. For each endogenous variable V in latent order:
   a. Let C(V) be the bounded set of earlier latent variables.
   b. Sample a parent count uniformly from the admissible range.
   c. Sample that many parents uniformly from C(V).
   d. Sample a Boolean DSL expression over those parents.
   e. Reject and resample unless every declared parent is semantically active
      and the truth table is nonconstant.
5. Return the acyclic Boolean SCM.
\end{Verbatim}

\subsubsection{Intervention-world construction}

A world is a table of unit rows under one intervention. A unit is a row identifier with latent root thresholds. Ordinary generated worlds contain 10--12 rows. Unit IDs share latent root thresholds across worlds, but non-intervened root values can change across worlds because the world-level environment changes. Specifically, each unit/root threshold is sampled uniformly from $[0,1]$, each world/root environment level lies in $\{0.2,0.35,0.5,0.65,0.8\}$, and the non-intervened root value is $1[\mathrm{threshold}<\mathrm{level}]$.

The intervention mode \texttt{none} leaves all variables unaltered. In \texttt{hard\_constant}, each target variable is clamped to one uniformly sampled Boolean value for the entire world. In \texttt{hard\_assigned}, each target receives row-level Boolean assignments sampled with bias in $\{0.3,0.5,0.7\}$; all-equal multi-row assignments are repaired to include both Boolean values. Roots and endogenous variables can both be targets, and a world may intervene on multiple variables.

Simulation clamps intervened variables first. Non-intervened roots are generated from unit thresholds and world environments. Endogenous variables are then evaluated in latent topological order, unless they are themselves intervention targets. The generator starts from an initial target of eight training worlds; final core records contain 8--11 training worlds after disambiguation additions, plus 8 held-out worlds.

\par\smallskip\noindent\textbf{Algorithm 2: Simulate an intervention world}
\begin{Verbatim}[fontsize=\footnotesize,breaklines=true]
Input: latent SCM M, units U, world intervention I
For each unit u in U:
1. For each root R:
   if R is intervened on, set R to its intervention value;
   otherwise set R from the unit threshold and world environment.
2. For each endogenous variable V in latent order:
   if V is intervened on, set V to its intervention value;
   otherwise evaluate f_V on the already assigned parent values.
3. Record the complete row over observed variables.
Return the world table and intervention metadata.
\end{Verbatim}

Training worlds are selected and then strengthened by the filters in Section~\ref{app:generation_filters}. Eight held-out worlds are simulated from the same latent SCM under intervention signatures withheld from the prompt. These held-out worlds are withheld from the model and used only for evaluation.

\subsubsection{Filters and bounded ambiguity reduction}
\label{app:generation_filters}

\textbf{Local support and scored exposure.} For an endogenous variable $V$, a local predecessor pattern is an assignment to a bounded subset of variables that precede $V$ in the latent order. Patterns are counted only on rows where $V$ is not intervened. The local-support probes use subset size 3 for nearly all records and subset size 4 for a small number of records. In stricter generation strata, each endogenous variable must also have 3--4 scored training worlds and 33/40/44 scored cells.

\textbf{Intervention coverage and held-out balance.} The intervention-coverage checks require enough direct intervention variation without making variables nearly always clamped. In stricter strata, records require 3--4 \texttt{hard\_assigned} worlds, 1--2 \texttt{hard\_constant} worlds, and at most 4--5 intervened worlds per endogenous variable. Held-out target novelty is constrained between lower bounds 0.20--0.25 and upper bounds 0.65--0.72, depending on the generation stratum.

\textbf{Shortcut resistance.} A shortcut is a bounded Boolean DSL formula over an admissible predecessor set that fits the training rows while differing from the intended local mechanism. The shortcut class uses AST cap 5 and floor 2--3. Survivor reduction maintains a set of shortcut formulas that fit the training worlds, proposes 170--340 candidate worlds, and uses 8--17 iterations or restarts to add worlds that remove survivors. Accepted records must reduce the survivor pool by a fraction between 0.75 and 0.95, depending on the stratum.

\textbf{Targeted disambiguation.} A local semantic alternative is a Boolean DSL mechanism that is consistent with the observed rows but has a different local truth table from the gold mechanism. The targeted search uses AST slack 2, maximum cap 8, 50,000 states per size, and 2.5 seconds per variable. When compact intervention worlds separate many such alternatives, they are added to the training set; accepted records contain 0--3 such added worlds.

\textbf{Post-generation audits and interpretation.} Stronger audits measure remaining ambiguity under larger bounded searches. The stronger local audit uses AST slack 4, cap 10, 80,000 states per size, and 4 seconds per variable. The coordinated pair audit uses AST slack 3, cap 9, up to 5 upstream alternatives per variable, and 120 seconds per problem. These searches do not establish uniqueness.

\par\smallskip\noindent\textbf{Algorithm 3: Accept or strengthen a candidate instance}
\begin{Verbatim}[fontsize=\footnotesize,breaklines=true]
Input: latent SCM M, initial training worlds W
1. Reject M if any mechanism has inactive parents or a constant truth table.
2. Simulate candidate training worlds from M.
3. Check local support, scored exposure, intervention coverage, and held-out balance.
4. Enumerate bounded shortcut formulas that fit W.
5. While shortcut survivors remain above threshold and the world-addition budget remains:
   a. propose candidate intervention worlds;
   b. simulate each candidate from M;
   c. add the world that rules out the most surviving shortcuts.
6. Enumerate bounded local semantic alternatives.
7. Add compact separating worlds when they rule out many alternatives.
8. Run bounded local and coordinated ambiguity audits.
9. Accept the instance if all checks pass; otherwise reject or retry.
\end{Verbatim}

\subsubsection{Benchmark variants}

Ordered reveals the root/endogenous partition and a full topological order, so mechanisms may reference only earlier variables. Block-order reveals the roots and coarse precedence blocks; within-block order is hidden, but submitted dependencies must remain compatible with block precedence and acyclicity. Hidden-order reveals the root/endogenous partition but hides the endogenous order, so the evaluator must infer whether the submitted dependency graph is acyclic before replay.

Hidden-roots hides the root set and requires the model to predict it. In Hidden-roots, the predicted root set determines which variables require mechanisms; root-set exactness is therefore reported separately from mechanism replay. An incorrect root set can still fit some finite observed worlds, so root-set prediction and downstream replay are separated in the results.

Matched pools share latent SCMs and worlds, changing only disclosed structure and output schema. The 100-problem matched same-latent subset yields Matched Ordered, Block-order, Matched Hidden-order, and Hidden-roots variants. The full Ordered and Hidden-order pools each contain 250 problems, with the matched pool serving as a controlled same-latent subset.

\subsubsection{Support-audit and Alternative-SCM derived settings}

\textbf{Support-audit variants.} The Extra Worlds level starts from the 100 matched Ordered/Hidden-order pairs. The same added worlds are used for Ordered + Extra Worlds and Hidden-order + Extra Worlds, and the held-out worlds are unchanged. Candidate additions are observational worlds, single-variable \texttt{hard\_constant} worlds, and single-variable \texttt{hard\_assigned} worlds, with held-out intervention signatures excluded. The selector adds 3--4 worlds per problem, mean 3.09. Mean local predecessor-pattern coverage rises from 0.8949 to 0.9815, and full local coverage rises from 2/100 to 42/100. The Counterexample Audit (CEx) level starts from these Extra Worlds records and adds further gold-simulated worlds until mean local predecessor-pattern coverage is 1.0. It then tests every semantically distinct LLM or symbolic exact-search answer found on the Extra Worlds records that fits the training worlds; if the answer still fits after the coverage additions, CEx adds a separating world. A 50-seed bnlearn+DSL sweep is then run for each problem and disclosure setting, and semantically distinct bnlearn+DSL alternatives that fit the training worlds are separated in the same way. After these additions, no discovered alternative in the audited pools still fits the training worlds. Because discovered alternatives can differ by disclosure, CEx may add setting-specific counterexample worlds while preserving the same latent SCMs and held-out worlds. The added worlds increase both the evidence supplied to the model and the prompt length, so CEx does not isolate prompt length from evidence.

\textbf{Alternative-SCM.} Alternative-SCM candidates are valid train-exact alternatives discovered in paired result pools, with sources including both LLMs and non-LLM systems. Semantic signatures deduplicate syntactic rewrites by recording effective parents and local truth-table behavior. Retained items must be semantically distinct from the supplied reference SCM and admit a single-variable separating intervention with a witness. Leave-model-out analyses exclude alternatives sourced from the evaluated model. Because Alternative-SCM is built from discovered alternatives that fit the training worlds, it tests local editing from a supplied SCM. It does not provide an independent proof of latent-SCM non-identifiability.

\subsubsection{Snapshot and artifact}

All experiments use one frozen benchmark snapshot. The released artifact will include the generator, configuration manifest, fixed benchmark records, rendered prompts, evaluator, and audit summaries. The manifest records the exact pool names, seeds, and configuration files needed to reproduce the snapshot. If a hidden-evaluation split is used, the public artifact will distinguish public prompt records from evaluator-only held-out data.

The filters and audits use bounded searches. Passing them means that no alternative was found within the specified candidate classes and budgets, and that the released worlds satisfy the stated support and shortcut-resistance criteria. ReplaySCM therefore evaluates executable replay from finite interventional evidence; it does not guarantee identification of a unique latent SCM.

\subsection{Prompt excerpt}
\label{app:prompt_excerpt}
\label{app:rendered_prompts}

All systems receive a structured prompt record containing task metadata, the DSL grammar, intervention-world tables, and the required JSON schema. The excerpt below shows the common input contract for an Ordered item. Other task variants preserve the same serialization style while changing the disclosed structural fields and output schema. The released artifact contains the full rendered prompts for every benchmark item.

\paragraph{System instruction excerpt.}
\begin{Verbatim}[fontsize=\footnotesize]
You are solving a formal causal mechanism induction task
over finite interventional worlds.
Treat the input as machine-checkable structure.
Output exactly one JSON object matching the required schema.
\end{Verbatim}

\paragraph{Task metadata excerpt.}
\begin{Verbatim}[fontsize=\footnotesize]
Task: CIND_A_SCM
ObservedVariables: X1, X2, X3, X4, X5
RootVariables: X1, X2
EndogenousVariables: X3, X4, X5
TopologicalOrder: X1, X2, X3, X4, X5
AllowedOperators: not, and, or, xor, iff
InterventionModes: none, hard_constant, hard_assigned
\end{Verbatim}

\paragraph{Scoring excerpt.}
For each world and row, intervened variables are clamped to their intervention values, non-intervened roots are copied from the row, and non-intervened endogenous variables are evaluated in topological order from the submitted mechanisms. Only non-intervened endogenous cells are scored. The same mechanism map is replayed on all training and held-out worlds.

\paragraph{DSL excerpt.}
\begin{Verbatim}[fontsize=\footnotesize]
expr ::= VAR
       | (not expr)
       | (and expr expr ...)
       | (or expr expr ...)
       | (xor expr expr ...)
       | (iff expr expr ...)
Constants are disallowed. Operators and variable names must match the metadata exactly.
\end{Verbatim}

\paragraph{Training-world excerpt.}
\begin{Verbatim}[fontsize=\footnotesize]
WorldId: train_00
InterventionMode: none
InterventionTargetsAssigned: []
InterventionTargetsConstant: {}
Rows:
- u00: X1=0 X2=1 X3=1 X4=0 X5=1
- u01: X1=1 X2=0 X3=0 X4=1 X5=0

WorldId: train_01
InterventionMode: hard_constant
InterventionTargetsConstant: {"X3": 1}
Rows:
- u00: X1=0 X2=1 X3=1 X4=1 X5=0
- u01: X1=1 X2=0 X3=1 X4=0 X5=1
\end{Verbatim}

\paragraph{Output schema excerpt.}
\begin{Verbatim}[fontsize=\footnotesize]
{"mechanisms":{"X3":"...","X4":"...","X5":"..."}}
\end{Verbatim}

\par\smallskip\noindent
The literal prompt also includes validity conditions, a formatting-only example, and the complete set of training worlds for the instance.

\FloatBarrier

\section{Failure analysis support}
\label{app:failure_decomposition_support}

Appendix C separates failures in validity, semantic parent structure, and held-out replay among train-exact responses. Across Tables C.1--C.4, frontier models often infer some dependencies before producing a fully correct executable mechanism.

% ---- inlined from generated/tables/mechanism_validity_funnel.tex ----
{\scriptsize
\setlength{\tabcolsep}{1.8pt}
\begin{longtable}{@{}p{0.12\linewidth}p{0.16\linewidth}*{8}{>{\raggedleft\arraybackslash}p{0.078\linewidth}}@{}}
\caption{\textbf{Validation-stage breakdown.} The columns separate successive evaluator stages before final executable validity. Mechanism strings are never semantically repaired. Strict JSON, Extracted JSON, and the later validation stages follow the response-handling policy in Appendix~\ref{app:llm_evaluation_protocol}. Appendix~\ref{app:llm_evaluation_protocol} reports selected stored responses instead of retry-normalized one-call attempts, so Strict JSON and Valid should be interpreted under that stored-response policy.}\label{tab:mechanism_validity_funnel_appendix}\\
\toprule
Setting & Model & \shortstack{Strict\\JSON} & \shortstack{Extracted\\JSON} & Schema & Keys & Parse & Legal & Acyclic & Valid \\
\midrule
\endfirsthead
\toprule
Setting & Model & \shortstack{Strict\\JSON} & \shortstack{Extracted\\JSON} & Schema & Keys & Parse & Legal & Acyclic & Valid \\
\midrule
\endhead
\bottomrule
\endlastfoot
Ord-Full & GPT-5.4 & 1.000 & 1.000 & 1.000 & 1.000 & 1.000 & 1.000 & 1.000 & 1.000 \\
Ord-Full & Opus~4.6 & 0.944 & 0.980 & 0.980 & 0.980 & 0.980 & 0.980 & 0.980 & 0.980 \\
Ord-Full & DeepSeek4Pro & 0.720 & 1.000 & 1.000 & 1.000 & 1.000 & 1.000 & 1.000 & 1.000 \\
Ord-Full & Gemini3.1 & 1.000 & 1.000 & 1.000 & 1.000 & 1.000 & 1.000 & 1.000 & 1.000 \\
Ord-Full & Grok~4.20 & 1.000 & 1.000 & 1.000 & 1.000 & 1.000 & 1.000 & 1.000 & 1.000 \\
Ord-Full & Grok~4 & 0.984 & 0.984 & 0.984 & 0.984 & 0.980 & 0.980 & 0.980 & 0.980 \\
Ord-Full & Grok~4.3 & 0.988 & 0.988 & 0.988 & 0.988 & 0.988 & 0.988 & 0.988 & 0.988 \\
Ord-Full & KimiK2t & 0.052 & 0.976 & 0.976 & 0.976 & 0.976 & 0.972 & 0.972 & 0.972 \\
Ord-Full & DSReasoner & 0.152 & 0.988 & 0.988 & 0.988 & 0.988 & 0.988 & 0.988 & 0.988 \\
\addlinespace
Block & GPT-5.4 & 1.000 & 1.000 & 1.000 & 1.000 & 1.000 & 1.000 & 1.000 & 1.000 \\
Block & Opus~4.6 & 0.600 & 0.950 & 0.950 & 0.950 & 0.950 & 0.950 & 0.950 & 0.950 \\
Block & DeepSeek4Pro & 0.440 & 1.000 & 1.000 & 1.000 & 1.000 & 1.000 & 1.000 & 1.000 \\
Block & Gemini3.1 & 1.000 & 1.000 & 1.000 & 1.000 & 1.000 & 1.000 & 1.000 & 1.000 \\
Block & Grok~4.20 & 1.000 & 1.000 & 1.000 & 1.000 & 1.000 & 1.000 & 1.000 & 1.000 \\
Block & Grok~4 & 1.000 & 1.000 & 1.000 & 1.000 & 1.000 & 1.000 & 0.990 & 0.990 \\
Block & Grok~4.3 & 0.990 & 0.990 & 0.990 & 0.990 & 0.990 & 0.990 & 0.970 & 0.970 \\
Block & KimiK2t & 0.050 & 0.990 & 0.990 & 0.990 & 0.990 & 0.990 & 0.980 & 0.980 \\
Block & DSReasoner & 0.160 & 1.000 & 1.000 & 1.000 & 1.000 & 1.000 & 0.990 & 0.990 \\
\addlinespace
Hid-Full & GPT-5.4 & 0.992 & 1.000 & 1.000 & 1.000 & 1.000 & 1.000 & 1.000 & 1.000 \\
Hid-Full & Opus~4.6 & 0.760 & 0.972 & 0.972 & 0.972 & 0.972 & 0.972 & 0.972 & 0.972 \\
Hid-Full & DeepSeek4Pro & 0.788 & 1.000 & 1.000 & 1.000 & 1.000 & 1.000 & 0.980 & 0.980 \\
Hid-Full & Gemini3.1 & 1.000 & 1.000 & 1.000 & 1.000 & 1.000 & 1.000 & 1.000 & 1.000 \\
Hid-Full & Grok~4.20 & 1.000 & 1.000 & 1.000 & 1.000 & 1.000 & 1.000 & 0.984 & 0.984 \\
Hid-Full & Grok~4 & 0.996 & 0.996 & 0.996 & 0.996 & 0.996 & 0.992 & 0.976 & 0.976 \\
Hid-Full & Grok~4.3 & 0.984 & 0.984 & 0.984 & 0.984 & 0.984 & 0.984 & 0.972 & 0.972 \\
Hid-Full & KimiK2t & 0.072 & 0.992 & 0.992 & 0.992 & 0.992 & 0.992 & 0.984 & 0.984 \\
Hid-Full & DSReasoner & 0.116 & 1.000 & 1.000 & 1.000 & 1.000 & 1.000 & 0.992 & 0.992 \\
\addlinespace
Hid-Roots & GPT-5.4 & 1.000 & 1.000 & 1.000 & 1.000 & 1.000 & 1.000 & 0.990 & 0.990 \\
Hid-Roots & Opus~4.6 & 0.590 & 0.930 & 0.930 & 0.930 & 0.930 & 0.930 & 0.930 & 0.930 \\
Hid-Roots & DeepSeek4Pro & 0.730 & 0.990 & 0.990 & 0.990 & 0.990 & 0.990 & 0.960 & 0.960 \\
Hid-Roots & Gemini3.1 & 1.000 & 1.000 & 1.000 & 1.000 & 1.000 & 1.000 & 1.000 & 1.000 \\
Hid-Roots & Grok~4.20 & 1.000 & 1.000 & 1.000 & 1.000 & 1.000 & 1.000 & 0.990 & 0.990 \\
Hid-Roots & Grok~4 & 1.000 & 1.000 & 1.000 & 1.000 & 1.000 & 1.000 & 1.000 & 1.000 \\
Hid-Roots & Grok~4.3 & 0.990 & 0.990 & 0.990 & 0.990 & 0.990 & 0.990 & 0.970 & 0.970 \\
Hid-Roots & KimiK2t & 0.020 & 0.990 & 0.990 & 0.990 & 0.990 & 0.990 & 0.980 & 0.980 \\
Hid-Roots & DSReasoner & 0.160 & 1.000 & 1.000 & 1.000 & 1.000 & 1.000 & 1.000 & 1.000 \\
\addlinespace
Alt-Ord & GPT-5.4 & 1.000 & 1.000 & 1.000 & 1.000 & 1.000 & 1.000 & 1.000 & 1.000 \\
Alt-Ord & Opus~4.6 & 1.000 & 1.000 & 1.000 & 1.000 & 1.000 & 1.000 & 1.000 & 1.000 \\
Alt-Ord & DeepSeek4Pro & 0.170 & 1.000 & 1.000 & 1.000 & 1.000 & 1.000 & 1.000 & 1.000 \\
Alt-Ord & Gemini3.1 & 1.000 & 1.000 & 1.000 & 1.000 & 1.000 & 1.000 & 1.000 & 1.000 \\
Alt-Ord & Grok~4.20 & 1.000 & 1.000 & 1.000 & 1.000 & 1.000 & 1.000 & 1.000 & 1.000 \\
Alt-Ord & Grok~4 & 1.000 & 1.000 & 1.000 & 1.000 & 1.000 & 1.000 & 1.000 & 1.000 \\
Alt-Ord & Grok~4.3 & 1.000 & 1.000 & 1.000 & 1.000 & 1.000 & 1.000 & 1.000 & 1.000 \\
Alt-Ord & KimiK2t & 0.050 & 1.000 & 1.000 & 1.000 & 1.000 & 1.000 & 1.000 & 1.000 \\
Alt-Ord & DSReasoner & 0.190 & 1.000 & 1.000 & 1.000 & 1.000 & 1.000 & 1.000 & 1.000 \\
\addlinespace
Alt-Hid & GPT-5.4 & 1.000 & 1.000 & 1.000 & 1.000 & 1.000 & 1.000 & 1.000 & 1.000 \\
Alt-Hid & Opus~4.6 & 0.990 & 1.000 & 0.990 & 0.990 & 0.990 & 0.990 & 0.990 & 0.990 \\
Alt-Hid & DeepSeek4Pro & 0.210 & 1.000 & 1.000 & 1.000 & 1.000 & 1.000 & 1.000 & 1.000 \\
Alt-Hid & Gemini3.1 & 1.000 & 1.000 & 1.000 & 1.000 & 1.000 & 1.000 & 1.000 & 1.000 \\
Alt-Hid & Grok~4.20 & 1.000 & 1.000 & 1.000 & 1.000 & 1.000 & 1.000 & 1.000 & 1.000 \\
Alt-Hid & Grok~4 & 1.000 & 1.000 & 1.000 & 1.000 & 1.000 & 1.000 & 1.000 & 1.000 \\
Alt-Hid & Grok~4.3 & 1.000 & 1.000 & 1.000 & 1.000 & 1.000 & 1.000 & 1.000 & 1.000 \\
Alt-Hid & KimiK2t & 0.010 & 0.980 & 0.980 & 0.980 & 0.980 & 0.980 & 0.980 & 0.980 \\
Alt-Hid & DSReasoner & 0.160 & 1.000 & 1.000 & 1.000 & 1.000 & 1.000 & 1.000 & 1.000 \\
\addlinespace
Ord-Ext & GPT-5.4 & 1.000 & 1.000 & 1.000 & 1.000 & 1.000 & 1.000 & 1.000 & 1.000 \\
Ord-Ext & Opus~4.6 & 0.950 & 0.980 & 0.980 & 0.980 & 0.980 & 0.980 & 0.980 & 0.980 \\
Ord-Ext & DeepSeek4Pro & 1.000 & 1.000 & 1.000 & 1.000 & 1.000 & 1.000 & 1.000 & 1.000 \\
Ord-Ext & Gemini3.1 & 1.000 & 1.000 & 1.000 & 1.000 & 1.000 & 1.000 & 1.000 & 1.000 \\
Ord-Ext & Grok~4.20 & 1.000 & 1.000 & 1.000 & 1.000 & 1.000 & 1.000 & 1.000 & 1.000 \\
Ord-Ext & Grok~4 & 1.000 & 1.000 & 1.000 & 1.000 & 1.000 & 0.990 & 0.990 & 0.990 \\
Ord-Ext & Grok~4.3 & 1.000 & 1.000 & 1.000 & 1.000 & 1.000 & 1.000 & 1.000 & 1.000 \\
Ord-Ext & KimiK2t & 0.040 & 0.950 & 0.950 & 0.950 & 0.950 & 0.950 & 0.950 & 0.950 \\
Ord-Ext & DSReasoner & 0.070 & 1.000 & 1.000 & 1.000 & 1.000 & 1.000 & 1.000 & 1.000 \\
\addlinespace
Hid-Ext & GPT-5.4 & 1.000 & 1.000 & 1.000 & 1.000 & 1.000 & 1.000 & 0.990 & 0.990 \\
Hid-Ext & Opus~4.6 & 0.730 & 0.950 & 0.950 & 0.950 & 0.950 & 0.950 & 0.950 & 0.950 \\
Hid-Ext & DeepSeek4Pro & 1.000 & 1.000 & 1.000 & 1.000 & 1.000 & 1.000 & 1.000 & 1.000 \\
Hid-Ext & Gemini3.1 & 1.000 & 1.000 & 1.000 & 1.000 & 1.000 & 1.000 & 1.000 & 1.000 \\
Hid-Ext & Grok~4.20 & 1.000 & 1.000 & 1.000 & 1.000 & 1.000 & 1.000 & 0.970 & 0.970 \\
Hid-Ext & Grok~4 & 1.000 & 1.000 & 1.000 & 1.000 & 0.990 & 0.990 & 0.990 & 0.990 \\
Hid-Ext & Grok~4.3 & 0.980 & 0.980 & 0.980 & 0.980 & 0.980 & 0.980 & 0.980 & 0.980 \\
Hid-Ext & KimiK2t & 0.020 & 0.990 & 0.990 & 0.990 & 0.990 & 0.990 & 0.970 & 0.970 \\
Hid-Ext & DSReasoner & 0.100 & 1.000 & 1.000 & 1.000 & 1.000 & 1.000 & 0.990 & 0.990 \\
\addlinespace
Ord-CEx & GPT-5.4 & 1.000 & 1.000 & 1.000 & 1.000 & 1.000 & 1.000 & 1.000 & 1.000 \\
Ord-CEx & Opus~4.6 & 0.940 & 0.990 & 0.990 & 0.990 & 0.990 & 0.990 & 0.990 & 0.990 \\
Ord-CEx & DeepSeek4Pro & 0.600 & 1.000 & 1.000 & 1.000 & 1.000 & 1.000 & 1.000 & 1.000 \\
Ord-CEx & Gemini3.1 & 1.000 & 1.000 & 1.000 & 1.000 & 1.000 & 1.000 & 1.000 & 1.000 \\
Ord-CEx & Grok~4.20 & 1.000 & 1.000 & 1.000 & 1.000 & 1.000 & 1.000 & 1.000 & 1.000 \\
Ord-CEx & Grok~4 & 0.990 & 0.990 & 0.990 & 0.990 & 0.990 & 0.990 & 0.990 & 0.990 \\
Ord-CEx & Grok~4.3 & 0.980 & 0.980 & 0.980 & 0.980 & 0.980 & 0.980 & 0.980 & 0.980 \\
Ord-CEx & KimiK2t & 0.080 & 0.970 & 0.970 & 0.970 & 0.970 & 0.960 & 0.960 & 0.960 \\
Ord-CEx & DSReasoner & 0.160 & 1.000 & 1.000 & 1.000 & 1.000 & 1.000 & 1.000 & 1.000 \\
\addlinespace
Hid-CEx & GPT-5.4 & 1.000 & 1.000 & 1.000 & 1.000 & 1.000 & 1.000 & 1.000 & 1.000 \\
Hid-CEx & Opus~4.6 & 0.670 & 0.950 & 0.950 & 0.950 & 0.950 & 0.950 & 0.950 & 0.950 \\
Hid-CEx & DeepSeek4Pro & 0.800 & 1.000 & 1.000 & 1.000 & 1.000 & 1.000 & 1.000 & 1.000 \\
Hid-CEx & Gemini3.1 & 1.000 & 1.000 & 1.000 & 1.000 & 1.000 & 1.000 & 1.000 & 1.000 \\
Hid-CEx & Grok~4.20 & 1.000 & 1.000 & 1.000 & 1.000 & 1.000 & 1.000 & 1.000 & 1.000 \\
Hid-CEx & Grok~4 & 0.950 & 0.950 & 0.950 & 0.950 & 0.940 & 0.940 & 0.930 & 0.930 \\
Hid-CEx & Grok~4.3 & 0.990 & 0.990 & 0.990 & 0.990 & 0.990 & 0.990 & 0.980 & 0.980 \\
Hid-CEx & KimiK2t & 0.060 & 1.000 & 1.000 & 1.000 & 1.000 & 1.000 & 0.980 & 0.980 \\
Hid-CEx & DSReasoner & 0.350 & 1.000 & 1.000 & 1.000 & 1.000 & 1.000 & 0.990 & 0.990 \\
\end{longtable}
}
% ---- end inlined generated/tables/mechanism_validity_funnel.tex ----

For the strongest frontier models, failures at the wrapper-text or schema stages are already uncommon. Deterministic extraction absorbs most wrapper variation without altering mechanism formulas, so the consequential losses arise later in the evaluator, after a candidate executable object has already been formed (Table~\ref{tab:mechanism_validity_funnel_appendix}).

% ---- inlined from generated/tables/effective_parent_recovery_appendix.tex ----
{\scriptsize
\setlength{\tabcolsep}{1.2pt}
\begin{longtable}{@{}p{0.105\linewidth}p{0.080\linewidth}>{\raggedleft\arraybackslash}p{0.055\linewidth}*{7}{>{\raggedleft\arraybackslash}p{0.102\linewidth}}@{}}
\caption{\textbf{Semantic parent-structure details.} Parent metrics ignore formula spelling and depend only on semantic functional-parent structure. For Hidden-roots, structural scoring is conditioned on exact root-set prediction, so the $n$ column counts root-exact executable submissions; for other benchmarks it counts executable submissions. This is why Hidden-roots structural counts can be much lower than Valid in Table~\ref{tab:structural_shd_core_appendix}.}\label{tab:effective_parent_structure_appendix}\\
\toprule
Setting & Model & \shortstack{$n$ structure\\scored} & \shortstack{Recall\\$|$ scored} & \shortstack{Parent\\F1\\$|$ scored} & \shortstack{Per-var.\\parent\\exact $|$ scored} & \shortstack{Parent\\map exact\\$|$ scored} & \shortstack{Mean\\local match\\$|$ scored} & \shortstack{Full\\local match\\$|$ scored} & \shortstack{Train\\Exact\\$|$ map\\exact} \\
\midrule
\endfirsthead
\toprule
Setting & Model & \shortstack{$n$ structure\\scored} & \shortstack{Recall\\$|$ scored} & \shortstack{Parent\\F1\\$|$ scored} & \shortstack{Per-var.\\parent\\exact $|$ scored} & \shortstack{Parent\\map exact\\$|$ scored} & \shortstack{Mean\\local match\\$|$ scored} & \shortstack{Full\\local match\\$|$ scored} & \shortstack{Train\\Exact\\$|$ map\\exact} \\
\midrule
\endhead
Ord-Full & GPT-5.4 & 250 & 0.956 & \textbf{0.945} & \textbf{0.798} & \textbf{0.388} & 0.715 & \textbf{0.272} & 0.835 \\
Ord-Full & Opus~4.6 & 245 & \textbf{0.958} & 0.944 & 0.793 & 0.335 & \textbf{0.717} & 0.220 & 0.854 \\
Ord-Full & DeepSeek4Pro & 250 & 0.906 & 0.880 & 0.645 & 0.168 & 0.578 & 0.136 & 0.976 \\
Ord-Full & Gemini3.1 & 250 & 0.914 & 0.919 & 0.764 & 0.332 & 0.686 & 0.236 & 0.807 \\
Ord-Full & Grok~4.20 & 250 & 0.904 & 0.894 & 0.672 & 0.180 & 0.606 & 0.144 & 0.889 \\
Ord-Full & Grok~4 & 245 & 0.917 & 0.881 & 0.638 & 0.147 & 0.583 & 0.118 & 0.917 \\
Ord-Full & Grok~4.3 & 247 & 0.750 & 0.764 & 0.473 & 0.073 & 0.419 & 0.057 & 0.944 \\
Ord-Full & KimiK2t & 243 & 0.746 & 0.722 & 0.393 & 0.029 & 0.321 & 0.021 & 0.857 \\
Ord-Full & DSReasoner & 247 & 0.638 & 0.683 & 0.354 & 0.036 & 0.310 & 0.036 & \textbf{1.000} \\
\addlinespace
Block & GPT-5.4 & 100 & \textbf{0.958} & \textbf{0.952} & \textbf{0.838} & \textbf{0.460} & \textbf{0.778} & \textbf{0.360} & 0.891 \\
Block & Opus~4.6 & 95 & 0.927 & 0.917 & 0.773 & 0.316 & 0.704 & 0.221 & 0.800 \\
Block & DeepSeek4Pro & 100 & 0.822 & 0.820 & 0.625 & 0.170 & 0.587 & 0.150 & 0.941 \\
Block & Gemini3.1 & 100 & 0.878 & 0.889 & 0.717 & 0.260 & 0.669 & 0.220 & 0.846 \\
Block & Grok~4.20 & 100 & 0.849 & 0.860 & 0.677 & 0.150 & 0.631 & 0.120 & 0.800 \\
Block & Grok~4 & 99 & 0.837 & 0.830 & 0.615 & 0.121 & 0.567 & 0.071 & 0.667 \\
Block & Grok~4.3 & 97 & 0.654 & 0.680 & 0.425 & 0.051 & 0.398 & 0.051 & * \\
Block & KimiK2t & 98 & 0.637 & 0.633 & 0.334 & 0.020 & 0.272 & 0.010 & * \\
Block & DSReasoner & 99 & 0.605 & 0.649 & 0.355 & 0.010 & 0.341 & 0.010 & * \\
\addlinespace
Hid-Full & GPT-5.4 & 250 & \textbf{0.903} & \textbf{0.891} & \textbf{0.742} & \textbf{0.280} & \textbf{0.689} & \textbf{0.192} & 0.957 \\
Hid-Full & Opus~4.6 & 243 & 0.852 & 0.844 & 0.658 & 0.189 & 0.611 & 0.123 & 0.913 \\
Hid-Full & DeepSeek4Pro & 245 & 0.745 & 0.740 & 0.495 & 0.086 & 0.463 & 0.065 & \textbf{1.000} \\
Hid-Full & Gemini3.1 & 250 & 0.786 & 0.805 & 0.619 & 0.156 & 0.571 & 0.100 & 0.795 \\
Hid-Full & Grok~4.20 & 246 & 0.768 & 0.774 & 0.560 & 0.122 & 0.524 & 0.069 & 0.733 \\
Hid-Full & Grok~4 & 244 & 0.770 & 0.752 & 0.514 & 0.066 & 0.476 & 0.049 & \textbf{1.000} \\
Hid-Full & Grok~4.3 & 243 & 0.520 & 0.543 & 0.285 & 0.029 & 0.258 & 0.012 & 0.857 \\
Hid-Full & KimiK2t & 246 & 0.457 & 0.456 & 0.153 & 0.000 & 0.119 & 0.000 & – \\
Hid-Full & DSReasoner & 248 & 0.351 & 0.391 & 0.123 & 0.000 & 0.107 & 0.000 & – \\
\addlinespace
Hid-Roots & GPT-5.4 & 24 & \textbf{0.676} & \textbf{0.690} & \textbf{0.494} & 0.083 & \textbf{0.469} & 0.083 & * \\
Hid-Roots & Opus~4.6 & 24 & 0.651 & 0.653 & 0.440 & 0.000 & 0.394 & 0.000 & – \\
Hid-Roots & DeepSeek4Pro & 34 & 0.573 & 0.589 & 0.388 & 0.029 & 0.362 & 0.029 & * \\
Hid-Roots & Gemini3.1 & 34 & 0.632 & 0.661 & 0.479 & 0.118 & 0.458 & 0.088 & * \\
Hid-Roots & Grok~4.20 & 39 & 0.630 & 0.657 & 0.434 & 0.128 & 0.420 & 0.128 & * \\
Hid-Roots & Grok~4 & 30 & 0.562 & 0.565 & 0.316 & 0.100 & 0.316 & 0.100 & * \\
Hid-Roots & Grok~4.3 & 17 & 0.490 & 0.503 & 0.302 & \textbf{0.176} & 0.269 & \textbf{0.176} & * \\
Hid-Roots & KimiK2t & 6 & 0.329 & 0.346 & 0.167 & 0.000 & 0.133 & 0.000 & – \\
Hid-Roots & DSReasoner & 4 & * & * & * & * & * & * & – \\
\addlinespace
Ord-Ext & GPT-5.4 & 100 & \textbf{0.979} & \textbf{0.975} & \textbf{0.903} & \textbf{0.610} & \textbf{0.866} & \textbf{0.500} & 0.836 \\
Ord-Ext & Opus~4.6 & 98 & 0.953 & 0.944 & 0.838 & 0.418 & 0.785 & 0.306 & 0.756 \\
Ord-Ext & DeepSeek4Pro & 100 & 0.898 & 0.873 & 0.644 & 0.110 & 0.619 & 0.080 & 0.727 \\
Ord-Ext & Gemini3.1 & 100 & 0.915 & 0.923 & 0.784 & 0.360 & 0.746 & 0.290 & 0.806 \\
Ord-Ext & Grok~4.20 & 100 & 0.901 & 0.898 & 0.737 & 0.270 & 0.704 & 0.200 & 0.741 \\
Ord-Ext & Grok~4 & 99 & 0.855 & 0.839 & 0.612 & 0.162 & 0.580 & 0.111 & 0.688 \\
Ord-Ext & Grok~4.3 & 100 & 0.682 & 0.714 & 0.416 & 0.020 & 0.388 & 0.020 & * \\
Ord-Ext & KimiK2t & 95 & 0.665 & 0.675 & 0.393 & 0.021 & 0.341 & 0.021 & * \\
Ord-Ext & DSReasoner & 100 & 0.641 & 0.686 & 0.376 & 0.020 & 0.347 & 0.020 & * \\
\addlinespace
Hid-Ext & GPT-5.4 & 99 & \textbf{0.918} & \textbf{0.920} & \textbf{0.793} & \textbf{0.313} & \textbf{0.774} & \textbf{0.263} & 0.839 \\
Hid-Ext & Opus~4.6 & 95 & 0.871 & 0.874 & 0.708 & 0.242 & 0.682 & 0.210 & 0.870 \\
Hid-Ext & DeepSeek4Pro & 100 & 0.719 & 0.728 & 0.531 & 0.120 & 0.529 & 0.120 & \textbf{1.000} \\
Hid-Ext & Gemini3.1 & 100 & 0.820 & 0.846 & 0.662 & 0.190 & 0.629 & 0.160 & 0.842 \\
Hid-Ext & Grok~4.20 & 97 & 0.774 & 0.790 & 0.598 & 0.134 & 0.577 & 0.103 & 0.769 \\
Hid-Ext & Grok~4 & 99 & 0.708 & 0.717 & 0.509 & 0.091 & 0.491 & 0.081 & 0.889 \\
Hid-Ext & Grok~4.3 & 96 & 0.512 & 0.541 & 0.298 & 0.052 & 0.286 & 0.052 & * \\
Hid-Ext & KimiK2t & 97 & 0.466 & 0.463 & 0.161 & 0.000 & 0.133 & 0.000 & – \\
Hid-Ext & DSReasoner & 99 & 0.373 & 0.403 & 0.155 & 0.000 & 0.142 & 0.000 & – \\
\bottomrule
\end{longtable}
}
% ---- end inlined generated/tables/effective_parent_recovery_appendix.tex ----

Semantic parent-structure estimates are much stronger than exact parent-map or local-semantic matching. This gap is especially clear on Hidden-order, where edge-level dependency estimates remain informative while exact structural and mechanism matching are much lower (Table~\ref{tab:effective_parent_structure_appendix}).

% ---- inlined from generated/tables/structural_shd_core_selected.tex ----
{\scriptsize
\setlength{\tabcolsep}{1.2pt}
\begin{longtable}{@{}p{0.105\linewidth}p{0.080\linewidth}*{8}{>{\raggedleft\arraybackslash}p{0.098\linewidth}}@{}}
\caption{\textbf{Structural-distance view of parent structure.} Valid is the fraction of benchmark items for which the model's selected final answer is an executable SCM: schema-compatible, parseable in the DSL, legal under the disclosure rules, and acyclic. The denominator is all benchmark items for that model and benchmark under the selected stored-response policy in Appendix~\ref{app:llm_evaluation_protocol}; missing or non-scorable final answers count as non-valid. Parent SHD is directed semantic functional-parent structural Hamming distance; lower values are better. For Hidden-roots, parent and local-semantic columns are additionally conditioned on exact root-set prediction.}\label{tab:structural_shd_core_appendix}\\
\toprule
Setting & Model & Valid & \shortstack{Parent\\F1\\$|$ scored} & \shortstack{Parent\\SHD $\downarrow$\\$|$ scored} & \shortstack{Parent\\map exact\\$|$ scored} & \shortstack{Mean\\local match\\$|$ scored} & \shortstack{Full\\local match\\$|$ scored} & \shortstack{Train\\Exact} & \shortstack{Heldout\\World} \\
\midrule
\endfirsthead
\toprule
Setting & Model & Valid & \shortstack{Parent\\F1\\$|$ scored} & \shortstack{Parent\\SHD $\downarrow$\\$|$ scored} & \shortstack{Parent\\map exact\\$|$ scored} & \shortstack{Mean\\local match\\$|$ scored} & \shortstack{Full\\local match\\$|$ scored} & \shortstack{Train\\Exact} & \shortstack{Heldout\\World} \\
\midrule
\endhead
Block & GPT-5.4 & 1.000 & \textbf{0.952} & \textbf{1.63} & \textbf{0.460} & \textbf{0.778} & \textbf{0.360} & \textbf{0.600} & \textbf{0.774} \\
Block & Opus~4.6 & 0.950 & 0.917 & 2.81 & 0.316 & 0.704 & 0.221 & 0.420 & 0.671 \\
Block & DeepSeek4Pro & 1.000 & 0.820 & 5.09 & 0.170 & 0.587 & 0.150 & 0.350 & 0.410 \\
Block & Gemini3.1 & 1.000 & 0.889 & 3.46 & 0.260 & 0.669 & 0.220 & 0.280 & 0.559 \\
Block & Grok~4.20 & 1.000 & 0.860 & 4.38 & 0.150 & 0.631 & 0.120 & 0.150 & 0.455 \\
Block & Grok~4 & 0.990 & 0.830 & 5.25 & 0.121 & 0.567 & 0.071 & 0.240 & 0.437 \\
Block & Grok~4.3 & 0.950 & 0.680 & 8.14 & 0.053 & 0.399 & 0.053 & 0.110 & 0.210 \\
Block & KimiK2t & 0.980 & 0.633 & 10.16 & 0.020 & 0.272 & 0.010 & 0.010 & 0.079 \\
Block & DSReasoner & 0.990 & 0.649 & 9.12 & 0.010 & 0.341 & 0.010 & 0.010 & 0.090 \\
\addlinespace
Hid-Roots & GPT-5.4 & 0.990 & \textbf{0.690} & \textbf{7.83} & 0.083 & \textbf{0.469} & 0.083 & \textbf{0.080} & 0.120 \\
Hid-Roots & Opus~4.6 & 0.930 & 0.653 & 9.21 & 0.000 & 0.394 & 0.000 & 0.020 & 0.138 \\
Hid-Roots & DeepSeek4Pro & 0.960 & 0.589 & 10.50 & 0.029 & 0.362 & 0.029 & 0.020 & 0.081 \\
Hid-Roots & Gemini3.1 & 1.000 & 0.661 & 8.44 & 0.118 & 0.458 & 0.088 & 0.040 & 0.116 \\
Hid-Roots & Grok~4.20 & 0.990 & 0.657 & 9.28 & 0.128 & 0.420 & 0.128 & 0.060 & \textbf{0.143} \\
Hid-Roots & Grok~4 & 1.000 & 0.565 & 12.60 & 0.100 & 0.316 & 0.100 & 0.050 & 0.104 \\
Hid-Roots & Grok~4.3 & 0.960 & 0.503 & 11.47 & \textbf{0.176} & 0.269 & \textbf{0.176} & 0.030 & 0.060 \\
Hid-Roots & KimiK2t & 0.980 & 0.346 & 15.67 & 0.000 & 0.133 & 0.000 & 0.000 & 0.014 \\
Hid-Roots & DSReasoner & 1.000 & * & * & * & * & * & 0.000 & 0.001 \\
\bottomrule
\end{longtable}
}
% ---- end inlined generated/tables/structural_shd_core_selected.tex ----

The same contrast appears in structural-distance form. Parent SHD remains far lower than chance-level structure would suggest, even when exact parent-map matching is still weak. Frontier models therefore often infer substantial dependency information without producing the full exact executable mechanism (Table~\ref{tab:structural_shd_core_appendix}).

To avoid overinterpreting unstable conditional estimates, the appendix suppresses any defined rate whose denominator is between 1 and 5 inclusive and marks it with *; only zero-denominator cases are shown as –.

% ---- inlined from generated/tables/conditional_retention_selected.tex ----
{\scriptsize
\setlength{\tabcolsep}{1.4pt}
\begin{longtable}{@{}p{0.14\linewidth}p{0.14\linewidth}*{5}{>{\raggedleft\arraybackslash}p{0.113\linewidth}}@{}}
\caption{\textbf{Held-out replay among train-exact responses.} These conditional rates distinguish submissions that fail to fit all training worlds from held-out replay errors among responses that are already TrainExact.}\label{tab:conditional_retention_appendix}\\
\toprule
Setting & Model & TrainExact & HeldoutWorld & \shortstack{HeldoutWorld\\$|$ TrainExact} & \shortstack{HeldoutExact\\$|$ TrainExact} & \shortstack{$n$ train-exact\\solutions} \\
\midrule
\endfirsthead
\toprule
Setting & Model & TrainExact & HeldoutWorld & \shortstack{HeldoutWorld\\$|$ TrainExact} & \shortstack{HeldoutExact\\$|$ TrainExact} & \shortstack{$n$ train-exact\\solutions} \\
\midrule
\endhead
\bottomrule
\endlastfoot
Ord-Full & GPT-5.4 & 0.612 & \textbf{0.731} & 0.849 & 0.562 & 153 \\
Ord-Full & Opus~4.6 & \textbf{0.640} & 0.725 & 0.816 & 0.456 & 160 \\
Ord-Full & DeepSeek4Pro & 0.360 & 0.518 & 0.783 & 0.422 & 90 \\
Ord-Full & Gemini3.1 & 0.392 & 0.640 & 0.885 & 0.663 & 98 \\
Ord-Full & Grok~4.20 & 0.332 & 0.532 & 0.825 & 0.494 & 83 \\
Ord-Full & Grok~4 & 0.296 & 0.519 & 0.818 & 0.500 & 74 \\
Ord-Full & Grok~4.3 & 0.120 & 0.243 & 0.892 & 0.567 & 30 \\
Ord-Full & KimiK2t & 0.084 & 0.164 & 0.786 & 0.286 & 21 \\
Ord-Full & DSReasoner & 0.048 & 0.116 & \textbf{0.917} & \textbf{0.750} & 12 \\
\addlinespace
Hid-Full & GPT-5.4 & \textbf{0.628} & \textbf{0.697} & 0.843 & 0.465 & 157 \\
Hid-Full & Opus~4.6 & 0.416 & 0.595 & 0.827 & 0.394 & 104 \\
Hid-Full & DeepSeek4Pro & 0.296 & 0.375 & 0.772 & 0.324 & 74 \\
Hid-Full & Gemini3.1 & 0.184 & 0.431 & \textbf{0.891} & \textbf{0.652} & 46 \\
Hid-Full & Grok~4.20 & 0.228 & 0.422 & 0.838 & 0.456 & 57 \\
Hid-Full & Grok~4 & 0.208 & 0.370 & 0.815 & 0.404 & 52 \\
Hid-Full & Grok~4.3 & 0.084 & 0.191 & 0.833 & 0.381 & 21 \\
Hid-Full & KimiK2t & 0.008 & 0.049 & * & * & 2 \\
Hid-Full & DSReasoner & 0.000 & 0.027 & – & – & 0 \\
\addlinespace
Ord-Match & GPT-5.4 & \textbf{0.580} & 0.748 & 0.909 & 0.690 & 58 \\
Ord-Match & Opus~4.6 & \textbf{0.580} & \textbf{0.757} & 0.888 & 0.621 & 58 \\
Ord-Match & DeepSeek4Pro & 0.280 & 0.499 & 0.888 & 0.571 & 28 \\
Ord-Match & Gemini3.1 & 0.370 & 0.656 & 0.943 & 0.784 & 37 \\
Ord-Match & Grok~4.20 & 0.300 & 0.539 & 0.879 & 0.600 & 30 \\
Ord-Match & Grok~4 & 0.310 & 0.504 & 0.831 & 0.484 & 31 \\
Ord-Match & Grok~4.3 & 0.100 & 0.207 & 0.900 & 0.600 & 10 \\
Ord-Match & KimiK2t & 0.010 & 0.098 & * & * & 1 \\
Ord-Match & DSReasoner & 0.010 & 0.062 & * & * & 1 \\
\addlinespace
Hid-Match & GPT-5.4 & \textbf{0.620} & \textbf{0.723} & 0.891 & 0.500 & 62 \\
Hid-Match & Opus~4.6 & 0.280 & 0.489 & 0.879 & 0.536 & 28 \\
Hid-Match & DeepSeek4Pro & 0.210 & 0.321 & 0.815 & 0.429 & 21 \\
Hid-Match & Gemini3.1 & 0.110 & 0.384 & \textbf{1.000} & \textbf{1.000} & 11 \\
Hid-Match & Grok~4.20 & 0.170 & 0.358 & 0.897 & 0.706 & 17 \\
Hid-Match & Grok~4 & 0.150 & 0.293 & 0.808 & 0.400 & 15 \\
Hid-Match & Grok~4.3 & 0.070 & 0.168 & 0.839 & 0.571 & 7 \\
Hid-Match & KimiK2t & 0.000 & 0.027 & – & – & 0 \\
Hid-Match & DSReasoner & 0.000 & 0.027 & – & – & 0 \\
\addlinespace
Block & GPT-5.4 & \textbf{0.600} & \textbf{0.774} & 0.912 & 0.683 & 60 \\
Block & Opus~4.6 & 0.420 & 0.671 & 0.884 & 0.571 & 42 \\
Block & DeepSeek4Pro & 0.350 & 0.410 & 0.861 & 0.486 & 35 \\
Block & Gemini3.1 & 0.280 & 0.559 & 0.942 & 0.786 & 28 \\
Block & Grok~4.20 & 0.150 & 0.455 & 0.967 & 0.867 & 15 \\
Block & Grok~4 & 0.240 & 0.437 & 0.745 & 0.292 & 24 \\
Block & Grok~4.3 & 0.110 & 0.206 & 0.875 & 0.545 & 11 \\
Block & KimiK2t & 0.010 & 0.079 & * & * & 1 \\
Block & DSReasoner & 0.010 & 0.090 & * & * & 1 \\
\addlinespace
Hid-Roots & GPT-5.4 & \textbf{0.080} & 0.120 & 0.734 & 0.250 & 8 \\
Hid-Roots & Opus~4.6 & 0.020 & 0.138 & * & * & 2 \\
Hid-Roots & DeepSeek4Pro & 0.020 & 0.081 & * & * & 2 \\
Hid-Roots & Gemini3.1 & 0.040 & 0.116 & * & * & 4 \\
Hid-Roots & Grok~4.20 & 0.060 & \textbf{0.143} & 0.917 & 0.833 & 6 \\
Hid-Roots & Grok~4 & 0.050 & 0.104 & * & * & 5 \\
Hid-Roots & Grok~4.3 & 0.030 & 0.061 & * & * & 3 \\
Hid-Roots & KimiK2t & 0.000 & 0.014 & – & – & 0 \\
Hid-Roots & DSReasoner & 0.000 & 0.001 & – & – & 0 \\
\addlinespace
Ord-Ext & GPT-5.4 & \textbf{0.630} & \textbf{0.880} & 0.958 & 0.841 & 63 \\
Ord-Ext & Opus~4.6 & 0.380 & 0.783 & 0.977 & 0.895 & 38 \\
Ord-Ext & DeepSeek4Pro & 0.140 & 0.465 & 0.938 & 0.714 & 14 \\
Ord-Ext & Gemini3.1 & 0.300 & 0.709 & 0.992 & 0.967 & 30 \\
Ord-Ext & Grok~4.20 & 0.230 & 0.571 & 0.967 & 0.913 & 23 \\
Ord-Ext & Grok~4 & 0.120 & 0.460 & 1.000 & 1.000 & 12 \\
Ord-Ext & Grok~4.3 & 0.020 & 0.146 & * & * & 2 \\
Ord-Ext & KimiK2t & 0.020 & 0.066 & * & * & 2 \\
Ord-Ext & DSReasoner & 0.020 & 0.100 & * & * & 2 \\
\addlinespace
Hid-Ext & GPT-5.4 & \textbf{0.430} & \textbf{0.783} & 0.962 & 0.767 & 43 \\
Hid-Ext & Opus~4.6 & 0.270 & 0.670 & 0.972 & 0.852 & 27 \\
Hid-Ext & DeepSeek4Pro & 0.170 & 0.343 & 0.941 & 0.824 & 17 \\
Hid-Ext & Gemini3.1 & 0.180 & 0.507 & 0.979 & 0.889 & 18 \\
Hid-Ext & Grok~4.20 & 0.130 & 0.466 & 0.971 & 0.846 & 13 \\
Hid-Ext & Grok~4 & 0.090 & 0.335 & 0.972 & 0.889 & 9 \\
Hid-Ext & Grok~4.3 & 0.050 & 0.137 & * & * & 5 \\
Hid-Ext & KimiK2t & 0.000 & 0.032 & – & – & 0 \\
Hid-Ext & DSReasoner & 0.000 & 0.034 & – & – & 0 \\
\end{longtable}
}
% ---- end inlined generated/tables/conditional_retention_selected.tex ----

Held-out replay is much stronger among train-exact responses. The main error for frontier models is failing to produce an executable SCM that fits all training worlds (Table C.4).

\FloatBarrier

\section{Disclosure-ladder support}
\label{app:disclosure_ladder_support}

Appendix D supports the disclosure-ladder analysis in Section 6.3. The paired same-latent deltas hold the latent SCM fixed while varying only the information revealed to the model. Matched-pool rates and train-versus-held-out comparisons show the same pattern.

\begin{figure}[!htbp]
\centering
\includegraphics[width=0.94\linewidth]{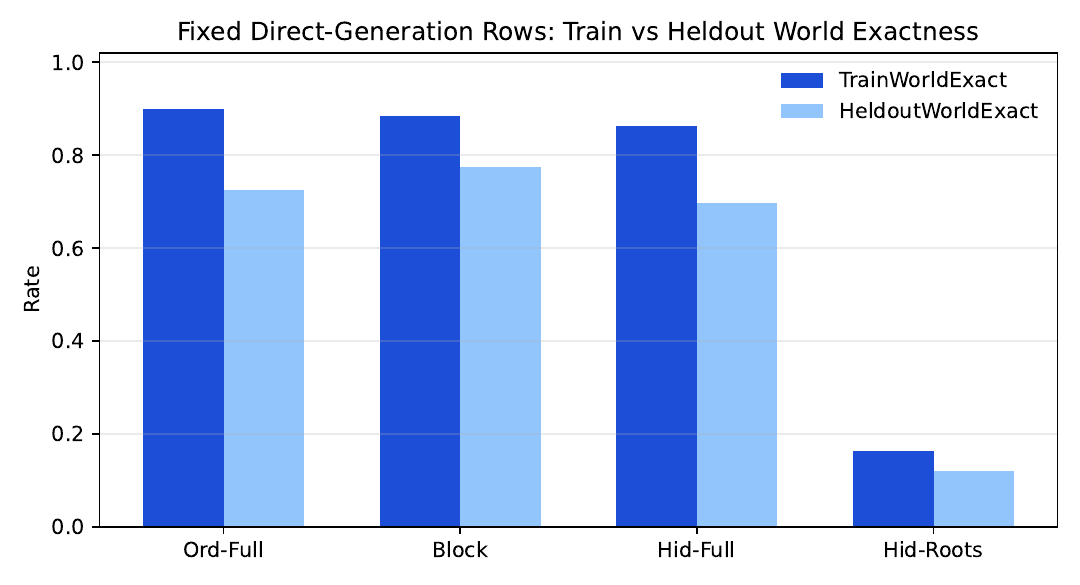}
\caption{\textbf{Training versus held-out replay for representative LLM rows.} Even among the strongest high-coverage LLM rows, executable replay degrades on held-out interventions.}
\label{fig:train_heldout_gap_appendix}
\end{figure}

The common 100-problem matched pool follows the same TrainExact ordering, although several structural facts change across that ladder at once. The train-versus-held-out comparison shows that executable mechanism induction cannot be reduced to fitting the exposed worlds alone (Table~\ref{tab:information_ladder_appendix} and Figure~\ref{fig:train_heldout_gap_appendix}).

% ---- inlined from generated/tables/information_tax_paired_deltas_selected.tex ----
\begin{table}[!htbp]
\centering
\scriptsize
\setlength{\tabcolsep}{3pt}
\begin{tabular}{p{0.25\linewidth}p{0.18\linewidth}ccc}
\toprule
Same-latent pair & Model & $n$ & $\Delta$ TrainExact & $\Delta$ HeldoutWorld \\
\midrule
Ord-Match -> Block & GPT-5.4 & 100 & -0.020 & -0.026 \\
Ord-Match -> Block & Opus~4.6 & 100 & 0.160 & 0.081 \\
Ord-Match -> Block & DeepSeek4Pro & 100 & \textbf{-0.070} & 0.089 \\
Ord-Match -> Block & Gemini3.1 & 100 & 0.090 & 0.098 \\
Ord-Match -> Block & Grok~4.20 & 100 & 0.150 & 0.084 \\
Ord-Match -> Block & Grok~4 & 100 & 0.070 & 0.051 \\
Ord-Match -> Block & Grok~4.3 & 100 & -0.010 & 0.003 \\
Ord-Match -> Block & KimiK2t & 100 & 0.000 & 0.016 \\
Ord-Match -> Block & DSReasoner & 100 & 0.000 & \textbf{-0.029} \\
\addlinespace
Block -> Hid-Match & GPT-5.4 & 100 & \textbf{-0.020} & \textbf{0.051} \\
Block -> Hid-Match & Opus~4.6 & 100 & 0.140 & 0.177 \\
Block -> Hid-Match & DeepSeek4Pro & 100 & 0.140 & 0.099 \\
Block -> Hid-Match & Gemini3.1 & 100 & 0.170 & 0.175 \\
Block -> Hid-Match & Grok~4.20 & 100 & \textbf{-0.020} & 0.107 \\
Block -> Hid-Match & Grok~4 & 100 & 0.090 & 0.145 \\
Block -> Hid-Match & Grok~4.3 & 100 & 0.040 & 0.054 \\
Block -> Hid-Match & KimiK2t & 100 & 0.010 & \textbf{0.051} \\
Block -> Hid-Match & DSReasoner & 100 & 0.010 & 0.062 \\
\addlinespace
Ord-Match -> Hid-Match & GPT-5.4 & 100 & \textbf{-0.040} & \textbf{0.025} \\
Ord-Match -> Hid-Match & Opus~4.6 & 100 & 0.300 & 0.259 \\
Ord-Match -> Hid-Match & DeepSeek4Pro & 100 & 0.070 & 0.188 \\
Ord-Match -> Hid-Match & Gemini3.1 & 100 & 0.260 & 0.273 \\
Ord-Match -> Hid-Match & Grok~4.20 & 100 & 0.130 & 0.191 \\
Ord-Match -> Hid-Match & Grok~4 & 100 & 0.160 & 0.196 \\
Ord-Match -> Hid-Match & Grok~4.3 & 100 & 0.030 & 0.056 \\
Ord-Match -> Hid-Match & KimiK2t & 100 & 0.010 & 0.068 \\
Ord-Match -> Hid-Match & DSReasoner & 100 & 0.010 & 0.034 \\
\addlinespace
Hid-Match -> Hid-Roots & GPT-5.4 & 100 & 0.540 & 0.604 \\
Hid-Match -> Hid-Roots & Opus~4.6 & 100 & 0.260 & 0.331 \\
Hid-Match -> Hid-Roots & DeepSeek4Pro & 100 & 0.190 & 0.234 \\
Hid-Match -> Hid-Roots & Gemini3.1 & 100 & 0.070 & 0.268 \\
Hid-Match -> Hid-Roots & Grok~4.20 & 100 & 0.110 & 0.206 \\
Hid-Match -> Hid-Roots & Grok~4 & 100 & 0.100 & 0.184 \\
Hid-Match -> Hid-Roots & Grok~4.3 & 100 & 0.040 & 0.087 \\
Hid-Match -> Hid-Roots & KimiK2t & 100 & \textbf{0.000} & \textbf{0.013} \\
Hid-Match -> Hid-Roots & DSReasoner & 100 & \textbf{0.000} & 0.025 \\
\bottomrule
\end{tabular}
\caption{\textbf{Paired disclosure deltas.} Each row compares paired versions of the same latent SCM under two disclosure settings. Deltas are computed as first setting minus second setting; positive values therefore indicate lower performance in the less-disclosed setting.}
\label{tab:paired_information_tax_appendix}
\end{table}
% ---- end inlined generated/tables/information_tax_paired_deltas_selected.tex ----

Most transitions toward less structural information reduce both TrainExact and HeldoutWorld, with the largest losses on Ordered→Hidden-order and Hidden-order→Hidden-roots. GPT-5.4 is the main exception on Ordered→Hidden-order TrainExact, but even there held-out replay still declines (Table~\ref{tab:paired_information_tax_appendix}).

% ---- inlined from generated/tables/information_ladder_common_pool.tex ----
\begin{table}[!htbp]
\centering
\small
\setlength{\tabcolsep}{3pt}
\resizebox{\linewidth}{!}{%
\begin{tabular}{lcccccccc}
\toprule
Model & Ord-Match & Block & Hid-Match & Hid-Roots & Ord-Ext & Hid-Ext & Ord-CEx & Hid-CEx \\
\midrule
GPT-5.4 & \textbf{0.580} & \textbf{0.600} & \textbf{0.620} & \textbf{0.080} & \textbf{0.630} & \textbf{0.430} & \textbf{0.530} & \textbf{0.434} \\
Opus~4.6 & \textbf{0.580} & 0.420 & 0.280 & 0.020 & 0.380 & 0.270 & 0.460 & 0.300 \\
DeepSeek4Pro & 0.280 & 0.350 & 0.210 & 0.020 & 0.140 & 0.170 & 0.210 & 0.120 \\
Gemini3.1 & 0.370 & 0.280 & 0.110 & 0.040 & 0.300 & 0.180 & 0.300 & 0.210 \\
Grok~4.20 & 0.300 & 0.150 & 0.170 & 0.060 & 0.230 & 0.130 & 0.230 & 0.170 \\
Grok~4 & 0.310 & 0.240 & 0.150 & 0.050 & 0.120 & 0.090 & 0.100 & 0.080 \\
Grok~4.3 & 0.100 & 0.110 & 0.070 & 0.030 & 0.020 & 0.050 & 0.020 & 0.060 \\
KimiK2t & 0.010 & 0.010 & 0.000 & 0.000 & 0.020 & 0.000 & 0.000 & 0.000 \\
DSReasoner & 0.010 & 0.010 & 0.000 & 0.000 & 0.020 & 0.000 & 0.060 & 0.000 \\
\bottomrule
\end{tabular}
}
\caption{\textbf{TrainExact on the common matched pool.} All entries are TrainExact on the same matched source problems or on stronger-support variants derived from them. These rates support the paired-delta analysis, although several structural facts change across the ladder at once.}
\label{tab:information_ladder_appendix}
\end{table}
% ---- end inlined generated/tables/information_ladder_common_pool.tex ----

TrainExact on the common matched pool is consistent with the paired-delta analysis, without isolating one disclosure change at a time (Table~\ref{tab:information_ladder_appendix}).

\FloatBarrier

\section{Robustness and support-audit summaries}
\label{app:robustness_analyses}

\subsection{Generation-time disambiguation}

Adding disambiguation during generation produces better-constrained cohorts. On Ordered (full), the disambiguated cohort preserves training replay while improving held-out replay, shrinking the train-minus-held-out gap, and raising retention. Cohort summaries, pooled uncertainty, per-model deltas, and shortcut-pressure bins all show this pattern (Tables~\ref{tab:cleanliness_main}--\ref{tab:ambiguity_proxy_appendix}).

Hidden-order behaves differently. There, the disambiguated cohort also reduces the train-to-held-out gap, while absolute held-out replay remains lower. This pattern is consistent with stronger finite-evidence support without making the task trivially easier. The ambiguity measures are consistent with this interpretation (Table~\ref{tab:ambiguity_proxy_appendix}).

\subsection{Three-level support-audit matched subsets}

The three-level matched ladder—Original, Extra Worlds, and CEx—asks whether residual ambiguity explains the Ordered/Hidden-order gap. All three levels use the same 100 matched latent SCMs and preserve the held-out worlds. Original and Extra Worlds are pure disclosure pairs: Ordered and Hidden-order share the same training worlds and differ only in whether the topological order is revealed. CEx adds counterexample worlds against discovered alternatives. Its counterexample worlds may be setting-specific because discovered alternatives can differ across disclosure settings.

Extra Worlds raises mean local predecessor-pattern coverage from 0.8949 to 0.9815 and increases the number of fully covered problems from 2/100 to 42/100. CEx then raises mean local predecessor-pattern coverage to 1.0, reaches full local coverage on all 100 problems, and adds separating worlds until no discovered semantic alternative from LLM outputs, symbolic exact-search, or 50-seed bnlearn+DSL searches still fits the training worlds. These additions materially weaken the local-support and discovered-alternative objections while leaving uniqueness and prompt-length control outside the claim.

% ---- inlined from generated/tables/identifiability_audit_performance.tex ----
\begin{table}[!htbp]
\centering
\scriptsize
\setlength{\tabcolsep}{3pt}
\resizebox{\linewidth}{!}{%
\begin{tabular}{p{0.20\linewidth}p{0.18\linewidth}rrrrrrr}
\toprule
Setting & Model & n & Valid & TrainExact & TrainWorld & HeldoutWorld & HeldoutExact & \shortstack{HeldoutWorld\\$|$ TrainExact} \\
\midrule
Ord-Ext & GPT-5.4 & 100 & 1.000 & \textbf{0.630} & \textbf{0.929} & \textbf{0.880} & \textbf{0.530} & 0.958 \\
Ord-Ext & Opus~4.6 & 100 & 0.980 & 0.380 & 0.824 & 0.783 & 0.340 & 0.977 \\
Ord-Ext & DeepSeek4Pro & 100 & 1.000 & 0.140 & 0.489 & 0.465 & 0.100 & 0.938 \\
Ord-Ext & Gemini3.1 & 100 & 1.000 & 0.300 & 0.715 & 0.709 & 0.290 & 0.992 \\
Ord-Ext & Grok~4.20 & 100 & 1.000 & 0.230 & 0.575 & 0.571 & 0.210 & 0.967 \\
Ord-Ext & Grok~4 & 100 & 0.990 & 0.120 & 0.463 & 0.460 & 0.120 & 1.000 \\
Ord-Ext & Grok~4.3 & 100 & 1.000 & 0.020 & 0.150 & 0.146 & 0.020 & * \\
Ord-Ext & KimiK2t & 100 & 0.950 & 0.020 & 0.085 & 0.066 & 0.020 & * \\
Ord-Ext & DSReasoner & 100 & 1.000 & 0.020 & 0.086 & 0.100 & 0.020 & * \\
\addlinespace
Hid-Ext & GPT-5.4 & 100 & 0.990 & \textbf{0.430} & \textbf{0.824} & \textbf{0.783} & \textbf{0.330} & 0.962 \\
Hid-Ext & Opus~4.6 & 100 & 0.950 & 0.270 & 0.729 & 0.670 & 0.230 & 0.972 \\
Hid-Ext & DeepSeek4Pro & 100 & 1.000 & 0.170 & 0.369 & 0.343 & 0.140 & 0.941 \\
Hid-Ext & Gemini3.1 & 100 & 1.000 & 0.180 & 0.545 & 0.507 & 0.160 & 0.979 \\
Hid-Ext & Grok~4.20 & 100 & 0.970 & 0.130 & 0.499 & 0.466 & 0.110 & 0.971 \\
Hid-Ext & Grok~4 & 100 & 0.990 & 0.090 & 0.387 & 0.335 & 0.080 & 0.972 \\
Hid-Ext & Grok~4.3 & 100 & 0.960 & 0.050 & 0.171 & 0.137 & 0.050 & * \\
Hid-Ext & KimiK2t & 100 & 0.970 & 0.000 & 0.047 & 0.032 & 0.000 & – \\
Hid-Ext & DSReasoner & 100 & 0.990 & 0.000 & 0.055 & 0.034 & 0.000 & – \\
\addlinespace
Ord-CEx & GPT-5.4 & 100 & 1.000 & \textbf{0.530} & \textbf{0.873} & \textbf{0.848} & \textbf{0.530} & 1.000 \\
Ord-CEx & Opus~4.6 & 100 & 0.990 & 0.460 & 0.834 & 0.820 & 0.460 & 1.000 \\
Ord-CEx & DeepSeek4Pro & 100 & 1.000 & 0.210 & 0.413 & 0.424 & 0.210 & 1.000 \\
Ord-CEx & Gemini3.1 & 100 & 1.000 & 0.300 & 0.705 & 0.705 & 0.300 & 1.000 \\
Ord-CEx & Grok~4.20 & 100 & 1.000 & 0.230 & 0.596 & 0.566 & 0.230 & 1.000 \\
Ord-CEx & Grok~4 & 100 & 0.990 & 0.100 & 0.473 & 0.474 & 0.100 & 1.000 \\
Ord-CEx & Grok~4.3 & 100 & 0.980 & 0.020 & 0.136 & 0.133 & 0.020 & * \\
Ord-CEx & KimiK2t & 100 & 0.960 & 0.000 & 0.030 & 0.028 & 0.000 & – \\
Ord-CEx & DSReasoner & 100 & 1.000 & 0.060 & 0.176 & 0.175 & 0.060 & 1.000 \\
\addlinespace
Hid-CEx & GPT-5.4 & 99 & 1.000 & \textbf{0.434} & \textbf{0.836} & \textbf{0.803} & \textbf{0.434} & 1.000 \\
Hid-CEx & Opus~4.6 & 100 & 0.950 & 0.300 & 0.715 & 0.694 & 0.300 & 1.000 \\
Hid-CEx & DeepSeek4Pro & 100 & 1.000 & 0.120 & 0.387 & 0.371 & 0.120 & 1.000 \\
Hid-CEx & Gemini3.1 & 100 & 1.000 & 0.210 & 0.561 & 0.554 & 0.210 & 1.000 \\
Hid-CEx & Grok~4.20 & 100 & 1.000 & 0.170 & 0.508 & 0.472 & 0.170 & 1.000 \\
Hid-CEx & Grok~4 & 100 & 0.930 & 0.080 & 0.394 & 0.374 & 0.080 & 1.000 \\
Hid-CEx & Grok~4.3 & 100 & 0.980 & 0.060 & 0.169 & 0.165 & 0.060 & 1.000 \\
Hid-CEx & KimiK2t & 100 & 0.980 & 0.000 & 0.035 & 0.028 & 0.000 & – \\
Hid-CEx & DSReasoner & 100 & 0.990 & 0.000 & 0.054 & 0.051 & 0.000 & – \\
\addlinespace
\bottomrule
\end{tabular}
}
\caption{\textbf{LLM results on the support-audit matched subsets.} $n$ is the number of stored scored responses. \texttt{Ord-Ext}/\texttt{Hid-Ext} denote Extra Worlds; \texttt{Ord-CEx}/\texttt{Hid-CEx} denote Counterexample Audit (CEx). CEx preserves the same held-out worlds and adds training worlds to complete bounded local predecessor-pattern coverage and separate discovered alternatives under the audited searches; these counterexample additions may be setting-specific. In the CEx rows, HeldoutExact equals TrainExact whenever TrainExact is nonzero, so train-exact CEx outputs also replay all held-out worlds exactly. HeldoutWorld is the unconditioned mean held-out world replay rate.}
\label{tab:identifiability_audit_performance}
\end{table}
% ---- end inlined generated/tables/identifiability_audit_performance.tex ----

Replay remains separated by disclosure under stronger evidence (Table E.1). In the CEx rows, train-exact LLM outputs also replay all held-out worlds exactly, yet Hidden-order still reaches TrainExact less often than Ordered. Structural measures show the same pattern. Adding worlds changes both evidence and prompt length, and CEx counterexample additions may be setting-specific; before/after comparisons therefore mix support, prompt-length, and counterexample-audit effects. Original and Extra Worlds are the pure disclosure comparisons; CEx is a same-latent, same-held-out condition with fewer discovered alternatives.

\FloatBarrier

\section{Construction and support-audit analyses}
\label{app:construction_added_world_analyses}

Appendix F gives the detailed construction and support-audit evidence behind Appendix E. Tables~\ref{tab:cleanliness_main}--\ref{tab:ambiguity_proxy_appendix} evaluate generation-time disambiguation and construction quality. Tables~\ref{tab:identifiability_audit_original_vs_audit_appendix}--\ref{tab:alternative_identification_solved_intersection_appendix} follow the same 100 matched SCMs through Original, Extra Worlds, and Counterexample Audit (CEx), testing whether Hidden-order remains harder than Ordered after adding stronger local support and counterexamples against discovered alternatives.

% ---- inlined from generated/tables/cleanliness_validation_main.tex ----
\begin{table}[!htbp]
\centering
\scriptsize
\setlength{\tabcolsep}{2.5pt}
\renewcommand{\arraystretch}{0.90}
\resizebox{\linewidth}{!}{%
\begin{tabular}{p{0.15\linewidth}p{0.24\linewidth}ccc}
\toprule
Setting & Comparison & $\Delta$ HeldoutWorld & $\Delta$ Train-minus-heldout gap & $\Delta$ Retention \\
\midrule
Ord-Match LLM delta (disambiguated $-$ base cohort) & – & +0.087 [+0.072, +0.103] & -0.092 [-0.117, -0.063] & +0.112 [+0.090, +0.131] \\
Hid-Match LLM delta (disambiguated $-$ base cohort) & – & -0.091 [-0.140, -0.028] & -0.032 [-0.044, -0.018] & +0.003 [-0.037, +0.034] \\
\bottomrule
\end{tabular}
}
\caption{\textbf{Pooled LLM cohort deltas with 95\% bootstrap CIs.} Each row compares a later cohort with the corresponding base cohort for the fixed LLM set. Positive $\Delta$ HeldoutWorld and $\Delta$ Retention indicate improvement; negative $\Delta$ Train-minus-heldout gap indicates a smaller overfitting gap.}
\label{tab:cleanliness_main}
\end{table}
% ---- end inlined generated/tables/cleanliness_validation_main.tex ----

The pooled deltas show the cohort effect most clearly on Ordered: held-out replay improves, retention rises, and the train-minus-held-out gap shrinks. On Hidden-order, the main effect is gap shrinkage rather than a comparable gain in absolute held-out replay (Table~\ref{tab:cleanliness_main}).

% ---- inlined from generated/tables/identifiability_audit_construction.tex ----
\begin{table}[!htbp]
\centering
\small
\begin{tabular}{lccc}
\toprule
\multicolumn{4}{l}{\textbf{Panel A: Local support under stronger evidence}} \\
Statistic & Original & Extra Worlds & Counterexample Audit \\
\midrule
Mean local predecessor-pattern coverage & 0.8949 & 0.9815 & 1.0000 \\
Problems with full local coverage & 2/100 & 42/100 & 100/100 \\
\bottomrule
\end{tabular}
\vspace{0.5em}
\begin{tabular}{lcc}
\toprule
\multicolumn{3}{l}{\textbf{Panel B: Training-world additions}} \\
Statistic & Extra Worlds & Counterexample Audit \\
\midrule
Mean Extra Worlds additions/problem & 3.09 (range 3--4) & – \\
Mean new predecessor-pattern gain/problem & 7.23 & – \\
Mean final train worlds/problem & 12.33 (range 11--14) & 14.27 (range 11--23) \\
Held-out worlds/problem & 8 & 8 \\
\bottomrule
\end{tabular}
\caption{\textbf{Three-level support-audit construction summary.} The same 100 latent SCMs underlie the Ordered and Hidden-order variants. Extra Worlds raises mean local predecessor-pattern coverage from 0.8949 to 0.9815. Counterexample Audit (CEx) starts from those records, completes bounded local predecessor-pattern coverage to 1.0, and adds worlds that separate discovered alternatives that fit the training worlds while preserving the original held-out worlds. The comparison strengthens finite evidence but also increases prompt length.}
\label{tab:identifiability_audit_construction}
\end{table}
% ---- end inlined generated/tables/identifiability_audit_construction.tex ----

The three-level support-audit ladder materially strengthens finite evidence. Mean local predecessor-pattern coverage rises from 0.8949 (Original) to 0.9815 (Extra Worlds) to 1.0 (CEx); full local coverage rises from 2/100 to 42/100 to 100/100. CEx also adds separating worlds until no discovered alternative from LLM outputs, symbolic exact-search, or 50-seed bnlearn+DSL searches still fits the training worlds. Together, these steps leave no discovered alternative train-consistent under the implemented audits. They do not prove uniqueness or isolate prompt length from evidence (Table~\ref{tab:identifiability_audit_construction}).

% ---- inlined from generated/tables/cleanliness_by_cohort_selected.tex ----
\begin{table}[!htbp]
\centering
\small
\resizebox{\linewidth}{!}{%
\begin{tabular}{p{0.15\linewidth}p{0.27\linewidth}cccc}
\toprule
Setting & Cohort & TrainWorldExact & HeldoutWorld & \shortstack{Train-minus-held-out\\gap} & Retention \\
\midrule
Ord-Full & base cohort & 0.692 & 0.521 & 0.172 & 0.752 \\
Ord-Match & disambiguated cohort & 0.655 & 0.568 & 0.087 & 0.867 \\
Ord-Ext & extension cohort & 0.567 & 0.505 & 0.062 & 0.891 \\
\addlinespace
Hid-Full & base cohort & 0.568 & 0.464 & 0.104 & 0.816 \\
Hid-Match & disambiguated cohort & 0.508 & 0.418 & 0.090 & 0.822 \\
Hid-Ext & extension cohort & 0.493 & 0.398 & 0.095 & 0.808 \\
\bottomrule
\end{tabular}
}
\caption{\textbf{Cohort-level construction summary.} Ordered shows the clearest held-out improvement under stronger generation-time disambiguation, while Hidden-order shows a narrower train-to-held-out gap without a comparable gain in absolute held-out replay.}
\label{tab:cleanliness_validation_appendix}
\end{table}
% ---- end inlined generated/tables/cleanliness_by_cohort_selected.tex ----

On Ordered, the disambiguated cohort preserves training replay while improving held-out replay and retention. Hidden-order behaves differently: the train-to-held-out gap narrows, but absolute held-out replay remains comparatively low. The newer Hidden-order cohort is therefore better constrained, even though absolute held-out replay remains low (Table~\ref{tab:cleanliness_validation_appendix}).

% ---- inlined from generated/tables/cleanliness_fixed_model_deltas.tex ----
\begin{table}[!htbp]
\centering
\footnotesize
\setlength{\tabcolsep}{1.8pt}
\resizebox{\linewidth}{!}{%
\begin{tabular}{p{0.14\linewidth}p{0.20\linewidth}ccccccc}
\toprule
Setting & Comparison & \shortstack{$\Delta$\\HeldoutWorld} & \shortstack{95\%\\CI} & \shortstack{$\Delta$ Train-minus\\held-out gap} & \shortstack{95\%\\CI} & \shortstack{$\Delta$\\Retention} & \shortstack{95\%\\CI} & \shortstack{$\Delta$\\TrainExact} \\
\midrule
Ord-Match & disambiguated - base cohort & +0.087 & [+0.072, +0.103] & -0.092 & [-0.117, -0.063] & +0.112 & [+0.090, +0.131] & -0.089 \\
Ord-Ext & extension - base cohort & -0.018 & [-0.130, +0.086] & -0.093 & [-0.159, -0.024] & +0.084 & [-0.036, +0.187] & -0.243 \\
Hid-Match & disambiguated - base cohort & -0.091 & [-0.140, -0.028] & -0.032 & [-0.044, -0.018] & +0.003 & [-0.037, +0.034] & -0.113 \\
Hid-Ext & extension - base cohort & -0.157 & [-0.332, -0.004] & -0.038 & [-0.063, -0.009] & -0.055 & [-0.213, +0.063] & -0.227 \\
\bottomrule
\end{tabular}
}
\caption{\textbf{Bootstrap uncertainty for cohort deltas.} All rows use the fixed LLM set ($m=5$). Ordered disambiguated minus base cohort improves HeldoutWorld and retention while shrinking the train-minus-held-out gap. Hidden-order shrinks the gap but has lower absolute HeldoutWorld.}
\label{tab:cleanliness_fixed_model_deltas_appendix}
\end{table}
% ---- end inlined generated/tables/cleanliness_fixed_model_deltas.tex ----

Bootstrap intervals show the same contrast. On Ordered, stronger disambiguation yields held-out gains after uncertainty is quantified. On Hidden-order, the main change is a smaller train-to-held-out gap rather than a clear gain in absolute held-out replay (Table~\ref{tab:cleanliness_fixed_model_deltas_appendix}).

% ---- inlined from generated/tables/cleanliness_by_model_deltas.tex ----
{\footnotesize
\setlength{\tabcolsep}{2pt}
\begin{longtable}{@{}p{0.14\linewidth}p{0.25\linewidth}p{0.12\linewidth}*{3}{>{\raggedleft\arraybackslash}p{0.13\linewidth}}@{}}
\caption{\textbf{Per-model cohort deltas for the fixed LLM set used in Table~\ref{tab:cleanliness_fixed_model_deltas_appendix}.}}\label{tab:cleanliness_by_model_deltas_appendix}\\
\toprule
Setting & Comparison & Model & \shortstack{$\Delta$\\HeldoutWorld} & \shortstack{$\Delta$ Train-minus\\held-out gap} & \shortstack{$\Delta$\\Retention} \\
\midrule
\endfirsthead
\toprule
Setting & Comparison & Model & \shortstack{$\Delta$\\HeldoutWorld} & \shortstack{$\Delta$ Train-minus\\held-out gap} & \shortstack{$\Delta$\\Retention} \\
\midrule
\endhead
\bottomrule
\endlastfoot
Ord-Match & disambiguated - base cohort & GPT-5.4 & \textbf{+0.114} & \textbf{-0.124} & \textbf{+0.135} \\
Ord-Match & disambiguated - base cohort & Opus~4.6 & +0.065 & -0.123 & +0.121 \\
Ord-Match & disambiguated - base cohort & Gemini3.1 & +0.100 & -0.102 & \textbf{+0.135} \\
Ord-Match & disambiguated - base cohort & Grok~4.20 & +0.069 & -0.037 & +0.067 \\
Ord-Match & disambiguated - base cohort & Grok~4 & +0.068 & -0.054 & +0.087 \\
\addlinespace
Ord-Ext & extension - base cohort & GPT-5.4 & -0.139 & \textbf{-0.215} & +0.209 \\
Ord-Ext & extension - base cohort & Opus~4.6 & +0.044 & -0.096 & +0.092 \\
Ord-Ext & extension - base cohort & Gemini3.1 & +0.116 & -0.159 & \textbf{+0.211} \\
Ord-Ext & extension - base cohort & Grok~4.20 & \textbf{+0.120} & -0.119 & +0.189 \\
Ord-Ext & extension - base cohort & Grok~4 & -0.254 & +0.010 & -0.182 \\
\addlinespace
Hid-Match & disambiguated - base cohort & GPT-5.4 & \textbf{+0.057} & -0.037 & +0.047 \\
Hid-Match & disambiguated - base cohort & Opus~4.6 & -0.172 & -0.038 & -0.003 \\
Hid-Match & disambiguated - base cohort & Gemini3.1 & -0.082 & \textbf{-0.044} & \textbf{+0.050} \\
Hid-Match & disambiguated - base cohort & Grok~4.20 & -0.105 & -0.019 & -0.005 \\
Hid-Match & disambiguated - base cohort & Grok~4 & -0.145 & -0.003 & -0.086 \\
\addlinespace
Hid-Ext & extension - base cohort & GPT-5.4 & -0.180 & -0.049 & +0.001 \\
Hid-Ext & extension - base cohort & Opus~4.6 & -0.187 & +0.020 & -0.079 \\
Hid-Ext & extension - base cohort & Gemini3.1 & -0.025 & \textbf{-0.063} & \textbf{+0.101} \\
Hid-Ext & extension - base cohort & Grok~4.20 & -0.110 & -0.015 & -0.014 \\
Hid-Ext & extension - base cohort & Grok~4 & \textbf{+0.120} & -0.037 & +0.090 \\
\end{longtable}
}
% ---- end inlined generated/tables/cleanliness_by_model_deltas.tex ----

Per-model deltas show the same pattern across the fixed LLM set: Ordered has the clearest held-out gains from stronger disambiguation, while Hidden-order mainly shows narrower train-to-held-out gaps (Table~\ref{tab:cleanliness_by_model_deltas_appendix}).

\clearpage
% ---- inlined from generated/tables/ambiguity_proxy_selected.tex ----
\begin{table}[!htbp]
\centering
\small
\begin{tabular}{p{0.16\linewidth}p{0.29\linewidth}ccc}
\toprule
Setting & Ambiguity proxy bin & HeldoutWorld & Retention & \shortstack{HeldoutWorld\\$|$ TrainExact} \\
\midrule
Ord-Full & shortcut-kill coverage <=0.50 & 0.549 & 0.855 & 0.873 \\
Ord-Full & shortcut-kill coverage >0.75 & 0.546 & 0.715 & 0.739 \\
Hid-Full & shortcut-kill coverage <=0.50 & 0.431 & 0.817 & 0.857 \\
Hid-Full & shortcut-kill coverage >0.75 & 0.474 & 0.773 & 0.838 \\
\bottomrule
\end{tabular}
\caption{\textbf{Ambiguity-proxy bins.} Stronger shortcut pressure is associated with weaker retention and held-out replay, especially on Ordered.}
\label{tab:ambiguity_proxy_appendix}
\end{table}
% ---- end inlined generated/tables/ambiguity_proxy_selected.tex ----

The ambiguity-proxy bins connect the cohort effect back to construction-time metadata. On Ordered, stronger shortcut pressure is associated with weaker held-out replay and weaker retention, suggesting that the generator's ambiguity measures track real difficulty (Table~\ref{tab:ambiguity_proxy_appendix}).

\subsection{Support-audit matched subsets}

The support-audit subsets address the residual-ambiguity objection by analyzing replay, structure, and discovered alternatives. Table~\ref{tab:identifiability_audit_original_vs_audit_appendix} tracks replay across Original, Extra Worlds, and CEx. Tables~\ref{tab:audit_structural_disclosure_appendix}--\ref{tab:structural_bootstrap_cis_appendix} track semantic structure for Original and Extra Worlds. Tables~\ref{tab:alternative_identification_problem_coverage_appendix}--\ref{tab:alternative_identification_solved_intersection_appendix} track discovered alternatives that fit the training worlds before CEx construction and show how Extra Worlds reduces them. Original and Extra Worlds are pure disclosure comparisons because Ordered and Hidden-order share latent SCMs, training worlds, and held-out worlds. CEx uses the same latent SCMs and held-out worlds and leaves zero surviving discovered alternatives under the audited pools; its counterexample additions may be setting-specific.

% ---- inlined from generated/tables/identifiability_audit_original_vs_audit.tex ----
\begin{table}[!htbp]
\centering
\scriptsize
\setlength{\tabcolsep}{3pt}
\resizebox{\linewidth}{!}{%
\begin{tabular}{p{0.14\linewidth}p{0.20\linewidth}cccccc}
\toprule
& & \multicolumn{2}{c}{Original} & \multicolumn{2}{c}{Extra Worlds} & \multicolumn{2}{c}{Counterexample Audit} \\
\cmidrule(lr){3-4}\cmidrule(lr){5-6}\cmidrule(lr){7-8}
Setting & Model & TrainExact & \shortstack{Held-out\\world exact} & TrainExact & \shortstack{Held-out\\world exact} & TrainExact & \shortstack{Held-out\\world exact} \\
\midrule
Ord-Match & GPT-5.4 & \textbf{0.580} & 0.748 & \textbf{0.630} & \textbf{0.880} & \textbf{0.530} & \textbf{0.848} \\
Ord-Match & Opus~4.6 & \textbf{0.580} & \textbf{0.757} & 0.380 & 0.783 & 0.460 & 0.820 \\
Ord-Match & DeepSeek4Pro & 0.280 & 0.499 & 0.140 & 0.465 & 0.210 & 0.424 \\
Ord-Match & Gemini3.1 & 0.370 & 0.656 & 0.300 & 0.709 & 0.300 & 0.705 \\
Ord-Match & Grok~4.20 & 0.300 & 0.539 & 0.230 & 0.571 & 0.230 & 0.566 \\
Ord-Match & Grok~4 & 0.310 & 0.504 & 0.120 & 0.460 & 0.100 & 0.474 \\
Ord-Match & Grok~4.3 & 0.100 & 0.207 & 0.020 & 0.146 & 0.020 & 0.133 \\
Ord-Match & KimiK2t & 0.010 & 0.098 & 0.020 & 0.066 & 0.000 & 0.028 \\
Ord-Match & DSReasoner & 0.010 & 0.062 & 0.020 & 0.100 & 0.060 & 0.175 \\
Hid-Match & GPT-5.4 & \textbf{0.620} & \textbf{0.723} & \textbf{0.430} & \textbf{0.783} & \textbf{0.434} & \textbf{0.803} \\
Hid-Match & Opus~4.6 & 0.280 & 0.489 & 0.270 & 0.670 & 0.300 & 0.694 \\
Hid-Match & DeepSeek4Pro & 0.210 & 0.321 & 0.170 & 0.343 & 0.120 & 0.371 \\
Hid-Match & Gemini3.1 & 0.110 & 0.384 & 0.180 & 0.507 & 0.210 & 0.554 \\
Hid-Match & Grok~4.20 & 0.170 & 0.358 & 0.130 & 0.466 & 0.170 & 0.472 \\
Hid-Match & Grok~4 & 0.150 & 0.293 & 0.090 & 0.335 & 0.080 & 0.374 \\
Hid-Match & Grok~4.3 & 0.070 & 0.168 & 0.050 & 0.137 & 0.060 & 0.165 \\
Hid-Match & KimiK2t & 0.000 & 0.027 & 0.000 & 0.032 & 0.000 & 0.028 \\
Hid-Match & DSReasoner & 0.000 & 0.027 & 0.000 & 0.034 & 0.000 & 0.051 \\
\bottomrule
\end{tabular}
}
\caption{\textbf{Replay across the three support-audit levels.} Original, Extra Worlds, and Counterexample Audit (CEx) compare the same matched source problems under increasingly strong finite-evidence support; Held-out world exact denotes HeldoutWorldExact, and the held-out worlds are preserved. Before/after differences combine stronger local evidence with longer prompts, and CEx can add setting-specific counterexample worlds; these rows therefore include prompt-length and audit-construction effects rather than a pure disclosure ablation.}
\label{tab:identifiability_audit_original_vs_audit_appendix}
\end{table}
% ---- end inlined generated/tables/identifiability_audit_original_vs_audit.tex ----

Within each support level, the Ordered/Hidden-order separation remains visible in replay. Original and Extra Worlds provide pure disclosure comparisons; CEx keeps the same latent SCMs and held-out worlds while reducing discovered alternatives (Table~\ref{tab:identifiability_audit_original_vs_audit_appendix}).

In the extra-worlds benchmarks, the Hidden-order gap remains visible in semantic parent structure and exact parent-map matching on the same latent problems (Table~\ref{tab:audit_structural_disclosure_appendix}).

% ---- inlined from generated/tables/audit_structural_disclosure_comparison.tex ----
\begin{table}[!htbp]
\centering
\footnotesize
\setlength{\tabcolsep}{2pt}
\begin{tabular}{@{}p{0.20\linewidth}*{6}{>{\raggedleft\arraybackslash}p{0.095\linewidth}}@{}}
\toprule
\multicolumn{7}{l}{\textbf{Panel A: Validity and parent-graph structure}} \\
Model & \shortstack{Valid\\Ord-Ext} & \shortstack{Valid\\Hid-Ext} & \shortstack{Parent\\F1\\Ord-Ext} & \shortstack{Parent\\F1\\Hid-Ext} & \shortstack{Parent\\SHD\\Ord-Ext} & \shortstack{Parent\\SHD\\Hid-Ext} \\
\midrule
GPT-5.4 & 1.000 & 0.990 & \textbf{0.975} & \textbf{0.920} & \textbf{0.88} & \textbf{2.52} \\
Opus~4.6 & 0.980 & 0.950 & 0.944 & 0.874 & 1.79 & 3.71 \\
DeepSeek4Pro & 1.000 & 1.000 & 0.873 & 0.728 & 4.01 & 7.77 \\
Gemini3.1 & 1.000 & 1.000 & 0.923 & 0.846 & 2.62 & 4.74 \\
Grok~4.20 & 1.000 & 0.970 & 0.898 & 0.790 & 3.38 & 6.28 \\
Grok~4 & 0.990 & 0.990 & 0.839 & 0.717 & 4.90 & 7.93 \\
Grok~4.3 & 1.000 & 0.960 & 0.714 & 0.541 & 7.31 & 11.18 \\
KimiK2t & 0.950 & 0.970 & 0.675 & 0.463 & 8.76 & 13.34 \\
DSReasoner & 1.000 & 0.990 & 0.686 & 0.403 & 7.86 & 13.75 \\
\bottomrule
\end{tabular}
\vspace{0.5em}
\begin{tabular}{@{}p{0.20\linewidth}*{6}{>{\raggedleft\arraybackslash}p{0.095\linewidth}}@{}}
\toprule
\multicolumn{7}{l}{\textbf{Panel B: Exact structure and replay}} \\
Model & \shortstack{Exact\\parent map\\Ord-Ext} & \shortstack{Exact\\parent map\\Hid-Ext} & \shortstack{TrainExact\\Ord-Ext} & \shortstack{TrainExact\\Hid-Ext} & \shortstack{Heldout\\World\\Ord-Ext} & \shortstack{Heldout\\World\\Hid-Ext} \\
\midrule
GPT-5.4 & \textbf{0.610} & \textbf{0.313} & \textbf{0.630} & \textbf{0.430} & \textbf{0.880} & \textbf{0.783} \\
Opus~4.6 & 0.418 & 0.242 & 0.380 & 0.270 & 0.783 & 0.670 \\
DeepSeek4Pro & 0.110 & 0.120 & 0.140 & 0.170 & 0.465 & 0.343 \\
Gemini3.1 & 0.360 & 0.190 & 0.300 & 0.180 & 0.709 & 0.507 \\
Grok~4.20 & 0.270 & 0.134 & 0.230 & 0.130 & 0.571 & 0.467 \\
Grok~4 & 0.162 & 0.091 & 0.120 & 0.090 & 0.460 & 0.335 \\
Grok~4.3 & 0.020 & 0.052 & 0.020 & 0.050 & 0.146 & 0.137 \\
KimiK2t & 0.021 & 0.000 & 0.020 & 0.000 & 0.066 & 0.032 \\
DSReasoner & 0.020 & 0.000 & 0.020 & 0.000 & 0.100 & 0.034 \\
\bottomrule
\end{tabular}
\caption{\textbf{Structural comparison on the Extra Worlds subset.} Ordered + Extra Worlds and Hidden-order + Extra Worlds use the same latent SCMs, the same additional training worlds, and the same held-out worlds; only revealed order differs. Panel A reports validity and parent-graph structure, and Panel B reports exact parent maps and replay.}
\label{tab:audit_structural_disclosure_appendix}
\end{table}
% ---- end inlined generated/tables/audit_structural_disclosure_comparison.tex ----

\FloatBarrier

% ---- inlined from generated/tables/audit_original_vs_audit_structural.tex ----
{\scriptsize
\setlength{\tabcolsep}{2pt}
\begin{longtable}{@{}p{0.17\linewidth}*{6}{>{\raggedleft\arraybackslash}p{0.108\linewidth}}@{}}
\caption{\textbf{Structure before and after added worlds.} The same source problems and held-out worlds are compared before and after additional training worlds are added. Panel A reports parent-graph changes, and Panel B reports local semantic match and replay.}\label{tab:audit_original_vs_audit_structural_appendix}\\
\toprule
\multicolumn{7}{l}{\textbf{Panel A: Parent-graph structure}} \\
Model & \shortstack{Parent\\F1\\Original} & \shortstack{Parent\\F1\\Extra\\worlds} & \shortstack{Parent\\SHD\\Original} & \shortstack{Parent\\SHD\\Extra\\worlds} & \shortstack{Exact parent\\map\\Original} & \shortstack{Exact parent\\map\\Extra\\worlds} \\
\midrule
\endfirsthead
\toprule
\multicolumn{7}{l}{\textbf{Panel B: Local semantics and replay (continued)}} \\
& \multicolumn{2}{c}{Local semantic match} & \multicolumn{2}{c}{TrainExact} & \multicolumn{2}{c}{HeldoutWorld} \\
\cmidrule(lr){2-3}\cmidrule(lr){4-5}\cmidrule(lr){6-7}
Model & Original & Extra Worlds & Original & Extra Worlds & Original & Extra Worlds \\
\midrule
\endhead
\bottomrule
\endlastfoot
\multicolumn{7}{@{}l}{\textbf{Ord-Match}} \\
GPT-5.4 & 0.952 & \textbf{0.975} & 1.70 & \textbf{0.88} & \textbf{0.460} & \textbf{0.610} \\
Opus~4.6 & \textbf{0.956} & 0.944 & \textbf{1.59} & 1.79 & 0.380 & 0.410 \\
DeepSeek4Pro & 0.872 & 0.873 & 4.11 & 4.01 & 0.160 & 0.110 \\
Gemini3.1 & 0.927 & 0.923 & 2.44 & 2.62 & 0.390 & 0.360 \\
Grok~4.20 & 0.903 & 0.898 & 3.34 & 3.38 & 0.210 & 0.270 \\
Grok~4 & 0.877 & 0.839 & 4.24 & 4.90 & 0.146 & 0.162 \\
Grok~4.3 & 0.749 & 0.714 & 6.71 & 7.31 & 0.071 & 0.020 \\
KimiK2t & 0.681 & 0.675 & 8.84 & 8.76 & 0.010 & 0.021 \\
DSReasoner & 0.674 & 0.686 & 8.14 & 7.86 & 0.010 & 0.020 \\
\addlinespace
\multicolumn{7}{@{}l}{\textbf{Hid-Match}} \\
GPT-5.4 & \textbf{0.896} & \textbf{0.920} & \textbf{3.23} & \textbf{2.52} & \textbf{0.310} & \textbf{0.313} \\
Opus~4.6 & 0.795 & 0.874 & 5.85 & 3.71 & 0.180 & 0.230 \\
DeepSeek4Pro & 0.732 & 0.728 & 7.47 & 7.77 & 0.093 & 0.120 \\
Gemini3.1 & 0.797 & 0.846 & 5.68 & 4.74 & 0.170 & 0.190 \\
Grok~4.20 & 0.756 & 0.790 & 6.85 & 6.28 & 0.150 & 0.130 \\
Grok~4 & 0.745 & 0.717 & 7.52 & 7.93 & 0.061 & 0.091 \\
Grok~4.3 & 0.525 & 0.541 & 11.86 & 11.18 & 0.038 & 0.052 \\
KimiK2t & 0.431 & 0.463 & 13.87 & 13.34 & 0.000 & 0.000 \\
DSReasoner & 0.376 & 0.403 & 13.76 & 13.75 & 0.000 & 0.000 \\
\addlinespace
\midrule
\multicolumn{7}{l}{\textbf{Panel B: Local semantics and replay}} \\
& \multicolumn{2}{c}{Local semantic match} & \multicolumn{2}{c}{TrainExact} & \multicolumn{2}{c}{HeldoutWorld} \\
\cmidrule(lr){2-3}\cmidrule(lr){4-5}\cmidrule(lr){6-7}
Model & Original & Extra Worlds & Original & Extra Worlds & Original & Extra Worlds \\
\midrule
\multicolumn{7}{@{}l}{\textbf{Ord-Match}} \\
GPT-5.4 & 0.763 & \textbf{0.866} & \textbf{0.580} & \textbf{0.630} & 0.748 & \textbf{0.880} \\
Opus~4.6 & \textbf{0.769} & 0.785 & \textbf{0.580} & 0.380 & \textbf{0.757} & 0.783 \\
DeepSeek4Pro & 0.616 & 0.619 & 0.280 & 0.140 & 0.499 & 0.465 \\
Gemini3.1 & 0.733 & 0.746 & 0.370 & 0.300 & 0.656 & 0.709 \\
Grok~4.20 & 0.656 & 0.704 & 0.300 & 0.230 & 0.539 & 0.571 \\
Grok~4 & 0.621 & 0.580 & 0.310 & 0.120 & 0.504 & 0.460 \\
Grok~4.3 & 0.429 & 0.388 & 0.100 & 0.020 & 0.207 & 0.146 \\
KimiK2t & 0.295 & 0.341 & 0.010 & 0.020 & 0.098 & 0.066 \\
DSReasoner & 0.308 & 0.347 & 0.010 & 0.020 & 0.062 & 0.100 \\
\addlinespace
\multicolumn{7}{@{}l}{\textbf{Hid-Match}} \\
GPT-5.4 & \textbf{0.710} & \textbf{0.774} & \textbf{0.620} & \textbf{0.430} & \textbf{0.723} & \textbf{0.783} \\
Opus~4.6 & 0.575 & 0.682 & 0.280 & 0.270 & 0.489 & 0.670 \\
DeepSeek4Pro & 0.468 & 0.529 & 0.210 & 0.170 & 0.321 & 0.343 \\
Gemini3.1 & 0.576 & 0.629 & 0.110 & 0.180 & 0.384 & 0.507 \\
Grok~4.20 & 0.519 & 0.577 & 0.170 & 0.130 & 0.358 & 0.467 \\
Grok~4 & 0.492 & 0.491 & 0.150 & 0.090 & 0.293 & 0.335 \\
Grok~4.3 & 0.259 & 0.286 & 0.070 & 0.050 & 0.168 & 0.137 \\
KimiK2t & 0.124 & 0.133 & 0.000 & 0.000 & 0.027 & 0.032 \\
DSReasoner & 0.120 & 0.142 & 0.000 & 0.000 & 0.027 & 0.034 \\
\addlinespace
\end{longtable}
}
% ---- end inlined generated/tables/audit_original_vs_audit_structural.tex ----

The comparison of original versus additional training worlds mixes two changes at once: increased intervention coverage, which reduces local mechanism ambiguity, and increased prompt length. Even so, the same held-out worlds become harder to explain with spurious alternatives once stronger local support is present (Table~\ref{tab:audit_original_vs_audit_structural_appendix}).

% ---- inlined from generated/tables/structural_bootstrap_cis.tex ----
{\footnotesize
\setlength{\tabcolsep}{2pt}
\begin{longtable}{@{}p{0.16\linewidth}*{3}{>{\raggedleft\arraybackslash}p{0.22\linewidth}}@{}}
\caption{\textbf{Bootstrap intervals for added-world analyses.} The Matched Ordered/Matched Hidden-order rows isolate revealed-order effects on the matched subset, whereas the original-to-extra-worlds rows quantify how much the added evidence changes the same source problems.}\label{tab:structural_bootstrap_cis_appendix}\\
\toprule
\multicolumn{4}{l}{\textbf{Panel A: Parent-graph intervals}} \\
Model & $\Delta$ Parent SHD $\downarrow$ & $\Delta$ Parent F1 & $\Delta$ Exact parent map \\
\midrule
\endfirsthead
\toprule
Model & $\Delta$ Local semantic match & $\Delta$ HeldoutWorld & \\
\midrule
\endhead
\bottomrule
\endlastfoot
\multicolumn{4}{@{}l}{\textbf{Hid-Ext - Ord-Ext}} \\
GPT-5.4 & +1.66 [+1.00, +2.36] & -0.055 [-0.080, -0.033] & -0.303 [-0.404, -0.192] \\
Opus~4.6 & +1.96 [+1.18, +2.72] & -0.071 [-0.102, -0.040] & -0.172 [-0.290, -0.043] \\
DeepSeek4Pro & +3.76 [+2.63, +4.87] & -0.145 [-0.191, -0.100] & +0.010 [-0.070, +0.090] \\
Gemini3.1 & +2.12 [+1.38, +2.89] & -0.076 [-0.102, -0.049] & -0.170 [-0.260, -0.090] \\
Grok~4.20 & +2.79 [+1.79, +3.85] & -0.106 [-0.148, -0.067] & -0.113 [-0.196, -0.041] \\
Grok~4 & +3.05 [+2.06, +4.09] & -0.122 [-0.164, -0.075] & -0.071 [-0.153, +0.010] \\
Grok~4.3 & +3.96 [+2.84, +5.00] & -0.178 [-0.228, -0.125] & +0.031 [-0.010, +0.083] \\
KimiK2t & +4.87 [+3.92, +5.79] & -0.219 [-0.260, -0.180] & -0.022 [-0.054, +0.000] \\
DSReasoner & +5.89 [+5.05, +6.73] & -0.282 [-0.325, -0.241] & -0.020 [-0.051, +0.000] \\
\addlinespace
\multicolumn{4}{@{}l}{\textbf{Ord-Ext - Ord-Match}} \\
GPT-5.4 & -0.82 [-1.27, -0.40] & +0.023 [+0.009, +0.038] & +0.150 [+0.040, +0.260] \\
Opus~4.6 & +0.14 [-0.33, +0.60] & -0.011 [-0.031, +0.006] & +0.021 [-0.107, +0.150] \\
DeepSeek4Pro & -0.10 [-0.83, +0.68] & +0.002 [-0.025, +0.031] & -0.050 [-0.130, +0.030] \\
Gemini3.1 & +0.18 [-0.49, +0.83] & -0.005 [-0.025, +0.015] & -0.030 [-0.120, +0.060] \\
Grok~4.20 & +0.04 [-0.62, +0.74] & -0.004 [-0.026, +0.018] & +0.060 [-0.020, +0.130] \\
Grok~4 & +0.63 [-0.15, +1.41] & -0.036 [-0.070, -0.007] & +0.021 [-0.063, +0.116] \\
Grok~4.3 & +0.57 [-0.12, +1.31] & -0.034 [-0.071, +0.001] & -0.051 [-0.102, +0.000] \\
KimiK2t & -0.08 [-0.98, +0.88] & -0.005 [-0.044, +0.034] & +0.011 [-0.022, +0.054] \\
DSReasoner & -0.23 [-0.88, +0.41] & +0.010 [-0.019, +0.040] & +0.000 [+0.000, +0.000] \\
\addlinespace
\multicolumn{4}{@{}l}{\textbf{Hid-Ext - Hid-Match}} \\
GPT-5.4 & -0.69 [-1.41, +0.03] & +0.024 [-0.001, +0.050] & +0.000 [-0.101, +0.091] \\
Opus~4.6 & -1.89 [-2.99, -0.89] & +0.076 [+0.033, +0.118] & +0.056 [-0.033, +0.144] \\
DeepSeek4Pro & +0.38 [-0.69, +1.47] & -0.008 [-0.054, +0.036] & +0.031 [-0.031, +0.093] \\
Gemini3.1 & -0.94 [-1.72, -0.13] & +0.049 [+0.017, +0.079] & +0.020 [-0.060, +0.090] \\
Grok~4.20 & -0.64 [-1.79, +0.53] & +0.037 [-0.013, +0.085] & -0.011 [-0.096, +0.074] \\
Grok~4 & +0.28 [-0.80, +1.38] & -0.023 [-0.069, +0.023] & +0.031 [-0.031, +0.102] \\
Grok~4.3 & -0.37 [-1.35, +0.68] & +0.005 [-0.045, +0.055] & +0.027 [-0.013, +0.080] \\
KimiK2t & -0.40 [-1.36, +0.51] & +0.030 [-0.012, +0.073] & +0.000 [+0.000, +0.000] \\
DSReasoner & -0.04 [-0.87, +0.75] & +0.027 [-0.013, +0.067] & +0.000 [+0.000, +0.000] \\
\addlinespace
\midrule
\multicolumn{4}{l}{\textbf{Panel B: Local semantics and held-out replay intervals}} \\
Model & $\Delta$ Local semantic match & $\Delta$ HeldoutWorld & \\
\midrule
\multicolumn{4}{@{}l}{\textbf{Hid-Ext - Ord-Ext}} \\
GPT-5.4 & -0.095 [-0.131, -0.060] & -0.100 [-0.155, -0.043] &  \\
Opus~4.6 & -0.106 [-0.161, -0.054] & -0.125 [-0.185, -0.060] &  \\
DeepSeek4Pro & -0.091 [-0.150, -0.035] & -0.122 [-0.214, -0.031] &  \\
Gemini3.1 & -0.117 [-0.162, -0.077] & -0.201 [-0.263, -0.139] &  \\
Grok~4.20 & -0.124 [-0.174, -0.077] & -0.095 [-0.166, -0.022] &  \\
Grok~4 & -0.093 [-0.148, -0.031] & -0.124 [-0.195, -0.050] &  \\
Grok~4.3 & -0.109 [-0.165, -0.056] & -0.016 [-0.074, +0.046] &  \\
KimiK2t & -0.212 [-0.261, -0.167] & -0.035 [-0.077, +0.004] &  \\
DSReasoner & -0.204 [-0.254, -0.156] & -0.067 [-0.115, -0.027] &  \\
\addlinespace
\multicolumn{4}{@{}l}{\textbf{Ord-Ext - Ord-Match}} \\
GPT-5.4 & +0.103 [+0.061, +0.147] & +0.133 [+0.077, +0.189] &  \\
Opus~4.6 & +0.022 [-0.019, +0.062] & +0.025 [-0.032, +0.085] &  \\
DeepSeek4Pro & +0.003 [-0.056, +0.058] & -0.034 [-0.120, +0.049] &  \\
Gemini3.1 & +0.013 [-0.030, +0.057] & +0.052 [-0.022, +0.125] &  \\
Grok~4.20 & +0.048 [+0.012, +0.087] & +0.033 [-0.030, +0.096] &  \\
Grok~4 & -0.039 [-0.092, +0.012] & -0.034 [-0.120, +0.045] &  \\
Grok~4.3 & -0.039 [-0.086, +0.006] & -0.057 [-0.117, +0.001] &  \\
KimiK2t & +0.042 [-0.005, +0.093] & -0.026 [-0.069, +0.024] &  \\
DSReasoner & +0.031 [-0.005, +0.063] & +0.028 [-0.008, +0.064] &  \\
\addlinespace
\multicolumn{4}{@{}l}{\textbf{Hid-Ext - Hid-Match}} \\
GPT-5.4 & +0.064 [+0.021, +0.109] & +0.056 [-0.008, +0.117] &  \\
Opus~4.6 & +0.099 [+0.044, +0.154] & +0.165 [+0.089, +0.240] &  \\
DeepSeek4Pro & +0.056 [-0.000, +0.110] & +0.026 [-0.043, +0.088] &  \\
Gemini3.1 & +0.053 [+0.016, +0.089] & +0.124 [+0.064, +0.185] &  \\
Grok~4.20 & +0.061 [+0.005, +0.119] & +0.116 [+0.036, +0.196] &  \\
Grok~4 & +0.003 [-0.049, +0.052] & +0.045 [-0.028, +0.117] &  \\
Grok~4.3 & +0.030 [-0.017, +0.077] & -0.005 [-0.058, +0.047] &  \\
KimiK2t & +0.011 [-0.027, +0.050] & +0.000 [-0.025, +0.024] &  \\
DSReasoner & +0.023 [-0.015, +0.063] & +0.006 [-0.022, +0.035] &  \\
\addlinespace
\end{longtable}
}
% ---- end inlined generated/tables/structural_bootstrap_cis.tex ----

Bootstrap intervals keep these effects separate. The matched Ordered-versus-Hidden-order rows quantify the residual disclosure penalty after additional worlds are added, whereas the original-to-extra-worlds rows quantify how much stronger local support changes the same source problems (Table~\ref{tab:structural_bootstrap_cis_appendix}).

Additional training worlds make TrainExact stricter, so before-versus-after comparisons couple support gains with task changes. At the same time, when the comparison is made on matched latent SCMs with matched held-out worlds, Hidden-order remains structurally and behaviorally harder than Ordered (Table~\ref{tab:structural_bootstrap_cis_appendix}).

\subsection{Alternative discovery under stronger support}

Because the search is over the available result pool, the numbers in this subsection are discovery counts rather than absence proofs. Even so, they provide a useful proxy for how much stronger local support narrows the space of discovered alternatives. These tables report discovery on Original and Extra Worlds. CEx adds separating worlds against discovered alternatives that survive Extra Worlds. Because none of those audited alternatives remain, CEx is summarized through its construction and replay results rather than as another discovery row.

% ---- inlined from generated/tables/alternative_identification_problem_coverage.tex ----
\begin{table}[!htbp]
\centering
\footnotesize
\setlength{\tabcolsep}{3.2pt}
\begin{tabular}{@{}p{0.22\linewidth}*{4}{>{\raggedleft\arraybackslash}p{0.13\linewidth}}@{}}
\toprule
Setting & \shortstack{No LLM-\\discovered alt.} & \shortstack{No symbolic\\alt.} & \shortstack{No alt. from\\any system} & \shortstack{Any alt.\\rate} \\
\midrule
Ord-Match & 38 & 58 & 24 & 0.760 \\
Hid-Match & 42 & 52 & 28 & 0.720 \\
\addlinespace
Ord-Ext & 73 & 80 & 66 & 0.340 \\
Hid-Ext & 78 & 75 & 66 & 0.340 \\
\addlinespace
Ord-CEx & 100 & 100 & 100 & 0.000 \\
Hid-CEx & 100 & 100 & 100 & 0.000 \\
\bottomrule
\end{tabular}
\caption{\textbf{Problem-level alternative discovery across support-audit levels.} The first three numeric columns are problem counts out of 100; Any alternative rate is the fraction of problems with at least one discovered semantic alternative that fits the training worlds. Original and Extra Worlds are discovery counts and rates in the available result pools. CEx rows are post-audit checks after adding counterexamples against surviving discovered alternatives; zero rates mean no discovered alternative remains under the implemented audited searches, not that uniqueness is proven.}
\label{tab:alternative_identification_problem_coverage_appendix}
\end{table}
% ---- end inlined generated/tables/alternative_identification_problem_coverage.tex ----

From Original to Extra Worlds, the fraction of problems with any discovered alternative that fits the training worlds drops sharply for both Ordered and Hidden-order. CEx then adds a stronger empirical filter by explicitly separating discovered alternatives that survive the Extra Worlds records until the audited discovered-alternative count is zero. This pattern is consistent with stronger finite-evidence support under the added evidence (Table~\ref{tab:alternative_identification_problem_coverage_appendix}).

% ---- inlined from generated/tables/alternative_identification_rates_by_model.tex ----
{\scriptsize
\setlength{\tabcolsep}{2.1pt}
\renewcommand{\arraystretch}{0.92}
\begin{longtable}{@{}p{0.15\linewidth}p{0.20\linewidth}*{5}{>{\raggedleft\arraybackslash}p{0.105\linewidth}}@{}}
\caption{\textbf{Per-model alternative-discovery rates across support-audit levels.} An alternative identification is an executable SCM that is TrainExact and semantically distinct from the gold SCM under local Boolean mechanism truth-table equality; syntactic rewrites of the gold mechanisms are excluded. The two rate columns divide alternative identifications by attempted submissions and by train-exact solutions. Original and Extra Worlds report discovered alternatives in the available result pools. CEx rows report the post-audit check after counterexample additions, where no discovered alternative remains under the implemented audited searches. The * and – notation follows Table~\ref{tab:topline_results}.}\label{tab:alternative_identification_rates_by_model_appendix}\\
\toprule
Setting & System & Attempted & \shortstack{Train-exact\\solutions} & \shortstack{Alternative\\identifications} & \shortstack{Alt. rate\\/ attempt} & \shortstack{Alt. rate\\/ train-exact} \\
\midrule
\endfirsthead
\toprule
Setting & System & Attempted & \shortstack{Train-exact\\solutions} & \shortstack{Alternative\\identifications} & \shortstack{Alt. rate\\/ attempt} & \shortstack{Alt. rate\\/ train-exact} \\
\midrule
\endhead
Ord-Match & GPT-5.4 & 100 & \textbf{58} & 27 & 0.270 & 0.466 \\
Ord-Match & Opus~4.6 & 100 & \textbf{58} & 31 & 0.310 & 0.534 \\
Ord-Match & DeepSeek4Pro & 100 & 28 & 13 & 0.130 & 0.464 \\
Ord-Match & Gemini3.1 & 100 & 37 & 8 & 0.080 & 0.216 \\
Ord-Match & Grok~4.20 & 100 & 30 & 14 & 0.140 & 0.467 \\
Ord-Match & Grok~4 & 100 & 31 & 20 & 0.200 & 0.645 \\
Ord-Match & Grok~4.3 & 100 & 10 & 5 & 0.050 & 0.500 \\
Ord-Match & KimiK2t & 100 & 1 & 1 & 0.010 & * \\
Ord-Match & DSReasoner & 100 & 1 & 0 & 0.000 & * \\
Ord-Match & bnlearn+DSL & 100 & 100 & 38 & 0.380 & 0.380 \\
Ord-Match & symbolic exact-search & 100 & 97 & 32 & 0.320 & 0.330 \\
\addlinespace
Hid-Match & GPT-5.4 & 100 & \textbf{62} & 36 & 0.360 & 0.581 \\
Hid-Match & Opus~4.6 & 100 & 28 & 15 & 0.150 & 0.536 \\
Hid-Match & DeepSeek4Pro & 100 & 21 & 12 & 0.120 & 0.571 \\
Hid-Match & Gemini3.1 & 100 & 11 & 0 & 0.000 & 0.000 \\
Hid-Match & Grok~4.20 & 100 & 17 & 6 & 0.060 & 0.353 \\
Hid-Match & Grok~4 & 100 & 15 & 9 & 0.090 & 0.600 \\
Hid-Match & Grok~4.3 & 100 & 7 & 5 & 0.050 & \textbf{0.714} \\
Hid-Match & KimiK2t & 100 & 0 & 0 & 0.000 & – \\
Hid-Match & DSReasoner & 100 & 0 & 0 & 0.000 & – \\
Hid-Match & bnlearn+DSL & 100 & 100 & 45 & 0.450 & 0.450 \\
Hid-Match & symbolic exact-search & 100 & 77 & 36 & 0.360 & 0.468 \\
\addlinespace
Ord-Ext & GPT-5.4 & 100 & \textbf{63} & 13 & 0.130 & 0.206 \\
Ord-Ext & Opus~4.6 & 100 & 38 & 8 & 0.080 & 0.210 \\
Ord-Ext & DeepSeek4Pro & 100 & 14 & 6 & 0.060 & \textbf{0.429} \\
Ord-Ext & Gemini3.1 & 100 & 30 & 1 & 0.010 & 0.033 \\
Ord-Ext & Grok~4.20 & 100 & 23 & 3 & 0.030 & 0.130 \\
Ord-Ext & Grok~4 & 100 & 12 & 1 & 0.010 & 0.083 \\
Ord-Ext & Grok~4.3 & 100 & 2 & 0 & 0.000 & * \\
Ord-Ext & KimiK2t & 100 & 2 & 0 & 0.000 & * \\
Ord-Ext & DSReasoner & 100 & 2 & 0 & 0.000 & * \\
Ord-Ext & bnlearn+DSL & 100 & 100 & 18 & 0.180 & 0.180 \\
Ord-Ext & symbolic exact-search & 100 & 94 & 14 & 0.140 & 0.149 \\
\addlinespace
Hid-Ext & GPT-5.4 & 100 & \textbf{43} & 17 & 0.170 & \textbf{0.395} \\
Hid-Ext & Opus~4.6 & 100 & 27 & 7 & 0.070 & 0.259 \\
Hid-Ext & DeepSeek4Pro & 100 & 17 & 5 & 0.050 & 0.294 \\
Hid-Ext & Gemini3.1 & 100 & 18 & 2 & 0.020 & 0.111 \\
Hid-Ext & Grok~4.20 & 100 & 13 & 3 & 0.030 & 0.231 \\
Hid-Ext & Grok~4 & 100 & 9 & 1 & 0.010 & 0.111 \\
Hid-Ext & Grok~4.3 & 100 & 5 & 0 & 0.000 & * \\
Hid-Ext & KimiK2t & 100 & 0 & 0 & 0.000 & – \\
Hid-Ext & DSReasoner & 100 & 0 & 0 & 0.000 & – \\
Hid-Ext & bnlearn+DSL & 100 & 99 & 24 & 0.240 & 0.242 \\
Hid-Ext & symbolic exact-search & 100 & 63 & 14 & 0.140 & 0.222 \\
\addlinespace
Ord-CEx & GPT-5.4 & 100 & \textbf{53} & 0 & 0.000 & 0.000 \\
Ord-CEx & Opus~4.6 & 100 & 46 & 0 & 0.000 & 0.000 \\
Ord-CEx & DeepSeek4Pro & 100 & 21 & 0 & 0.000 & 0.000 \\
Ord-CEx & Gemini3.1 & 100 & 30 & 0 & 0.000 & 0.000 \\
Ord-CEx & Grok~4.20 & 100 & 23 & 0 & 0.000 & 0.000 \\
Ord-CEx & Grok~4 & 100 & 10 & 0 & 0.000 & 0.000 \\
Ord-CEx & Grok~4.3 & 100 & 2 & 0 & 0.000 & * \\
Ord-CEx & KimiK2t & 100 & 0 & 0 & 0.000 & – \\
Ord-CEx & DSReasoner & 100 & 6 & 0 & 0.000 & 0.000 \\
Ord-CEx & bnlearn+DSL & 100 & 100 & 0 & 0.000 & 0.000 \\
Ord-CEx & symbolic exact-search & 100 & 95 & 0 & 0.000 & 0.000 \\
Ord-CEx & bnlearn+DSL 50-seed audit & 50 & 50 & 0 & 0.000 & 0.000 \\
\addlinespace
Hid-CEx & GPT-5.4 & 99 & \textbf{43} & 0 & 0.000 & 0.000 \\
Hid-CEx & Opus~4.6 & 100 & 30 & 0 & 0.000 & 0.000 \\
Hid-CEx & DeepSeek4Pro & 100 & 12 & 0 & 0.000 & 0.000 \\
Hid-CEx & Gemini3.1 & 100 & 21 & 0 & 0.000 & 0.000 \\
Hid-CEx & Grok~4.20 & 100 & 17 & 0 & 0.000 & 0.000 \\
Hid-CEx & Grok~4 & 100 & 8 & 0 & 0.000 & 0.000 \\
Hid-CEx & Grok~4.3 & 100 & 6 & 0 & 0.000 & 0.000 \\
Hid-CEx & KimiK2t & 100 & 0 & 0 & 0.000 & – \\
Hid-CEx & DSReasoner & 100 & 0 & 0 & 0.000 & – \\
Hid-CEx & bnlearn+DSL & 100 & 100 & 0 & 0.000 & 0.000 \\
Hid-CEx & symbolic exact-search & 100 & 95 & 0 & 0.000 & 0.000 \\
Hid-CEx & bnlearn+DSL 50-seed audit & 50 & 50 & 0 & 0.000 & 0.000 \\
\addlinespace
\bottomrule
\end{longtable}
}
% ---- end inlined generated/tables/alternative_identification_rates_by_model.tex ----

The reduction also appears at the system level. Systems that solve many original matched problems contribute most heavily to the available alternative pool, and they also show some of the clearest drops after augmentation (Table~\ref{tab:alternative_identification_rates_by_model_appendix}).

% ---- inlined from generated/tables/alternative_identification_original_vs_audit.tex ----
\begin{table}[!htbp]
\centering
\scriptsize
\setlength{\tabcolsep}{1.6pt}
\begin{tabular}{@{}p{0.17\linewidth}*{7}{>{\raggedleft\arraybackslash}p{0.098\linewidth}}@{}}
\toprule
& \multicolumn{3}{c}{Original} & \multicolumn{3}{c}{Extra Worlds} & \\
\cmidrule(lr){2-4}\cmidrule(lr){5-7}
System & \shortstack{Train-exact\\solutions} & \shortstack{Alt. rate\\/ attempt} & \shortstack{Alt. rate\\/ train-\\exact} & \shortstack{Train-exact\\solutions} & \shortstack{Alt. rate\\/ attempt} & \shortstack{Alt. rate\\/ train-\\exact} & \shortstack{$\Delta$ Alt. rate\\/ attempt} \\
\midrule
\multicolumn{8}{@{}l}{\textbf{Ord-Match}} \\
GPT-5.4 & 58 & 0.270 & 0.466 & 63 & 0.130 & 0.206 & -0.140 \\
Opus~4.6 & 58 & 0.310 & 0.534 & 38 & 0.080 & 0.210 & -0.230 \\
DeepSeek4Pro & 28 & 0.130 & 0.464 & 14 & 0.060 & \textbf{0.429} & -0.070 \\
Gemini3.1 & 37 & 0.080 & 0.216 & 30 & 0.010 & 0.033 & -0.070 \\
Grok~4.20 & 30 & 0.140 & 0.467 & 23 & 0.030 & 0.130 & -0.110 \\
Grok~4 & 31 & 0.200 & 0.645 & 12 & 0.010 & 0.083 & -0.190 \\
Grok~4.3 & 10 & 0.050 & 0.500 & 2 & 0.000 & * & -0.050 \\
KimiK2t & 1 & 0.010 & * & 2 & 0.000 & * & -0.010 \\
DSReasoner & 1 & 0.000 & * & 2 & 0.000 & * & \textbf{+0.000} \\
bnlearn+DSL & \textbf{100} & \textbf{0.380} & 0.380 & \textbf{100} & \textbf{0.180} & 0.180 & -0.200 \\
symbolic exact-search & 97 & 0.320 & 0.330 & 94 & 0.140 & 0.149 & -0.180 \\
\addlinespace
\multicolumn{8}{@{}l}{\textbf{Hid-Match}} \\
GPT-5.4 & 62 & 0.360 & 0.581 & 43 & 0.170 & \textbf{0.395} & -0.190 \\
Opus~4.6 & 28 & 0.150 & 0.536 & 27 & 0.070 & 0.259 & -0.080 \\
DeepSeek4Pro & 21 & 0.120 & 0.571 & 17 & 0.050 & 0.294 & -0.070 \\
Gemini3.1 & 11 & 0.000 & 0.000 & 18 & 0.020 & 0.111 & \textbf{+0.020} \\
Grok~4.20 & 17 & 0.060 & 0.353 & 13 & 0.030 & 0.231 & -0.030 \\
Grok~4 & 15 & 0.090 & 0.600 & 9 & 0.010 & 0.111 & -0.080 \\
Grok~4.3 & 7 & 0.050 & \textbf{0.714} & 5 & 0.000 & * & -0.050 \\
KimiK2t & 0 & 0.000 & – & 0 & 0.000 & – & +0.000 \\
DSReasoner & 0 & 0.000 & – & 0 & 0.000 & – & +0.000 \\
bnlearn+DSL & \textbf{100} & \textbf{0.450} & 0.450 & \textbf{99} & \textbf{0.240} & 0.242 & -0.210 \\
symbolic exact-search & 77 & 0.360 & 0.468 & 63 & 0.140 & 0.222 & -0.220 \\
\addlinespace
\bottomrule
\end{tabular}
\caption{\textbf{Alternative discovery before and after Extra Worlds.} Original and Extra Worlds columns separate train-exact solution counts, alternative-identification rates among all attempts, and alternative-identification rates among train-exact solutions. The delta column is Extra Worlds minus Original for the rate among attempts. These before/after rates summarize discovered alternatives in the available result pool; they do not prove that no other alternatives exist. CEx then adds separating worlds against surviving discovered alternatives. The * and – notation follows Table~\ref{tab:topline_results}.}
\label{tab:alternative_identification_original_vs_audit_appendix}
\end{table}
% ---- end inlined generated/tables/alternative_identification_original_vs_audit.tex ----

Alternative discovery falls both in aggregate and within train-exact responses, so the drop cannot be attributed only to fewer systems solving the training worlds (Table~\ref{tab:alternative_identification_original_vs_audit_appendix}).

% ---- inlined from generated/tables/alternative_identification_solved_intersection.tex ----
\begin{table}[!htbp]
\centering
\scriptsize
\setlength{\tabcolsep}{2.6pt}
\begin{tabular}{@{}p{0.20\linewidth}*{6}{>{\raggedleft\arraybackslash}p{0.11\linewidth}}@{}}
\toprule
System & $n$ solved both & \shortstack{Original\\alternatives} & \shortstack{Extra Worlds\\alternatives} & \shortstack{Original\\alt. rate} & \shortstack{Extra Worlds\\alt. rate} & $\Delta$ rate \\
\midrule
\multicolumn{7}{@{}l}{\textbf{Ord-Match}} \\
GPT-5.4 & 44 & 18 & 7 & 0.409 & 0.159 & -0.250 \\
Opus~4.6 & 30 & 12 & 6 & 0.400 & \textbf{0.200} & -0.200 \\
DeepSeek4Pro & 5 & 0 & 0 & * & * & * \\
Gemini3.1 & 21 & 3 & 1 & 0.143 & 0.048 & -0.095 \\
Grok~4.20 & 18 & 5 & 1 & 0.278 & 0.056 & -0.222 \\
Grok~4 & 11 & 6 & 0 & \textbf{0.545} & 0.000 & -0.545 \\
Grok~4.3 & 1 & 0 & 0 & * & * & * \\
KimiK2t & – & – & – & – & – & – \\
DSReasoner & 1 & 0 & 0 & * & * & * \\
bnlearn+DSL & 100 & \textbf{38} & \textbf{18} & 0.380 & 0.180 & -0.200 \\
symbolic exact-search & 94 & 30 & 14 & 0.319 & 0.149 & -0.170 \\
\addlinespace
\multicolumn{7}{@{}l}{\textbf{Hid-Match}} \\
GPT-5.4 & 33 & 14 & 11 & 0.424 & \textbf{0.333} & -0.091 \\
Opus~4.6 & 15 & 5 & 4 & 0.333 & 0.267 & -0.067 \\
DeepSeek4Pro & 12 & 6 & 2 & 0.500 & 0.167 & -0.333 \\
Gemini3.1 & 9 & 0 & 0 & 0.000 & 0.000 & \textbf{+0.000} \\
Grok~4.20 & 6 & 2 & 0 & 0.333 & 0.000 & -0.333 \\
Grok~4 & 5 & 3 & 0 & * & * & * \\
Grok~4.3 & 4 & 2 & 0 & * & * & * \\
KimiK2t & – & – & – & – & – & – \\
DSReasoner & – & – & – & – & – & – \\
bnlearn+DSL & 99 & \textbf{44} & \textbf{24} & 0.444 & 0.242 & -0.202 \\
symbolic exact-search & 62 & 25 & 14 & 0.403 & 0.226 & -0.177 \\
\addlinespace
\bottomrule
\end{tabular}
\caption{\textbf{Solved-intersection control for alternative discovery before CEx construction.} Restricting to problems solved on both sides tests whether the reduction in discovered alternatives is a solved-set artifact. The drop from Original to Extra Worlds persists within this solved intersection. The * and – notation follows Table~\ref{tab:topline_results}.}
\label{tab:alternative_identification_solved_intersection_appendix}
\end{table}
% ---- end inlined generated/tables/alternative_identification_solved_intersection.tex ----

The drop in discovered alternatives persists after restricting attention to the solved intersection. The reduction reflects more than a solved-set artifact (Table~\ref{tab:alternative_identification_solved_intersection_appendix}).

Extra Worlds and CEx improve local support and reduce discovered alternatives, while Hidden-order remains harder than Ordered on matched latent SCMs.

\FloatBarrier

\section{Supplementary-setting analyses}
\label{app:supplementary_setting_analyses}

Appendix G analyzes the two supplementary settings. Alternative-SCM tests local editing when a valid SCM is supplied (Tables G.1--G.5). Hidden-roots separates root-set prediction from downstream mechanism induction (Tables G.6--G.8).

% ---- inlined from generated/tables/alt_exp_scaffold_diagnostic.tex ----
\begin{table}[!htbp]
\centering
\footnotesize
\setlength{\tabcolsep}{3pt}
\begin{tabular}{llccc}
\toprule
Diagnostic pair & Model & \shortstack{Paired hidden-task\\correctness} & \shortstack{Alternative-SCM\\joint success} & \shortstack{Alternative-SCM success\\$|$ hidden-task failure} \\
\midrule
Ord-Match -> Alt-Ord & GPT-5.4 & \textbf{0.580} & \textbf{0.960} & \textbf{0.929} \\
Ord-Match -> Alt-Ord & Opus~4.6 & \textbf{0.580} & 0.730 & 0.667 \\
Ord-Match -> Alt-Ord & DeepSeek4Pro & 0.280 & 0.800 & 0.806 \\
Ord-Match -> Alt-Ord & Gemini3.1 & 0.370 & 0.920 & 0.873 \\
Ord-Match -> Alt-Ord & Grok~4.20 & 0.300 & 0.800 & 0.743 \\
Ord-Match -> Alt-Ord & Grok~4 & 0.310 & 0.890 & 0.870 \\
Ord-Match -> Alt-Ord & Grok~4.3 & 0.100 & 0.420 & 0.400 \\
Ord-Match -> Alt-Ord & KimiK2t & 0.010 & 0.240 & 0.242 \\
Ord-Match -> Alt-Ord & DSReasoner & 0.010 & 0.030 & 0.030 \\
Hid-Match -> Alt-Hid & GPT-5.4 & \textbf{0.620} & 0.840 & 0.789 \\
Hid-Match -> Alt-Hid & Opus~4.6 & 0.280 & 0.700 & 0.708 \\
Hid-Match -> Alt-Hid & DeepSeek4Pro & 0.210 & 0.670 & 0.633 \\
Hid-Match -> Alt-Hid & Gemini3.1 & 0.110 & \textbf{0.950} & \textbf{0.944} \\
Hid-Match -> Alt-Hid & Grok~4.20 & 0.170 & 0.730 & 0.699 \\
Hid-Match -> Alt-Hid & Grok~4 & 0.150 & 0.870 & 0.859 \\
Hid-Match -> Alt-Hid & Grok~4.3 & 0.070 & 0.360 & 0.344 \\
Hid-Match -> Alt-Hid & KimiK2t & 0.000 & 0.180 & 0.180 \\
Hid-Match -> Alt-Hid & DSReasoner & 0.000 & 0.030 & 0.030 \\
\bottomrule
\end{tabular}
\caption{\textbf{Alternative-SCM versus paired hidden induction.} The comparison between Paired hidden-task correctness and Alternative-SCM joint success quantifies how much performance rises once a valid SCM is supplied. The final column conditions Alternative-SCM success on paired hidden-task failure.}
\label{tab:alt_exp_scaffold_diagnostic_appendix}
\end{table}
% ---- end inlined generated/tables/alt_exp_scaffold_diagnostic.tex ----

Supplying a valid reference SCM substantially raises performance. Joint success on Alternative-SCM is much higher than performance on the paired hidden-induction task, especially for Hidden-order source problems. Models can often edit a supplied SCM on problems where they failed to infer that SCM from the intervention worlds (Table~\ref{tab:alt_exp_scaffold_diagnostic_appendix}).

\FloatBarrier

% ---- inlined from generated/tables/alt_exp_leave_model_out.tex ----
{\scriptsize
\setlength{\tabcolsep}{2.2pt}
\renewcommand{\arraystretch}{0.90}
\begin{longtable}{@{}p{0.11\linewidth}p{0.07\linewidth}p{0.05\linewidth}*{5}{>{\raggedleft\arraybackslash}p{0.105\linewidth}}@{}}
\caption{\textbf{Alternative-SCM leave-model-out results.} KP-all uses all known-pair alternatives; OM-src excludes alternatives mined from the evaluated model; SM-src uses alternatives mined from the same model; OE uses open-ended cases. Bold marks the best LLM value within each benchmark and source-relation group.}\label{tab:alt_exp_leave_model_out_appendix}\\
\toprule
Model & \shortstack{Source\\relation} & $n$ & \shortstack{Joint\\success} & \shortstack{Alternative\\success} & \shortstack{Experiment\\valid} & Separates & Witness \\
\midrule
\endfirsthead
\toprule
Model & \shortstack{Source\\relation} & $n$ & \shortstack{Joint\\success} & \shortstack{Alternative\\success} & \shortstack{Experiment\\valid} & Separates & Witness \\
\midrule
\endhead
\multicolumn{8}{@{}l}{\textbf{Alt-Ord}} \\
GPT-5.4 & KP-all & 71 & \textbf{0.972} & \textbf{0.972} & 1.000 & \textbf{0.972} & \textbf{0.972} \\
GPT-5.4 & OM-src & 54 & \textbf{0.963} & \textbf{0.963} & 1.000 & \textbf{0.963} & \textbf{0.963} \\
GPT-5.4 & SM-src & 17 & 1.000 & 1.000 & 1.000 & 1.000 & 1.000 \\
GPT-5.4 & OE & 29 & \textbf{0.931} & \textbf{0.931} & 1.000 & \textbf{0.931} & \textbf{0.931} \\
Opus~4.6 & KP-all & 71 & 0.761 & 0.761 & 1.000 & 0.761 & 0.761 \\
Opus~4.6 & OM-src & 51 & 0.765 & 0.765 & 1.000 & 0.765 & 0.765 \\
Opus~4.6 & SM-src & 20 & 0.750 & 0.750 & 1.000 & 0.750 & 0.750 \\
Opus~4.6 & OE & 29 & 0.655 & 0.655 & 1.000 & 0.655 & 0.655 \\
DeepSeek4Pro & KP-all & 71 & 0.789 & 0.789 & 1.000 & 0.789 & 0.789 \\
DeepSeek4Pro & OM-src & 71 & 0.789 & 0.789 & 1.000 & 0.789 & 0.789 \\
DeepSeek4Pro & OE & 29 & 0.828 & 0.828 & 1.000 & 0.828 & 0.828 \\
Gemini3.1 & KP-all & 71 & 0.944 & 0.944 & 1.000 & 0.944 & 0.944 \\
Gemini3.1 & OM-src & 68 & 0.941 & 0.941 & 1.000 & 0.941 & 0.941 \\
Gemini3.1 & SM-src & 3 & * & * & * & * & * \\
Gemini3.1 & OE & 29 & 0.862 & 0.862 & 1.000 & 0.862 & 0.862 \\
Grok~4.20 & KP-all & 71 & 0.803 & 0.803 & 1.000 & 0.803 & 0.803 \\
Grok~4.20 & OM-src & 70 & 0.800 & 0.800 & 1.000 & 0.800 & 0.800 \\
Grok~4.20 & SM-src & 1 & * & * & * & * & * \\
Grok~4.20 & OE & 29 & 0.793 & 0.793 & 1.000 & 0.793 & 0.793 \\
Grok~4 & KP-all & 71 & 0.930 & 0.930 & 1.000 & 0.930 & 0.930 \\
Grok~4 & OM-src & 64 & 0.922 & 0.922 & 1.000 & 0.922 & 0.922 \\
Grok~4 & SM-src & 7 & 1.000 & 1.000 & 1.000 & 1.000 & 1.000 \\
Grok~4 & OE & 29 & 0.793 & 0.828 & 1.000 & 0.793 & 0.793 \\
Grok~4.3 & KP-all & 71 & 0.479 & 0.479 & 1.000 & 0.479 & 0.479 \\
Grok~4.3 & OM-src & 71 & 0.479 & 0.479 & 1.000 & 0.479 & 0.479 \\
Grok~4.3 & OE & 29 & 0.276 & 0.310 & 1.000 & 0.276 & 0.276 \\
KimiK2t & KP-all & 71 & 0.296 & 0.380 & 0.901 & 0.296 & 0.296 \\
KimiK2t & OM-src & 71 & 0.296 & 0.380 & 0.901 & 0.296 & 0.296 \\
KimiK2t & OE & 29 & 0.103 & 0.138 & 0.862 & 0.103 & 0.103 \\
DSReasoner & KP-all & 71 & 0.042 & 0.070 & 1.000 & 0.042 & 0.042 \\
DSReasoner & OM-src & 71 & 0.042 & 0.070 & 1.000 & 0.042 & 0.042 \\
DSReasoner & OE & 29 & 0.000 & 0.000 & 1.000 & 0.000 & 0.000 \\
\addlinespace
\multicolumn{8}{@{}l}{\textbf{Alt-Hid}} \\
GPT-5.4 & KP-all & 71 & 0.831 & 0.831 & 1.000 & 0.831 & 0.831 \\
GPT-5.4 & OM-src & 54 & 0.833 & 0.833 & 1.000 & 0.833 & 0.833 \\
GPT-5.4 & SM-src & 17 & 0.824 & 0.824 & 1.000 & 0.824 & 0.824 \\
GPT-5.4 & OE & 29 & 0.862 & 0.862 & 1.000 & 0.862 & 0.862 \\
Opus~4.6 & KP-all & 71 & 0.704 & 0.704 & 1.000 & 0.704 & 0.704 \\
Opus~4.6 & OM-src & 51 & 0.745 & 0.745 & 1.000 & 0.745 & 0.745 \\
Opus~4.6 & SM-src & 20 & 0.600 & 0.600 & 1.000 & 0.600 & 0.600 \\
Opus~4.6 & OE & 29 & 0.690 & 0.690 & 0.966 & 0.690 & 0.690 \\
DeepSeek4Pro & KP-all & 71 & 0.704 & 0.704 & 1.000 & 0.704 & 0.704 \\
DeepSeek4Pro & OM-src & 71 & 0.704 & 0.704 & 1.000 & 0.704 & 0.704 \\
DeepSeek4Pro & OE & 29 & 0.586 & 0.621 & 1.000 & 0.586 & 0.586 \\
Gemini3.1 & KP-all & 71 & \textbf{0.930} & \textbf{0.958} & 1.000 & \textbf{0.944} & \textbf{0.930} \\
Gemini3.1 & OM-src & 68 & \textbf{0.926} & \textbf{0.956} & 1.000 & \textbf{0.941} & \textbf{0.926} \\
Gemini3.1 & SM-src & 3 & * & * & * & * & * \\
Gemini3.1 & OE & 29 & \textbf{1.000} & \textbf{1.000} & 1.000 & \textbf{1.000} & \textbf{1.000} \\
Grok~4.20 & KP-all & 71 & 0.747 & 0.747 & 1.000 & 0.747 & 0.747 \\
Grok~4.20 & OM-src & 70 & 0.757 & 0.757 & 1.000 & 0.757 & 0.757 \\
Grok~4.20 & SM-src & 1 & * & * & * & * & * \\
Grok~4.20 & OE & 29 & 0.690 & 0.690 & 1.000 & 0.690 & 0.690 \\
Grok~4 & KP-all & 71 & 0.915 & 0.915 & 1.000 & 0.915 & 0.915 \\
Grok~4 & OM-src & 64 & 0.922 & 0.922 & 1.000 & 0.922 & 0.922 \\
Grok~4 & SM-src & 7 & 0.857 & 0.857 & 1.000 & 0.857 & 0.857 \\
Grok~4 & OE & 29 & 0.759 & 0.759 & 1.000 & 0.759 & 0.759 \\
Grok~4.3 & KP-all & 71 & 0.408 & 0.422 & 1.000 & 0.408 & 0.408 \\
Grok~4.3 & OM-src & 71 & 0.408 & 0.422 & 1.000 & 0.408 & 0.408 \\
Grok~4.3 & OE & 29 & 0.241 & 0.310 & 1.000 & 0.241 & 0.241 \\
KimiK2t & KP-all & 71 & 0.211 & 0.239 & 0.915 & 0.211 & 0.211 \\
KimiK2t & OM-src & 71 & 0.211 & 0.239 & 0.915 & 0.211 & 0.211 \\
KimiK2t & OE & 29 & 0.103 & 0.103 & 0.828 & 0.103 & 0.103 \\
DSReasoner & KP-all & 71 & 0.042 & 0.042 & 1.000 & 0.042 & 0.042 \\
DSReasoner & OM-src & 71 & 0.042 & 0.042 & 1.000 & 0.042 & 0.042 \\
DSReasoner & OE & 29 & 0.000 & 0.000 & 1.000 & 0.000 & 0.000 \\
\addlinespace
\bottomrule
\end{longtable}
}
% ---- end inlined generated/tables/alt_exp_leave_model_out.tex ----

This increase extends beyond models recognizing alternatives sourced from the same model. The leave-model-out rows remain high for the strongest frontier models, so local editing with a supplied SCM remains strong even when same-model sourced alternatives are excluded (Table~\ref{tab:alt_exp_leave_model_out_appendix}).

% ---- inlined from generated/tables/alt_exp_frontier_highlights.tex ----
\begin{table}[!htbp]
\centering
\small
\begin{tabular}{p{0.24\linewidth}p{0.24\linewidth}cc}
\toprule
Setting & Model & \shortstack{Alternative-SCM\\joint success} & \shortstack{Separating-intervention\\optimality} \\
\midrule
Alt-Hid & GPT-5.4 & 0.840 & \textbf{0.968} \\
Alt-Hid & Opus~4.6 & 0.700 & 0.959 \\
Alt-Hid & DeepSeek4Pro & 0.670 & 0.949 \\
Alt-Hid & Gemini3.1 & \textbf{0.950} & 0.928 \\
Alt-Hid & Grok~4.20 & 0.730 & 0.960 \\
Alt-Hid & Grok~4 & 0.870 & 0.966 \\
Alt-Hid & Grok~4.3 & 0.360 & 0.885 \\
Alt-Hid & KimiK2t & 0.180 & 0.944 \\
Alt-Hid & DSReasoner & 0.030 & * \\
\addlinespace
Alt-Ord & GPT-5.4 & \textbf{0.960} & \textbf{0.969} \\
Alt-Ord & Opus~4.6 & 0.730 & 0.933 \\
Alt-Ord & DeepSeek4Pro & 0.800 & 0.950 \\
Alt-Ord & Gemini3.1 & 0.920 & 0.951 \\
Alt-Ord & Grok~4.20 & 0.800 & 0.926 \\
Alt-Ord & Grok~4 & 0.890 & 0.939 \\
Alt-Ord & Grok~4.3 & 0.420 & 0.898 \\
Alt-Ord & KimiK2t & 0.240 & 0.904 \\
Alt-Ord & DSReasoner & 0.030 & * \\
\bottomrule
\end{tabular}
\caption{\textbf{Alternative-SCM intervention quality.} High joint success is usually paired with strong separating interventions, indicating that successful models often provide structured witnesses.}
\label{tab:alt_exp_appendix}
\end{table}
% ---- end inlined generated/tables/alt_exp_frontier_highlights.tex ----

Successful Alternative-SCM outputs are usually paired with strong separating interventions. The setting tests more than the existence of an alternative: it also tests whether the model can separate that alternative efficiently once an SCM is supplied (Table~\ref{tab:alt_exp_appendix}).

The overlap analysis shows how much the supplied SCM helps. Successful alternatives usually remain close to the reference SCM in both semantic parent structure and local semantics, and they are overwhelmingly one-variable edits rather than global rewrites. Supplying the SCM removes most of the structure search (Table~\ref{tab:alt_exp_reference_overlap_appendix}).

% ---- inlined from generated/tables/alt_exp_reference_overlap.tex ----
\begin{table}[!htbp]
\centering
\scriptsize
\setlength{\tabcolsep}{2.2pt}
\resizebox{\linewidth}{!}{%
\begin{tabular}{llrrrrrrrr}
\toprule
Setting & Model & \shortstack{$n$ valid\\alt.} & \shortstack{$n$ alt.\\success} & \shortstack{Parent F1\\vs ref.} & \shortstack{Exact parent\\map vs ref.} & \shortstack{Mean local\\semantic match\\vs ref.} & \shortstack{One-var\\edits} & \shortstack{Same-parent\\one-var} & \shortstack{Changed-parent\\one-var} \\
\midrule
Alt-Ord & GPT-5.4 & 100 & 96 & 0.956 & 0.200 & 0.808 & 0.990 & 0.156 & 0.833 \\
Alt-Ord & Opus~4.6 & 100 & 73 & 0.976 & 0.460 & 0.810 & 1.000 & 0.397 & 0.603 \\
Alt-Ord & DeepSeek4Pro & 100 & 80 & 0.968 & 0.300 & 0.801 & 0.975 & 0.275 & 0.700 \\
Alt-Ord & Gemini3.1 & 100 & 92 & 0.962 & 0.300 & 0.803 & 0.967 & 0.261 & 0.707 \\
Alt-Ord & Grok~4.20 & 100 & 80 & 0.972 & 0.400 & 0.810 & 0.988 & 0.312 & 0.675 \\
Alt-Ord & Grok~4 & 100 & 90 & 0.951 & 0.120 & 0.810 & 1.000 & 0.100 & 0.900 \\
Alt-Ord & Grok~4.3 & 100 & 43 & 0.966 & 0.440 & 0.813 & 1.000 & 0.302 & 0.698 \\
Alt-Ord & KimiK2t & 100 & 31 & 0.981 & 0.600 & 0.759 & 0.935 & 0.419 & 0.516 \\
Alt-Ord & DSReasoner & 100 & 5 & 0.959 & 0.310 & 0.795 & * & * & * \\
\addlinespace[4pt]
Alt-Hid & GPT-5.4 & 100 & 84 & 0.960 & 0.370 & 0.803 & 0.988 & 0.333 & 0.655 \\
Alt-Hid & Opus~4.6 & 99 & 70 & 0.969 & 0.505 & 0.810 & 1.000 & 0.414 & 0.586 \\
Alt-Hid & DeepSeek4Pro & 100 & 68 & 0.960 & 0.380 & 0.806 & 0.985 & 0.382 & 0.603 \\
Alt-Hid & Gemini3.1 & 100 & 97 & 0.949 & 0.310 & 0.788 & 0.907 & 0.289 & 0.619 \\
Alt-Hid & Grok~4.20 & 100 & 73 & 0.974 & 0.510 & 0.807 & 1.000 & 0.466 & 0.534 \\
Alt-Hid & Grok~4 & 100 & 87 & 0.978 & 0.440 & 0.810 & 1.000 & 0.402 & 0.598 \\
Alt-Hid & Grok~4.3 & 100 & 39 & 0.961 & 0.500 & 0.805 & 1.000 & 0.462 & 0.538 \\
Alt-Hid & KimiK2t & 98 & 20 & 0.960 & 0.541 & 0.733 & 1.000 & 0.600 & 0.400 \\
Alt-Hid & DSReasoner & 100 & 3 & 0.950 & 0.420 & 0.778 & * & * & * \\
\bottomrule
\end{tabular}
}
\caption{\textbf{Alternative-SCM reference overlap and edit locality.} Parent-overlap and local-semantic columns compare the model-supplied alternative with the provided reference SCM over valid executable alternatives. The edit-type columns condition on alternative success and separate one-variable same-parent edits from one-variable parent-changing edits.}
\label{tab:alt_exp_reference_overlap_appendix}
\end{table}
% ---- end inlined generated/tables/alt_exp_reference_overlap.tex ----

% ---- inlined from generated/tables/alt_exp_depth_profile.tex ----
\begin{table}[!htbp]
\centering
\scriptsize
\setlength{\tabcolsep}{3pt}
\begin{tabular}{llrrrr}
\toprule
\multicolumn{6}{@{}l}{\textbf{Panel A: Per-variable change rates on alternative-success responses}} \\
Setting & \shortstack{Ref.\\depth} & \shortstack{$n$\\vars} & Changed & \shortstack{Structure +\\mechanism} & \shortstack{Mechanism-\\only} \\
\midrule
Alt-Ord & 1 & 817 & 0.122 & 0.122 & 0.000 \\
Alt-Ord & 2 & 781 & 0.264 & 0.137 & 0.127 \\
Alt-Ord & 3+ & 1529 & 0.193 & 0.150 & 0.043 \\
\addlinespace[4pt]
Alt-Hid & 1 & 809 & 0.048 & 0.048 & 0.000 \\
Alt-Hid & 2 & 778 & 0.273 & 0.119 & 0.153 \\
Alt-Hid & 3+ & 1487 & 0.232 & 0.154 & 0.078 \\
\bottomrule
\end{tabular}

\vspace{0.5em}

\begin{tabular}{llrrrr}
\toprule
\multicolumn{6}{@{}l}{\textbf{Panel B: Composition of changed variables}} \\
Setting & \shortstack{Ref.\\depth} & \shortstack{$n$ changed\\vars} & \shortstack{Share of\\changed vars} & \shortstack{Structure +\\mechanism} & \shortstack{Mechanism-\\only} \\
\midrule
Alt-Ord & 1 & 100 & 0.166 & 1.000 & 0.000 \\
Alt-Ord & 2 & 206 & 0.343 & 0.519 & 0.481 \\
Alt-Ord & 3+ & 295 & 0.491 & 0.776 & 0.224 \\
\addlinespace[4pt]
Alt-Hid & 1 & 39 & 0.065 & 1.000 & 0.000 \\
Alt-Hid & 2 & 212 & 0.356 & 0.439 & 0.561 \\
Alt-Hid & 3+ & 345 & 0.579 & 0.664 & 0.336 \\
\bottomrule
\end{tabular}
\caption{\textbf{Alternative-SCM edit depth relative to the provided reference SCM.} Panel A reports per-variable change rates on the alternative-success subset. Panel B conditions on changed variables and shows where successful edits concentrate by reference depth.}
\label{tab:alt_exp_depth_profile_appendix}
\end{table}
% ---- end inlined generated/tables/alt_exp_depth_profile.tex ----

The local edits cluster across the supplied SCM. Early endogenous layers remain the most stable part of the supplied SCM, whereas most successful changes occur at depth 2 and deeper. In Alternative-SCM (Hidden-order), the remaining flexibility shifts further downstream and more often preserves the parent set while changing only the local rule. In Alternative-SCM (Ordered), successful edits are somewhat shallower and more often change the parent set itself (Table~\ref{tab:alt_exp_depth_profile_appendix}).

\FloatBarrier

Hidden-roots separates two kinds of error. Once the root set is hidden, the model must first identify the correct roots and then produce executable mechanisms for the remaining variables.

% ---- inlined from generated/tables/root_unknown_decomposition.tex ----
\begin{table}[!htbp]
\centering
\small
\begin{tabular}{p{0.26\linewidth}cccc}
\toprule
Model & \shortstack{Full task\\success} & RootExact & \shortstack{Mechanism\\TrainExact} & \shortstack{Mechanism\\HeldoutExact} \\
\midrule
GPT-5.4 & \textbf{0.080} & 0.240 & \textbf{0.080} & 0.020 \\
Opus~4.6 & 0.020 & 0.240 & 0.020 & 0.000 \\
DeepSeek4Pro & 0.020 & 0.360 & 0.020 & 0.010 \\
Gemini3.1 & 0.040 & 0.340 & 0.040 & 0.030 \\
Grok~4.20 & 0.050 & \textbf{0.390} & 0.060 & \textbf{0.050} \\
Grok~4 & 0.040 & 0.300 & 0.050 & 0.030 \\
Grok~4.3 & 0.030 & 0.170 & 0.030 & 0.030 \\
KimiK2t & 0.000 & 0.060 & 0.000 & 0.000 \\
DSReasoner & 0.000 & 0.040 & 0.000 & 0.000 \\
\bottomrule
\end{tabular}
\caption{\textbf{Hidden-roots breakdown.} RootExact, mechanism train exactness, and mechanism held-out replay are reported separately so that root-set prediction is not conflated with downstream mechanism induction. Mechanism HeldoutExact is strict: it requires exact mechanism training replay and exact replay of every held-out world.}
\label{tab:root_unknown_appendix}
\end{table}
% ---- end inlined generated/tables/root_unknown_decomposition.tex ----

Root-set prediction is already hard, and exact mechanism induction remains weak even on the subset where the root set is correct (Table~\ref{tab:root_unknown_appendix}).

% ---- inlined from generated/tables/root_unknown_conditionals.tex ----
\begin{table}[!htbp]
\centering
\scriptsize
\setlength{\tabcolsep}{2pt}
\begin{tabular}{@{}p{0.15\linewidth}>{\raggedleft\arraybackslash}p{0.05\linewidth}>{\raggedleft\arraybackslash}p{0.08\linewidth}>{\raggedleft\arraybackslash}p{0.08\linewidth}>{\raggedleft\arraybackslash}p{0.19\linewidth}>{\raggedleft\arraybackslash}p{0.19\linewidth}>{\raggedleft\arraybackslash}p{0.19\linewidth}@{}}
\toprule
\multicolumn{7}{l}{\textbf{Panel A: Overall rates and RootExact-conditioned replay}} \\
Model & $n$ & RootExact & \shortstack{$n$\\RootExact} & \shortstack{TrainExact\\$|$ RootExact} & \shortstack{HeldoutWorld\\$|$ RootExact} & \shortstack{HeldoutExact\\$|$ RootExact} \\
\midrule
GPT-5.4 & 100 & 0.240 & 24 & \textbf{0.333} & \textbf{0.307} & 0.083 \\
Opus~4.6 & 100 & 0.240 & 24 & 0.083 & 0.260 & 0.000 \\
DeepSeek4Pro & 100 & 0.360 & 36 & 0.056 & 0.195 & 0.028 \\
Gemini3.1 & 100 & 0.340 & 34 & 0.118 & 0.290 & 0.088 \\
Grok~4.20 & 100 & \textbf{0.390} & 39 & 0.128 & 0.215 & 0.128 \\
Grok~4 & 100 & 0.300 & 30 & 0.133 & 0.192 & 0.100 \\
Grok~4.3 & 100 & 0.170 & 17 & 0.176 & 0.243 & \textbf{0.176} \\
KimiK2t & 100 & 0.060 & 6 & 0.000 & 0.062 & 0.000 \\
DSReasoner & 100 & 0.040 & 4 & * & * & * \\
\bottomrule
\end{tabular}
\vspace{0.5em}
\begin{tabular}{@{}p{0.24\linewidth}>{\raggedleft\arraybackslash}p{0.19\linewidth}>{\raggedleft\arraybackslash}p{0.21\linewidth}>{\raggedleft\arraybackslash}p{0.16\linewidth}>{\raggedleft\arraybackslash}p{0.16\linewidth}@{}}
\toprule
\multicolumn{5}{l}{\textbf{Panel B: Root-wrong cases and residual failures}} \\
Model & \shortstack{TrainExact\\| root-wrong} & \shortstack{HeldoutWorld\\| root-wrong} & \shortstack{Train\\failures} & \shortstack{Held-out\\failures} \\
\midrule
GPT-5.4 & 0.000 & 0.060 & 16 & 6 \\
Opus~4.6 & 0.000 & \textbf{0.096} & 22 & 2 \\
DeepSeek4Pro & 0.000 & 0.018 & 34 & 1 \\
Gemini3.1 & 0.000 & 0.026 & 30 & 1 \\
Grok~4.20 & \textbf{0.016} & 0.096 & 34 & \textbf{0} \\
Grok~4 & 0.014 & 0.066 & 26 & 1 \\
Grok~4.3 & 0.000 & 0.022 & 14 & \textbf{0} \\
KimiK2t & 0.000 & 0.011 & 6 & \textbf{0} \\
DSReasoner & 0.000 & 0.001 & \textbf{4} & \textbf{0} \\
\bottomrule
\end{tabular}
\caption{\textbf{Mechanism replay after exact root-set prediction.} The all-problems columns give the total number of problems and RootExact. The RootExact-conditioned columns condition on exact root-set prediction, and the root-wrong columns condition on an incorrect root set. Failure counts report RootExact cases that fail mechanism training fit and RootExact TrainExact cases that fail strict held-out exactness. The * and – notation follows Table~\ref{tab:topline_results}.}
\label{tab:root_unknown_conditionals_appendix}
\end{table}
% ---- end inlined generated/tables/root_unknown_conditionals.tex ----

Even when the root set is correct, many submissions still make downstream mechanism errors (Table~\ref{tab:root_unknown_conditionals_appendix}).

% ---- inlined from generated/tables/root_unknown_mechanism_decomposition.tex ----
\begin{table}[!htbp]
\centering
\scriptsize
\setlength{\tabcolsep}{2pt}
\begin{tabular}{@{}p{0.15\linewidth}>{\raggedleft\arraybackslash}p{0.05\linewidth}>{\raggedleft\arraybackslash}p{0.07\linewidth}>{\raggedleft\arraybackslash}p{0.08\linewidth}>{\raggedleft\arraybackslash}p{0.13\linewidth}>{\raggedleft\arraybackslash}p{0.23\linewidth}>{\raggedleft\arraybackslash}p{0.23\linewidth}@{}}
\toprule
\multicolumn{7}{l}{\textbf{Panel A: Root-set exactness and structural decomposition}} \\
Model & $n$ & Valid & RootExact & \shortstack{$n$ RootExact\\+ valid} & \shortstack{Parent F1\\$|$ RootExact + valid} & \shortstack{Exact parent map\\$|$ RootExact + valid} \\
\midrule
GPT-5.4 & 100 & 0.990 & 0.240 & 24 & \textbf{0.690} & 0.083 \\
Opus~4.6 & 100 & 0.930 & 0.240 & 24 & 0.653 & 0.000 \\
DeepSeek4Pro & 100 & 0.960 & 0.360 & 34 & 0.589 & 0.029 \\
Gemini3.1 & 100 & 1.000 & 0.340 & 34 & 0.661 & 0.118 \\
Grok~4.20 & 100 & 0.990 & \textbf{0.390} & 39 & 0.657 & 0.128 \\
Grok~4 & 100 & 1.000 & 0.300 & 30 & 0.565 & 0.100 \\
Grok~4.3 & 100 & 0.970 & 0.170 & 17 & 0.503 & \textbf{0.176} \\
KimiK2t & 100 & 0.980 & 0.060 & 6 & 0.346 & 0.000 \\
DSReasoner & 100 & 1.000 & 0.040 & 4 & * & * \\
\bottomrule
\end{tabular}
\vspace{0.5em}
\begin{tabular}{@{}p{0.24\linewidth}>{\raggedleft\arraybackslash}p{0.36\linewidth}>{\raggedleft\arraybackslash}p{0.36\linewidth}@{}}
\toprule
\multicolumn{3}{l}{\textbf{Panel B: Mechanism replay after exact root-set prediction}} \\
Model & \shortstack{TrainExact\\$|$ RootExact + valid} & \shortstack{HeldoutWorld\\$|$ RootExact + TrainExact} \\
\midrule
GPT-5.4 & \textbf{0.333} & 0.734 \\
Opus~4.6 & 0.083 & * \\
DeepSeek4Pro & 0.059 & * \\
Gemini3.1 & 0.118 & * \\
Grok~4.20 & 0.128 & * \\
Grok~4 & 0.133 & * \\
Grok~4.3 & 0.176 & * \\
KimiK2t & 0.000 & – \\
DSReasoner & * & – \\
\bottomrule
\end{tabular}
\caption{\textbf{Nested Hidden-roots breakdown.} RootExact separates root-set prediction from downstream mechanism induction. Parent F1, Exact parent map, and TrainExact are conditioned on submissions that are both RootExact and valid; HeldoutWorld is further conditioned on exact training fit. The * and – notation follows Table~\ref{tab:topline_results}.}
\label{tab:root_unknown_mechanism_decomposition_appendix}
\end{table}
% ---- end inlined generated/tables/root_unknown_mechanism_decomposition.tex ----

Nested conditioning on root-exact and valid submissions leads to the same qualitative conclusion: residual structural and mechanism errors remain even after the root set is correct (Table~\ref{tab:root_unknown_mechanism_decomposition_appendix}).

Taken together, root-set prediction and downstream mechanism induction are separate sources of error in Hidden-roots (Tables~\ref{tab:root_unknown_appendix}--\ref{tab:root_unknown_mechanism_decomposition_appendix}).

\FloatBarrier

\section{Non-LLM calibration rows}
\label{app:baseline_details}

The non-LLM rows are fixed benchmark-specific procedures, distinct from off-the-shelf end-to-end causal discovery systems. The symbolic exact-search baseline searches directly in the benchmark SCM language. The bnlearn+DSL baseline uses bnlearn for structure proposal and then invokes the same exact Boolean fitter used by the symbolic pipeline. Both procedures receive only the training worlds and the same revealed-structure fields available to the corresponding LLM prompt, and both are scored by the same evaluator as the LLM submissions.

These rows show which parts of exact executable induction remain difficult under fixed symbolic or hybrid search budgets. They are not a comprehensive comparison to all causal-discovery or program-synthesis methods.

\begin{table}[!htbp]
\centering
\scriptsize
\setlength{\tabcolsep}{2.5pt}
\resizebox{\linewidth}{!}{%
\begin{tabular}{p{0.16\linewidth}p{0.39\linewidth}p{0.39\linewidth}}
\toprule
Component & Symbolic exact-search baseline & bnlearn+DSL baseline \\
\midrule
Input & Training worlds and revealed-structure fields only. & Training worlds and revealed-structure fields only. \\
Structure search & Ordered restricts parents to earlier variables; Block-order respects block precedence and searches admissible within-block dependencies; Hidden-order searches acyclic orders and parent assignments; Hidden-roots enumerates root-set hypotheses. & bnlearn proposes structures with mbde-scored tabu search and hill climbing under the disclosed constraints: full order in Ordered, block constraints in Block-order, no disclosed endogenous order in Hidden-order, and explicit root-set hypotheses in Hidden-roots. \\
Root handling & Uses disclosed roots except in Hidden-roots, where candidate root sets are enumerated before mechanism fitting. & Uses disclosed roots except in Hidden-roots, where roots are latent and searched as hypotheses before structure proposal and fitting. \\
Parent-set candidates & Enumerates admissible parent sets under the setting disclosure, with candidate pools capped by the staged mechanism-search budget. & Tries learned DAG parent sets first, then parent proposals from the graph ensemble and bootstrap arc supports in fallback stages. \\
Mechanism language & Benchmark DSL over \texttt{not}, \texttt{and}, \texttt{or}, \texttt{xor}, and \texttt{iff}; constants are disallowed. & Same benchmark DSL; constants are disallowed. \\
Mechanism fitting & Exact Boolean fitter searches formulas over each candidate parent set and accepts only train-exact mechanisms. & The learned graph is not submitted directly; each proposed parent set is passed to the same exact Boolean fitter. \\
Staged budgets & Three stages: AST caps 8, 10, and 12; AST slack 2, 2, and 3; at most 32, 32, and 256 candidates per variable; 50,000, 50,000, and 100,000 states per size; and 2-second, 20-second, and 600-second per-problem residual budgets. & Four stages: initial tabu mbde search with max 5 parents and 500 iterations; an ensemble with tabu and hill climbing using max 5 parents and up to 2,000 iterations; the same ensemble with parent-proposal fallback; and a final residual stage with max 6 parents, tabu up to 4,000 iterations, hill climbing up to 5,000 iterations, 96 bootstrap replicates where used, exact-fitter AST caps 10, 12, and 13, and a 600-second per-problem budget. \\
Selection rule & Stages are tried in order; the selected row is the first train-exact executable candidate found by the staged ordering, with later stages filling only rows not solved earlier. & Stage order is fixed; exact fits to learned DAGs are tried before parent-proposal fallback, and later stages fill only rows not solved earlier. \\
Failure behavior & If no exact candidate is found within budget, emits a schema-correct failure object that is scored by the same evaluator. & If no exact candidate is found within budget, emits a schema-correct failure object that is scored by the same evaluator. \\
Scoring & Same parser, legality checks, acyclicity check, and train/held-out replay evaluator as LLM submissions. & Same parser, legality checks, acyclicity check, and train/held-out replay evaluator as LLM submissions. \\
\bottomrule
\end{tabular}
}
\caption{\textbf{Non-LLM baseline protocol summary.} Both procedures are fixed calibration rows evaluated under the same final-object contract as the LLM submissions; they are not budget-matched direct-generation LLM baselines.}
\label{tab:nonllm_protocol_summary}
\end{table}

\paragraph{Symbolic exact-search baseline.}
The symbolic exact-search baseline treats each item as an exact finite-world synthesis problem in the benchmark DSL. A candidate consists of a root set, an acyclic functional-parent graph, and one Boolean mechanism for every predicted endogenous variable. In Ordered, parent sets are restricted to earlier variables in the disclosed order. In Block-order, the search respects block precedence and searches admissible within-block dependencies. In Hidden-order, it searches over acyclic orders and parent assignments. In Hidden-roots, it also enumerates root-set hypotheses before mechanism fitting.

The search is run as a staged portfolio with the budgets shown in Table~\ref{tab:nonllm_protocol_summary}. Each accepted candidate must replay all training worlds exactly under the benchmark evaluator. When multiple candidates satisfy the training worlds, the baseline selects the first train-exact executable candidate found by the fixed staged ordering. If no exact candidate is found within budget, it emits a schema-correct failure object, which is then scored by the same evaluator.

\paragraph{bnlearn+DSL baseline.}
The bnlearn+DSL baseline separates structure proposal from mechanism fitting. It first converts the training worlds into the data representation used by bnlearn and runs mbde-scored tabu search and hill climbing under the budgets in Table~\ref{tab:nonllm_protocol_summary}. The structural search respects the disclosure condition: full order in Ordered, block constraints in Block-order, no disclosed endogenous order in Hidden-order, and explicit root-set hypotheses in Hidden-roots.

The learned graph is an intermediate structure proposal. Each learned graph, or the parent proposals derived from the graph ensemble, is passed to the shared exact Boolean fitter, which searches for benchmark-DSL mechanisms over the proposed parents. The fitter uses the same operator set and fitting budgets shown in Table~\ref{tab:nonllm_protocol_summary}. The final candidate is selected by the fixed stage order and then scored by the same evaluator as every other system.

Table H.2 separates full task correctness, training replay, and held-out replay.

% ---- inlined from generated/tables/classical_baseline_decomposition.tex ----
\begin{table}[!htbp]
\centering
\scriptsize
\setlength{\tabcolsep}{2.6pt}
\resizebox{\linewidth}{!}{%
\begin{tabular}{p{0.17\linewidth}p{0.26\linewidth}ccccc}
\toprule
Setting & System & $n$ & TaskCorrect & TrainExact & HeldoutWorld & HeldoutExact \\
\midrule
Ord-Full & bnlearn+DSL & 250 & 0.996 & 0.996 & 0.880 & 0.596 \\
Ord-Full & symbolic exact-search & 250 & 0.980 & 0.980 & 0.884 & 0.596 \\
\addlinespace
Hid-Full & bnlearn+DSL & 250 & 0.996 & 0.996 & 0.896 & 0.620 \\
Hid-Full & symbolic exact-search & 250 & 0.900 & 0.900 & 0.892 & 0.536 \\
\addlinespace
Block & bnlearn+DSL & 100 & 1.000 & 1.000 & 0.899 & 0.620 \\
Block & symbolic exact-search & 100 & 0.960 & 0.960 & 0.927 & 0.650 \\
\addlinespace
Hid-Roots & bnlearn+DSL & 100 & 0.790 & 0.900 & 0.867 & 0.530 \\
Hid-Roots & symbolic exact-search & 100 & 0.820 & 0.950 & 0.880 & 0.560 \\
\bottomrule
\end{tabular}
}
\caption{\textbf{Non-LLM calibration-row breakdown.} The rows separate full task correctness, training replay, and held-out replay for the fixed symbolic exact-search and bnlearn+DSL procedures. TaskCorrect is full task correctness under the benchmark evaluator; on Hidden-roots it differs from TrainExact because root-set prediction is part of the task. HeldoutWorld is the unconditioned mean held-out world replay rate. HeldoutExact is strict: it requires exact training replay and exact replay of every held-out world.}
\label{tab:classical_baseline_decomposition_appendix}
\end{table}
% ---- end inlined generated/tables/classical_baseline_decomposition.tex ----

\FloatBarrier

\section{Illustrative case studies}
\label{app:case_studies}

These case studies illustrate three recurrent failure modes already visible in the aggregate results. The first shows a hidden-structure shortcut: revealing order removes a plausible but incorrect endogenous-parent substitution. The second shows a train-exact but mechanistically wrong solution that fails only when a new parent configuration is reached. The third shows that the same model can edit a supplied SCM successfully on a paired source problem where it failed to infer that SCM from interventions alone.

\paragraph{Case 1: Revealed order rules out a downstream surrogate.}
In one paired item, GPT-5.4 solves the Ordered version exactly, with TrainWorldExact and HeldoutWorldExact both 1.000. On the corresponding Hidden-order version, the submitted SCM is still executable but no longer train-exact: TrainWorldExact falls to 0.556 and HeldoutWorldExact to 0.625. The key difference is causal admissibility. When the true order is revealed, \texttt{X1} is known to be downstream of \texttt{X2}, so it cannot be used as an input to \texttt{X2}. When order is hidden, the model uses \texttt{X1} as a plausible surrogate for the root/input signal \texttt{X3}. The surrogate is correlated enough to look reasonable on many rows, but it is causally backwards.

{\footnotesize
\begin{center}
\setlength{\tabcolsep}{3pt}
\renewcommand{\arraystretch}{1.08}
\begin{tabular}{@{}p{0.24\linewidth}p{0.70\linewidth}@{}}
\toprule
Item & Detail \\
\midrule
Revealed order & \texttt{X8, X7, X3, X5, X2, X1, X6, X4}; thus \texttt{X1} is not an admissible parent of \texttt{X2}. \\
Gold \texttt{X2} & \texttt{(xor X5 (and (iff X7 X8) (xor X3 X7)))} \\
Ordered submitted \texttt{X2} & \texttt{(or (and X5 (or (iff X3 X7) (iff X3 X8))) (and X7 X8 (not X3) (not X5)) (and X3 (not X5) (not X7) (not X8)))} \\
Hidden-order submitted \texttt{X2} & \texttt{(or (xor X1 X5 X7 X8) (and X5 X8 (not X1)))} \\
Held-out separator & Under \texttt{do(X1=assigned\_per\_row, X5=assigned\_per\_row, X7=assigned\_per\_row)}, with \texttt{X1=1, X3=1, X5=0, X7=1, X8=1}, gold gives \texttt{X2=0}; the Hidden-order answer predicts \texttt{X2=1}. \\
\bottomrule
\end{tabular}
\end{center}
}

\paragraph{Case 2: Train-exact replay can still miss a truth-table corner.}
In a Hidden-order item, GPT-5.4 returns a valid SCM that is fully train-exact: TrainWorldExact is 1.000 and cell-level training accuracy is 1.000. It nevertheless fails strict held-out exactness, with HeldoutWorldExact 0.750 and cell-level held-out accuracy 0.970. The failure is concentrated in one local mechanism. The submitted \texttt{X5} agrees with every scored training row for \texttt{X5}, but the held-out intervention reaches a parent assignment that the training worlds never query.

{\footnotesize
\begin{center}
\setlength{\tabcolsep}{3pt}
\renewcommand{\arraystretch}{1.08}
\begin{tabular}{@{}p{0.24\linewidth}p{0.70\linewidth}@{}}
\toprule
Item & Detail \\
\midrule
Gold \texttt{X5} & \texttt{(iff (and X3 X6) (or X4 X7))} \\
Submitted \texttt{X5} & \texttt{(or (and X3 (or X4 (iff X6 X7))) (and X6 (not (or X3 X4 X7))))} \\
Held-out separator & Under \texttt{do(X2=assigned\_per\_row, X3=assigned\_per\_row, X6=assigned\_per\_row)}, the parent assignment for \texttt{X5} is \texttt{X3=0, X4=0, X6=0, X7=0}; gold outputs \texttt{X5=1}, while the submission outputs \texttt{X5=0}. \\
Training support & The exposing parent assignment occurs zero times in the training rows for \texttt{X5}. \\
\bottomrule
\end{tabular}
\end{center}
}

\paragraph{Case 3: A supplied SCM changes the task.}
For a linked Hidden-order and Alternative-SCM pair, GPT-5.4 fails the discovery version but succeeds once a valid reference SCM is supplied. On Hidden-order, the model returns an executable but non-exact SCM: TrainWorldExact is 0.444 and HeldoutWorldExact is 0.250. In Alternative-SCM, the same model constructs a valid, train-exact, semantically distinct alternative together with a separating intervention and witness. This mirrors the main-table pattern: editing an executable causal object is much easier once the object is already supplied.

{\footnotesize
\begin{center}
\setlength{\tabcolsep}{3pt}
\renewcommand{\arraystretch}{1.08}
\begin{tabular}{@{}p{0.24\linewidth}p{0.70\linewidth}@{}}
\toprule
Item & Detail \\
\midrule
Hidden-order task key error & The hidden submission uses \texttt{X7 = (xor X1 X3 X8)} and \texttt{X6 = X7}, with \texttt{X5 = (and (not (and X1 X6)) (or X4 X8 (not X2)))}. \\
Supplied reference structure & \texttt{X1 = (xor X3 X8)}, \texttt{X2 = (xor X3 X8)}, \texttt{X6 = (xor X1 X2)}, \texttt{X7 = (xor X1 X2)}, and \texttt{X5 = (iff X4 (xor (and X1 X6) (or X2 X4)))}. \\
Submitted alternative & \texttt{X1 = (xor X3 X8)}, \texttt{X2 = (xor X3 X8)}, \texttt{X7 = (xor X1 X2)}, \texttt{X6 = X7}, and \texttt{X5 = (iff X4 (xor (and X1 X6) (or X2 X4)))}. \\
Separating experiment and witness & \texttt{hard\_do} with \texttt{X7=1}; witness roots \texttt{X4=1, X8=0, X3=1}. The reference gives \texttt{X6=0, X5=1}; the alternative gives \texttt{X6=1, X5=0}. PairDisagreementRate is 1.000 and CellDifferenceRate is 0.375. \\
\bottomrule
\end{tabular}
\end{center}
}

\paragraph{Case 4: Short formulas can still fail.}
AST size is useful for diagnosing behavior, but smaller is not automatically better. In an Ordered item, GPT-5.4 returns a valid train-exact SCM whose total AST size is 18, compared with 28 for the gold mechanism map. The submission is almost a compressed version of the gold map: \texttt{X1} is equivalent up to reordering, \texttt{X5} is a Boolean simplification of the gold expression, and \texttt{X7} agrees on the training worlds. The problem is \texttt{X2}. The model replaces a three-parent mechanism with the much shorter \texttt{(or X6 X7)}, fits every training world, and then fails on a held-out intervention.

{\footnotesize
\begin{center}
\setlength{\tabcolsep}{3pt}
\renewcommand{\arraystretch}{1.08}
\begin{tabular}{@{}p{0.24\linewidth}p{0.70\linewidth}@{}}
\toprule
Item & Detail \\
\midrule
Outcome & Valid and train-exact; HeldoutWorldExact 0.875 and cell-level held-out accuracy 0.995. \\
AST contrast & Candidate total AST 18 versus gold total AST 28. \\
Gold \texttt{X2} & \texttt{(xor X1 X4 (and X4 (xor X1 X4 X7)))} \\
Submitted \texttt{X2} & \texttt{(or X6 X7)} \\
Held-out separator & Under \texttt{do(X4=assigned\_per\_row, X7=assigned\_per\_row)}, with \texttt{X1=1, X4=1, X6=1, X7=0}, gold gives \texttt{X2=0}; the shorter submission predicts \texttt{X2=1}. \\
\bottomrule
\end{tabular}
\end{center}
}

\paragraph{Case 5: Train-exact bloat can overfit observed worlds.}
The opposite failure also appears. In another Ordered item, GPT-5.4 returns a valid train-exact SCM with total AST size 95, while the gold map has total AST size 23. The submission keeps some simple equations exactly, but inflates others into DNF-like clauses with many negated guards and extra parents. This achieves training fit at the expense of robustness: HeldoutWorldExact is 0.375 and cell-level held-out accuracy is 0.848. The first held-out mismatch already shows the problem. The gold \texttt{X1} is the simple mechanism \texttt{(xor X2 X5)}, while the submitted \texttt{X1} adds a spurious dependence on \texttt{X6}.

{\footnotesize
\begin{center}
\setlength{\tabcolsep}{3pt}
\renewcommand{\arraystretch}{1.08}
\begin{tabular}{@{}p{0.24\linewidth}p{0.70\linewidth}@{}}
\toprule
Item & Detail \\
\midrule
Outcome & Valid and train-exact; HeldoutWorldExact 0.375 and cell-level held-out accuracy 0.848. \\
AST contrast & Candidate total AST 95 versus gold total AST 23; candidate parent references 18 versus 14 in the gold map. \\
Gold \texttt{X1} & \texttt{(xor X2 X5)} \\
Submitted \texttt{X1} & \texttt{(or (and (not X2) X5) X6)} \\
Held-out separator & Under \texttt{do(X3=assigned\_per\_row, X5=assigned\_per\_row)}, with \texttt{X2=0, X5=0, X6=1}, gold gives \texttt{X1=0}; the bloated submission predicts \texttt{X1=1}. \\
\bottomrule
\end{tabular}
\end{center}
}

\paragraph{Case 6: Long formulas can still replay correctly.}
A final case illustrates why the evaluator prioritizes replay behavior over formula style. In one Ordered item, Grok 4 returns a much larger valid SCM: total AST 100 versus 39 for the gold map. Several local mechanisms are expanded into longer case-like formulas, and the submitted map uses 22 parent references, compared with 19 in the gold map. Unlike the previous case, however, the submitted SCM is exact on both training and held-out worlds: TrainWorldExact, HeldoutWorldExact, and cell-level held-out accuracy are all 1.000. This leaves global equivalence beyond the benchmark worlds unresolved, but it shows the intended scoring principle. The benchmark requires an executable causal object that survives replay, independent of gold-string matching or formula length.

{\footnotesize
\begin{center}
\setlength{\tabcolsep}{3pt}
\renewcommand{\arraystretch}{1.08}
\begin{tabular}{@{}p{0.24\linewidth}p{0.70\linewidth}@{}}
\toprule
Item & Detail \\
\midrule
Outcome & Valid, train-exact, and strict held-out exact; HeldoutWorldExact and cell-level held-out accuracy are both 1.000. \\
AST contrast & Candidate total AST 100 versus gold total AST 39. \\
Gold \texttt{X3} & \texttt{(xor (and X5 X7) (or X4 X6))} \\
Submitted \texttt{X3} & \texttt{(or (and X7 (not (or (and X5 X4) (and X5 X6) (not (or X5 X4 X6))))) (and (not X7) (or X4 X6)))} \\
Interpretation & The formula is larger and less transparent, but it replays every benchmark world exactly. Long formulas can help interpret model behavior, but the benchmark scores replay behavior, not formula brevity. \\
\bottomrule
\end{tabular}
\end{center}
}

\end{document}